%% file: emnlp2022.tex
\pdfoutput=1

\documentclass[11pt]{article}

\usepackage{emnlp2022}

\usepackage{times}
\usepackage{latexsym}

\usepackage[T1]{fontenc}

\usepackage[utf8]{inputenc}

\usepackage{microtype}

\usepackage{graphicx}

\usepackage{amsmath}

\usepackage{amsfonts}
\usepackage{multirow}
\usepackage{enumitem}
\usepackage{multicol}
\usepackage{float}
\usepackage{boxedminipage} 

\usepackage{booktabs}
\usepackage{bbm}

\usepackage{hyperref}

\usepackage{xcolor} 

\definecolor{gred}{RGB}{219,68,55}
\definecolor{gblue}{RGB}{66,133,244}
\definecolor{ggreen}{RGB}{30,150,30}

\newcommand{\colorred}[1]{\textcolor{gred}{\textbf{#1}}} 
\newcommand{\colorblue}[1]{\textcolor{gblue}{\textbf{#1}}} 
\newcommand{\colorgreen}[1]{\textcolor{ggreen}{\textbf{#1}}} 

\def\ev{\mathbf{e}}
\def\xv{\mathbf{x}}
\def\yv{\mathbf{y}}
\def\wv{\mathbf{w}}
\def\Dc{\mathcal{D}}
\def\Ac{\mathcal{A}}

\graphicspath{ {./figures/} }

\title{Active Learning for Abstractive Text Summarization}

\author{Akim Tsvigun\textsuperscript{1,2}, 
Ivan Lysenko\textsuperscript{2}, 
Danila Sedashov\textsuperscript{2}, 
Ivan Lazichny\textsuperscript{1}, \\ \bf
Eldar Damirov\textsuperscript{2,4}, 
Vladimir Karlov\textsuperscript{2}, 
Artemy Belousov\textsuperscript{2}, 
Leonid Sanochkin\textsuperscript{1,2}, \\ \bf
Maxim Panov\textsuperscript{7},
Alexander Panchenko\textsuperscript{3}, 
Mikhail Burtsev\textsuperscript{1,5}, \\ \bf 
Artem Shelmanov\textsuperscript{1,6,8} \\
\textsuperscript{1}AIRI,
\textsuperscript{2}HSE, 
\textsuperscript{3}Skoltech,
\textsuperscript{4}SberDevices, 
\textsuperscript{5}MIPT,
\textsuperscript{6}MBZUAI, 
\textsuperscript{7}TII, \\
\textsuperscript{8}ISP RAS Research Center for Trusted Artificial Intelligence \\
\href{mailto:tsvigun@airi.net}{\{tsvigun, shelmanov\}@airi.net}, \href{mailto:artem.shelmanov@mbzuai.ac.ae}{artem.shelmanov@mbzuai.ac.ae}
}

\begin{document}

\maketitle

\begin{abstract}
Construction of human-curated annotated datasets for abstractive text summarization (ATS) is very time-consuming and expensive because creating each instance requires a human annotator to read a long document and compose a shorter summary that would preserve the key information relayed by the original document. Active Learning (AL) is a technique developed to reduce the amount of annotation required to achieve a certain level of machine learning model performance. In information extraction and text classification, AL can reduce the amount of labor up to multiple times. Despite its potential for aiding expensive annotation, as far as we know, there were no effective AL query strategies for ATS. This stems from the fact that many AL strategies rely on uncertainty estimation, while as we show in our work, uncertain instances are usually noisy, and selecting them can degrade the model performance compared to passive annotation. We address this problem by proposing the first effective query strategy for AL in ATS based on diversity principles. We show that given a certain annotation budget, using our strategy in AL annotation helps to improve the model performance in terms of ROUGE and consistency scores. Additionally, we analyze the effect of self-learning and show that it can further increase the performance of the model. 
\end{abstract}

\section{Introduction}

Abstractive text summarization (ATS) aims to compress a document into a brief yet informative and readable summary, which would retain the key information of the original document. State-of-the-art results in this task are achieved by neural seq-to-seq models~\cite{lewis_bart,zhang_pegasus,qi_prophetnet,long_t5,simcls} based on the Transformer architecture~\cite{vaswani_attention}. Training a model for ATS requires a dataset that contains pairs of original documents and their short summaries, which are usually written by human annotators. Manually composing a summary is a very tedious task, which requires one to read a long original document, select crucial information, and finally write a small text. Each of these steps is very time-consuming, resulting in the fact that constructing each instance in annotated corpora for text summarization is very expensive. 

Active Learning (AL; \citet{cohn1996active}) is a well-known technique that helps to substantially reduce the amount of annotation required to achieve a certain level of machine learning model performance. For example, in tasks related to named entity recognition, researchers report annotation reduction by 2-7 times with a loss of only 1\% of  F1-score~\cite{settles-craven-2008-analysis}. This makes AL especially important when annotation is expensive, which is the case for ATS.

AL works iteratively: on each iteration, (1) a model is trained on the so far annotated dataset; (2) the model is used to select some informative instances from a large unlabeled pool using a \textit{query strategy}; (3) informative instances are presented to human experts, which provide gold-standard annotations; (4) finally, the instances with annotations are added to the labeled dataset, and a new iteration begins. Traditional AL query strategies are based on uncertainty estimation techniques \cite{lc, bt_1}. The hypothesis is that the most uncertain instances for the model trained on the current iteration are informative for training the model on the next iteration. We argue that uncertain predictions of ATS models (uncertain summaries) are not more useful than randomly selected instances. Moreover, usually, they introduce more noise and detriment to the performance of summarization models. Therefore, it is not possible to straightforwardly adapt the uncertainty-based approach to AL in text summarization.

In this work, we present the first effective query strategy for AL in ATS, which we call in-domain diversity sampling (IDDS). It is based on the idea of the selection of diverse instances that are semantically dissimilar from already annotated documents but at the same time similar to the core documents of the considered domain. 
The empirical investigation shows that while techniques based on uncertainty cannot overcome the random sampling baseline, IDDS substantially increases the performance of summarization models. We also experiment with the self-learning technique that leverages a training dataset expanded with summaries automatically generated by an ATS model trained only on the human-annotated dataset. This approach shows improvements when one needs to generate short summaries. The code for reproducing the experiments is available online\footnote{\url{https://github.com/AIRI-Institute/al_ats}}. The contributions of this paper are the following:
\begin{itemize}[itemsep=1mm, parsep=0pt]
    \item We propose the first effective AL query strategy for ATS that beats the random sampling baseline.
    \item We conduct a vast empirical investigation and show that in contrast to such tasks as text classification and information extraction, in ATS, uncertainty-based AL query strategies cannot outperform the random sampling baseline.
    \item To our knowledge, we are the first to investigate the effect of self-learning in conjunction with AL for ATS and demonstrate that it can substantially improve results on the datasets with short summaries.
\end{itemize}

\section{Related Work}

\paragraph{Abstractive Text Summarization.}
The advent of seq2seq models~\citep{sutskever_seq2seq} along with the development of the attention mechanism~\cite{bahdanau_attention} consolidated neural networks as a primary tool for ATS. The attention-based Transformer~\citep{vaswani_attention} architecture has formed the basis of many large-scale pre-trained language models that achieve state-of-the-art results in ATS~\citep{lewis_bart,zhang_pegasus,qi_prophetnet, long_t5}.
Recent efforts in this area mostly focus on minor modifications of the existing architectures \cite{simcls, bart_rxf, brio}. 

\paragraph{Active Learning in Natural Language Generation.}
While many recent works leverage AL for text classification or sequence-tagging tasks~\citep{yuan_cold_al,zhang_cartography,shelmanov_active,margatina_bayesian_al}, little attention has been paid to natural language generation tasks. Among  the works in this area, it is worth mentioning~\cite{haffari_stat_mt_al,ambati_smt,ananthakrishnan_smt}. These works focus on neural machine translation (NMT) and suggest several uncertainty-based query strategies for AL. \newcite{peris_imt} successfully apply AL in the interactive machine translation. \citet{liu_translation_al_rl} exploit reinforcement learning to train a policy-based query strategy for NMT. Although there is an attempt to apply AL in ATS \cite{gidiotis_al}, to the best of our knowledge, there is no published work on this topic yet. 

\paragraph{Uncertainty Estimation in Natural Language Generation.}
A simple yet effective approach for uncertainty estimation in generation is proposed by \newcite{wang_unc_back}. 
They use a combination of expected translation probability and variance of the translation probability, demonstrating that it can handle noisy instances better and noticeably improve the quality of back-translation.
\newcite{malinin_uncertainty_autoregressive} investigate the ensemble-based measures of uncertainty for NMT. Their results demonstrate the superiority of these methods for OOD detection and for identifying generated translations of low-quality. \newcite{xiao_wat_zei} propose a method for uncertainty estimation of long sequences of discrete random variables, which they dub ``BLEU Variance'', and apply it for OOD sentence detection in NMT. It is also shown to be useful for identifying instances of questionable quality in ATS \cite{gidiotis_unc_as}. In this work, we investigate these uncertainty estimation techniques in AL and show that they do not provide any benefits over annotating randomly selected instances.

\paragraph{Diversity-based Active Learning.}
Along with the uncertainty-based query strategies, a series of diversity-based methods have been suggested for AL \cite{mmr_al,sener2018active,ash2019deep,citovsky2021batch}. The most relevant work among them is \cite{mmr_al}, where the authors propose to use a Maximal Marginal Relevance (MMR; \citet{mmr})-based function as a query strategy in  AL for named entity recognition. This function aims to capture uncertainty and diversity and selects instances for annotation based on these two perspectives. We adapt this strategy for the ATS task and compare the proposed method with it.



\section{Uncertainty-based Active Learning for Text Generation}

In this section, we give a brief formal definition of the AL procedure for text generation and uncertainty-based query strategies. 
Here and throughout the rest of the paper, we denote an input sequence as $\xv = (x_1 \dots x_m$) and the output sequence as $\yv = (y_1 \dots y_n$), with $m$ and $n$ being lengths of $\xv$ and $\yv$ respectively.

Let $\Dc = \{(\xv^{(k)}, \yv^{(k)})\}_{k = 1}^K$ be a dataset of pairs (documents, summaries). Consider a probabilistic model $p_\wv(\yv \mid \xv)$ parametrized by a vector $\wv$. Usually, $p_\wv(\yv \mid \xv)$ is a neural network, while the parameter estimation is done via the maximum likelihood approach:
\begin{equation}
    \hat{\wv} = \arg\max_\wv L(\Dc, \wv),
\end{equation}
where $L(\Dc, \wv) = \sum_{k = 1}^K \log p_\wv(\yv^{(k)} \mid \xv^{(k)})$ is log-likelihood.

Many AL methods are based on greedy query strategies that select instances for annotation, optimizing a certain criterion $\Ac(\xv \mid \Dc, \hat{\wv})$ called an \textit{acquisition function}:
\begin{equation}
    \xv^* = \arg\max_{\xv} \Ac(\xv \mid \Dc, \hat{\wv}).
\end{equation}
The selected instance $\xv^*$ is then annotated with a target value $\yv^*$ (document summary) and added to the training dataset: $\Dc := \Dc \cup (\xv^*, \yv^*)$. Subsequently, the model parameters $\wv$ are updated and the instance selection process continues until the desired model quality is achieved or the available annotation budget is depleted.

The right choice of an acquisition function is crucial for AL. A common heuristic for acquisition is selecting instances with high uncertainty.  Below, we consider several measures of uncertainty used in text generation. 




\paragraph{Normalized Sequence Probability (NSP)} was originally proposed by~\newcite{ueffing_sp} and has been used in many subsequent works~\citep{haffari_stat_mt_al,wang_unc_back,xiao_wat_zei,lyu_qa}. This measure is given by
\begin{equation}
    \text{NSP}(\xv) = 1 - \bar{p}_{\hat{\wv}}(\yv \mid \xv),
\end{equation}
where we define the geometric mean of probabilities of tokens predicted by the model as: $\bar{p}_{\hat{\wv}}(\yv \mid \xv) = \exp\bigl\{\frac{1}{n} \log p_{\hat{\wv}}(\yv \mid \xv)\bigr\}$ .

A wide family of uncertainty measures can be derived using the Bayesian approach to modeling. Under the Bayesian approach, it is assumed that model parameters have a prior distribution $\pi(\wv)$. Optimization of the log-likelihood $L(\Dc, \wv)$ in this case leads to the optimization of the posterior distribution of the model parameters:
\begin{equation}
    \pi(\wv \mid \Dc) \propto \exp\{L(\Dc, \wv)\} \cdot \pi(\wv).
\end{equation}
Usually, the exact computation of the posterior is intractable, and to perform training and inference, a family of distributions $q_{\theta}(\wv)$ parameterized by $\theta$ is introduced. The parameter estimate $\hat{\theta}$ minimizes the KL-divergence between the true posterior $\pi(\wv \mid \Dc)$ and the approximation $q_{\hat{\theta}}(\wv)$. Given such an approximation, several uncertainty measures can be constructed.

\paragraph{Expected Normalized Sequence Probability (ENSP)} is proposed by \newcite{wang_unc_back} and is also used in \citep{xiao_wat_zei,lyu_qa}: 
\begin{equation}
    \text{ENSP}(\xv) = 1 - \mathbb{E}_{\wv \sim q_{\hat{\theta}}} \bar{p}_\wv(\yv \mid \xv).
\end{equation}
In practice, the expectation is approximated via Monte Carlo dropout~\cite{gal_mc_dropout}, i.e. averaging multiple predictions obtained with activated dropout layers in the network.


\paragraph{Expected Normalized Sequence Variance (ENSV; \citet{wang_unc_back})} measures the variance of the sequence probabilities obtained via Monte Carlo dropout: 
\begin{equation}
    \text{ENSV}(\xv) = \text{Var}_{\wv \sim q_{\hat{\theta}}} \bar{p}_\wv(\yv \mid \xv).
\end{equation}

\paragraph{BLEU Variance (BLEUVar)} is proposed by \newcite{xiao_wat_zei}. 
It treats documents as points in some high dimensional space and uses the BLEU metric~\cite{bleu} for measuring the difference between them. In such a setting, it is possible to calculate the variance of generated texts in the following way:
\begin{align}
    & \quad \text{BLEUVar}(\xv) =  
    \\
    &= \mathbb{E}_{\wv \sim q_{\hat{\theta}}}\mathbb{E}_{\yv, \yv' \sim p_\wv(\cdot | \xv)} \bigl(1 - \text{BLEU}(\yv, \yv')\bigr)^2.
    \notag
\end{align}

The BLEU metric is calculated as a geometric mean of n-grams overlap up to 4-grams. Consequently, when summaries consist of less than 4 tokens, the metric is equal to zero since there will be no common 4-grams. This problem can be mitigated with the SacreBLEU metric~\cite{sacrebleu}, which smoothes the n-grams with zero counts. When we use this query strategy with the SacreBLUE metric, we refer to it as \textbf{SacreBLEUVar}.

\section{Proposed Methods}

\subsection{In-Domain Diversity Sampling}

\input{figures/idds}

We argue that uncertainty-based query strategies tend to select noisy instances that have little value for training ATS models. To alleviate this issue, we propose a novel query strategy named in-domain diversity sampling (IDDS). It aims to maximize the diversity of the annotated instances by selecting instances that are dissimilar from the already annotated ones. At the same time, it avoids selecting noisy outliers. These noisy documents that are harmful to training an ATS model are usually semantically dissimilar from the core documents of the domain represented by the unlabeled pool.
Therefore, IDDS queries instances that are dissimilar to the annotated instances but at the same time are similar to unannotated ones (Figure~\ref{fig:idds}).

We propose the following acquisition function that implements the aforementioned idea (the higher the value -- the higher the priority for the annotation):
\begin{equation}
    \text{IDDS}(\xv) = \lambda \frac{\sum\limits_{j = 1}^{|U|} s(\xv, \xv_j)}{|U|} - (1 - \lambda) \frac{\sum\limits_{i = 1}^{|L|} s(\xv, \xv_i)}{|L|},
\label{eq:idds}
\end{equation}
where $s(\xv, \xv')$ is a similarity function between texts, $U$ is the unlabeled set, $L$ is the labeled set, and $\lambda \in [0;1]$ is a hyperparameter. 

Below, we formalize the resulting algorithm of the IDDS query strategy.
\vspace{-0.2cm}
\begin{enumerate}[itemsep=1mm, parsep=0pt]
    \item For each document in the unlabeled pool $\xv$, we obtain an embedding vector $\ev(\xv)$. For this purpose, we suggest using the [CLS] pooled sequence embeddings from BERT. We note that using a pre-trained checkpoint straightforwardly may lead to unreasonably high similarity scores between instances since they all belong to the same domain, which can be quite specific. We mitigate this problem by using the task-adaptive pre-training (TAPT; \citet{tapt}) on the unlabeled pool. TAPT performs several epochs of self-supervised training of the pre-trained model on the target dataset to acquaint it with the peculiarities of the data. 
    
    \item Deduplicate the unlabeled pool. Instances with duplicates will have an overrated similarity score with the unlabeled pool.


    \item Calculate the informativeness scores using the IDDS acquisition function~\eqref{eq:idds}. As a similarity function, we suggest using a scalar product between document representations: $s(\xv, \xv') = \langle \ev(\xv), \ev(\xv') \rangle$.
\end{enumerate}

The idea of IDDS is close to the MMR-based strategy proposed in \cite{mmr_al}. Yet, despite the resemblance, IDDS differs from it in several crucial aspects. The MMR-based strategy focuses on the uncertainty and diversity components. However, as shown in Section~\ref{sec:exp_unc}, selecting instances by uncertainty leads to worse results compared to random sampling.
Consequently, instead of using uncertainty, IDDS leverages the unlabeled pool to capture the \textit{representativeness} of the instances. Furthermore, IDDS differs from the MMR-based strategy in how they calculate the diversity component. MMR  directly specifies the usage of the ``max'' aggregation function for calculating the similarity with the already annotated data, while IDDS uses ``average'' similarity instead and achieves better results as shown in Section~\ref{sec:idds_exps}. 

We note that IDDS does not require retraining an acquisition model in contrast to uncertainty-based strategies since document vector representations and document similarities can be calculated before starting the AL annotation process. This results in the fact that no heavy computations during AL are required. Consequently, IDDS does not harm the interactiveness of the annotation process, which is a common bottleneck \cite{plasm}.  


\subsection{Self-learning}

Pool-based AL assumes that there is a large unlabeled pool of data. We propose to use this data source during AL to improve text summarization models with the help of self-learning. 
We train the model on the labeled data and generate summaries for the whole unlabeled pool. Then, we concatenate the generated summaries with the labeled set and use this data to fine-tune the final model.
We note that generated summaries can be noisy: irrelevant, grammatically incorrect, contain factual inconsistency, and can harm the model performance. We detect such instances using the uncertainty estimates obtained via NSP scores and exclude $k_l$\% instances with the lowest scores and $k_h$\% of instances with the highest scores. We choose this uncertainty metric because according to our experiments in Section \ref{sec:exp_unc}, high NSP scores correspond to the noisiest instances. 
We note that adding the filtration step does not introduce additional computational overhead, since the NSP scores are calculated simultaneously with the summary generation for self-learning. 



\section{Experimental Setup}

\subsection{Active Learning Setting}

We evaluate IDDS and other query strategies using the conventional scheme of AL annotation emulation applied in many previous works~\citep{settles_al,shen_2017,siddhant_2018,shelmanov_active,dor2020active}. 
For uncertainty-based query strategies and random sampling, we start from a small annotated seeding set selected randomly. This set is used for fine-tuning the summarization model on the first iteration. For IDDS, the seeding set is not used, since this query strategy does not require fine-tuning the model to make a query. On each AL iteration, we select top-k instances from the unlabeled pool according to the informativeness score obtained with a query strategy. The selected instances with their gold-standard summaries are added to the so-far annotated set and are excluded from the unlabeled pool. On each iteration, we fine-tune a summarization model from scratch and evaluate it on a held-out test set. We report the performance of the model on each iteration to demonstrate the dynamics of the model performance depending on the invested annotation effort. 

The query size (the number of instances selected for annotation on each iteration) is set to 10 documents. We repeat each experiment 9 times with different random seeds and report the mean and the standard deviation of the obtained scores. For the WikiHow and PubMed datasets, on each iteration, we use a random subset from the unlabeled pool since generating predictions for the whole unlabeled dataset is too computationally expensive. In the experiments, the subset size is set to 10,000 for WikiHow and 1,000 for PubMed.

\subsection{Baselines}

We use random sampling as the main baseline. To our knowledge, in the ATS task, this baseline has not been outperformed by any other query strategy yet. In this baseline, an annotator is given randomly selected instances from the unlabeled pool, which means that AL is not used at all. We also report results of uncertainty-based query strategies and an MMR-based query strategy \cite{mmr_al} that is shown to be useful for named entity recognition.

\subsection{Metrics}

\paragraph{Quality of Text Summarization.}
To measure the quality of the text summarization model, we use the commonly adopted ROUGE metric~\cite{rouge}. Following previous works~\citep{see_2017,nallapati_2017,rouge_chen,lewis_bart, zhang_pegasus}, we report ROUGE-1, ROUGE-2, and ROUGE-L. Since we found the dynamics of these metrics coinciding, for brevity, in the main part of the paper, we keep only the results with the ROUGE-1 metric. The results with other metrics are presented in the appendix.



\paragraph{Factual Consistency.}
Inconsistency (hallucination) of the generated summaries is one of the most crucial problems in summarization~\cite{kryscinski_consistency,inconsistency_survey,improving_consistency,dynamics_goyal}. Therefore, in addition to the ROUGE metrics, we measure the factual consistency of the generated summaries with the original documents. We use the SummaC-ZS~\cite{summac} -- a state-of-the-art model for inconsistency detection. We set \textit{granularity = ``sentence''} and \textit{model\_name = ``vitc''}.

\input{figures/uncertainty_rouge1}

\input{figures/idds_rouge1}

\subsection{Datasets}

We experiment with three datasets widely-used for evaluation of ATS models: AESLC~\cite{aeslc}, PubMed~\cite{pubmed}, and WikiHow~\cite{wikihow}. AESLC consists of emails with their subject lines as summaries. WikiHow contains articles from WikiHow pages with their headlines as summaries. PubMed~\cite{pubmed} is a collection of scientific articles from the PubMed archive with their abstracts. The choice of datasets is stipulated by the fact that AESLC contains short documents and summaries, WikiHow contains medium-sized documents and summaries, and PubMed contains long documents and summaries. We also use two non-intersecting subsets of the Gigaword dataset~\citep{graff_gigaword, rush_gigaword} of sizes 2,000 and 10,000 for hyperparameter optimization of ATS models and  additional experiments with self-learning, respectively. Gigaword consists of news articles and their headlines representing summaries. The dataset statistics is presented in Table~\ref{tab:datasets_stats} in Appendix~\ref{app:data_stats}.

\subsection{Models and Hyperparameters}

We conduct experiments using the state-of-the-art text summarization models: BART~\cite{lewis_bart} and PEGASUS~\cite{zhang_pegasus}. In all experiments, we use the ``base'' pre-trained version of BART and the ``large'' version of PEGASUS. Most of the experiments are conducted with the BART model, while PEGASUS is only used for the AESLC dataset (results are presented in Appendices~\ref{app:unc}, ~\ref{app:idds}) since running it on two other datasets in AL introduces a computational bottleneck.

We tune hyperparameter values of ATS models using the ROUGE-L score on the subset of the Gigaword dataset. The hyperparameter values are provided in Table~\ref{tab:hyperparams} in Appendix~\ref{app:data_stats}.

For the IDDS query strategy, we use $\lambda = 0.67$. We analyze the effect of different values of this parameter in Section~\ref{sec:idds_exps}.

\input{figures/examples}

\section{Results and Discussion}

\subsection{Uncertainty-based Query Strategies}
\label{sec:exp_unc}

In this series of experiments, we demonstrate that selected uncertainty-based query strategies are not suitable for AL in ATS. Figure~\ref{fig:unc_rouge1}a and Figures~\ref{fig:aeslc_uncertainty}, \ref{fig:aeslc_uncertainty_pegasus} in Appendix~\ref{app:unc} present the results on the AESLC dataset. As we can see, none of the uncertainty-based query strategies outperform the random sampling baseline for both BART and PEGASUS models. NSP and ENSP strategies demonstrate the worst results with the former having the lowest performance for both ATS models. 
Similar results are obtained for the WikiHow and PubMed datasets (Figures~\ref{fig:unc_rouge1}b and~\ref{fig:unc_rouge1}c). 


In some previous work on NMT, uncertainty-based query strategies outperform the random sampling baseline \cite{haffari_stat_mt_al,ambati_smt,ananthakrishnan_smt}. Their low results for ATS compared to NMT might stem from the differences between these tasks. Both NMT and ATS are seq2seq tasks and can be solved via similar models. However, NMT is somewhat easier, since the output is usually of similar length as the input and its variability is smaller. It is much easier to train a model to reproduce an exact translation rather than make it generate an exact summary. Therefore, uncertainty estimates of ATS models are way less reliable than estimates of translation models. These estimates often select for annotation noisy documents that are useless or even harmful for training summarization models. Table~\ref{tab:examples} reveals several documents selected by the worst-performing strategy NSP on AESLC. We can see that NSP selects 
domain-irrelevant documents or very specific ones. Their summaries can hardly be restored from the source documents, which means that they most likely have little positive impact on the generalization ability of the model. More examples of instances selected by different query strategies are presented in Table~\ref{tab:larger_examples} in Appendix~\ref{app:diversity}.


\subsection{In-Domain Diversity Sampling} 
\label{sec:idds_exps}

In this series of experiments, we analyze the proposed IDDS query strategy. Figure ~\ref{fig:idds_rouge1}a and Figures~\ref{fig:aeslc_embs}, ~\ref{fig:aeslc_embs_pegasus} in Appendix~\ref{app:idds} show the performance of the models with various query strategies on AESLC. We can see that the proposed strategy outperforms random sampling on all iterations for both ATS models and subsequently outperforms the uncertainty-based strategy NSP. 
IDDS demonstrates similar results on the WikiHow and PubMed datasets, outperforming the baseline with a large margin as depicted in Figures~\ref{fig:idds_rouge1}b and~\ref{fig:idds_rouge1}c. We also report the improvement of IDDS over random sampling in percentage on several AL iterations in Table~\ref{tab:increase_idds}. We can see that IDDS provides an especially large improvement in the cold-start AL scenario when the amount of labeled data is very small.


We carry out several ablation studies for the proposed query strategy. First, we investigate the effect of various models for document embeddings construction and the necessity of performing TAPT. Figures~\ref{fig:as_emb_model_aeslc} and~\ref{fig:as_emb_model_wikiall} in Appendix~\ref{app:ablation_embs} illustrate that TAPT substantially enhances the performance of IDDS. Figure~\ref{fig:as_emb_model_aeslc} also shows that the BERT-base encoder appears to be better than SentenceBERT~\cite{sentbert} and LongFormer~\cite{longformer}.

Second, we try various functions for calculating the similarity between instances. Figures~\ref{fig:as_sim_aeslc}, \ref{fig:as_sim_wiki} in Appendix~\ref{app:ablation_embs} compare the originally used dot product with Mahalanobis and Euclidean distances on AESLC and WikiHow. On AESLC, IDDS with Mahalanobis distance persistently demonstrates lower performance. IDDS with the Euclidean distance shows a performance drop on the initial AL iterations compared to the dot product. On WikiHow, however, all the variants perform roughly the same. Therefore, we suggest keeping the dot product for computing the document similarity in IDDS since it provides the most robust results across the datasets.

We also compare the dot product with its normalized version -- cosine similarity on AESLC and PubMed, see Figures~\ref{fig:as_norm_aeslc} and~\ref{fig:as_norm_pubmed} in Appendix~\ref{app:ablation_embs}. On both datasets, adding normalization leads to substantially worse results on the initial AL iterations. We deem that this happens because normalization may damage the representativeness component since the norm of the embedding can be treated as a measure of the representativeness of the corresponding document.


Third, we investigate how different values for the lambda coefficient influence the performance of IDDS. Table~\ref{tab:lamb_ae} and Figure~\ref{fig:as_lambda_aeslc} in Appendix~\ref{app:ablation_embs} shows that smaller values of $\lambda \in \{0, 0.33, 0.5\}$ substantially deteriorate the performance. Smaller values correspond to selecting instances that are highly dissimilar from the documents in the unlabeled pool, which leads to picking many outliers. Higher values lead to the selection of instances from the core of the unlabeled dataset, but also very similar to the annotated part. This also results in a lower quality on the initial AL iterations. The best and most stable results are obtained with $\lambda=0.67$.

Fourth, we compare IDDS with the MMR-based strategy suggested in \cite{mmr_al}. Since it uses uncertainty, it requires a trained model to calculate the scores. Consequently, the initial query is taken randomly as no trained model is available on the initial AL iteration. Therefore, we use the modification, when the initial query is done with IDDS because it provides substantially better results on the initial iteration.
We also experiment with different values of the $\lambda$ hyperparameter of the MMR-based strategy.
Figure~\ref{fig:as_mmr_aeslc} illustrates a large gap in performance of IDDS and the MMR-based strategy regardless of the initialization / $\lambda$ on AESLC. We believe that this is attributed to the fact that strategies incorporating uncertainty are harmful to AL in ATS as shown in Section~\ref{sec:exp_unc}.

Finally, we compare ``aggregation'' functions for estimating the similarity between a document and a collection of documents (labeled and unlabeled pools). Following the MMR-based strategy \cite{mmr_al}, instead of calculating the \textit{average} similarity between the embedding of the target document and the embeddings of documents from the labeled set, we calculate the \textit{maximum} similarity.
We also try replacing the ``average'' aggregation function with ``maximum'' in both IDDS components in ~\eqref{eq:idds}. 
Figures~\ref{fig:as_agg_aeslc} and \ref{fig:as_agg_wiki} in Appendix~\ref{app:ablation_embs} show that \textit{average} leads to better performance on both AESLC and WikiHow datasets. 


The importance of diversity sampling is illustrated in Table~\ref{tab:diversity} in Appendix~\ref{app:diversity}. We can see that NSP-based query batches contain a large number of overlapping instances. This may partly stipulate the poor performance of the NSP strategy since almost 9\% of labeled instances are redundant. IDDS, on the contrary, does not have instances with overlapping summaries inside batches at all.



\subsection{Self-learning}
\label{sec:self_sup_learning}

\input{figures/self_supervised_rouge1}

In this section, we investigate the effect of self-learning in the AL setting. Figures~\ref{fig:self_supervised_rouge1}a, ~\ref{fig:self_supervised_rouge1}b illustrate the effect of self-learning on the AESLC and Gigaword datasets. For this experiment, we use $k_l = 10, k_h = 1$, filtering out 11\% of automatically generated summaries. In both cases: with AL and without, adding automatically generated summaries of documents from the unlabeled pool to the training set improves the performance of the summarization model. On AESLC, the best results are obtained with both AL and self-learning: their combination achieves up to 58\% improvement in all ROUGE metrics compared to using passive annotation without self-learning.

The same experiment on the WikiHow dataset is presented in Figure~\ref{fig:self_supervised_rouge1}c. To make sure that the quality is not deteriorated due to the addition of noisy uncertain instances, we use $k_l = 38, k_h = 2$ for this experiment, filtering out 40\% of automatically generated summaries. On this dataset, self-learning reduces the performance for both cases (with AL and without). We deem that the benefit of self-learning depends on the length of the summaries in the dataset. AESLC and Gigaword contain very short summaries (less than 13 tokens on average, see Table~\ref{tab:datasets_stats}). Since the model is capable of generating short texts that are grammatically correct and logically consistent, such data augmentation does not introduce much noise into the dataset, resulting in performance improvement. WikiHow, on the contrary, contains \textit{long} summaries (77 tokens on average). Generation of long, logically consistent, and grammatically correct summaries is still a challenging task even for the state-of-the-art ATS models. Therefore, the generated summaries are of low quality, and using them as an additional training signal deteriorates the model performance. Consequently, we suggest using self-learning only if the dataset consists of relatively short texts. We leave a more detailed investigation of this topic for future research.

\subsection{Consistency}

We analyze how various AL strategies and self-learning affect the consistency of model output in two ways. We measure the consistency of the generated summaries with the original documents on the test set on each AL iteration. Figure~\ref{fig:aeslc_consistency_model_bart} shows that the model trained on instances queried by IDDS generates the most consistent summaries across all considered AL query strategies on AESLC. On the contrary, the model trained on the instances selected by the uncertainty-based NSP query strategy generates summaries with the lowest consistency. 

\input{figures/aeslc_consistency_model_bart}

Figure~\ref{fig:aeslc_consistency_pl} in Appendix~\ref{app:consist} demonstrates that on AESLC, self-learning also improves consistency regardless of the AL strategy. The same trend is observed on Gigaword (Figure~\ref{fig:gigaword_consistency_pl} in Appendix~\ref{app:consist}).


However, for WikiHow, there is no clear trend. Figure~\ref{fig:wikihow_consistency_model_bart} in Appendix~\ref{app:consist} shows that all query strategies lead to similar consistency results, with NSP producing slightly higher consistency, and BLEUVar -- slightly lower. We deem that this may be due to the fact that summaries generated by the model on WikiHow are of lower quality than the golden summaries regardless of the strategy. Therefore, this leads to biased scores of the SummaC model with similar results on average.


\subsection{Query Duration}

We compare the average duration of AL iterations for various query strategies. Figure~\ref{fig:time_strategy_aeslc} in the Appendix~\ref{app:query_dur} presents the average training time and the average duration of making a query. We can see that training a model takes considerably less time than selecting the instances from the unlabeled pool for annotation. Therefore, the duration of AL iterations is mostly determined by the efficiency of the query strategy. 
The IDDS query strategy does not require any heavy computations during AL, which makes it also the best option for keeping the AL process interactive.


\section{Conclusion}

In this work, we convey the first study of AL in ATS and propose the first active learning query strategy that outperforms the baseline random sampling. The query strategy aims at selecting for annotation the instances with high similarity with the documents in the unlabeled pool and low similarity with the already annotated documents. It outperforms the random sampling in terms of ROUGE metrics on all considered datasets. It also outperforms random sampling in terms of the consistency score calculated via the SummaC model on the AESLC dataset. We also demonstrate that uncertainty-based query strategies fail to outperform random sampling, resulting in the same or even worse performance. Finally, we show that self-learning can improve the performance of an ATS model in terms of both the ROUGE metrics and consistency. This is especially favorable in AL since there is always a large unlabeled pool of data. We show that combining AL and self-learning can give an improvement of up to 58\% in terms of ROUGE metrics.

In future work, we look forward to investigating IDDS in other sequence generation tasks. This query strategy might be beneficial for tasks with the highly variable output when uncertainty estimates of model predictions are unreliable and cannot outperform the random sampling baseline. IDDS facilitates the representativeness of instances in the training dataset without leveraging uncertainty scores.


\section*{Limitations}

Despite the benefits, the proposed methods require some conditions to be met to be successfully applied in practice. IDDS strategy requires making TAPT of the embeddings-generated model, which may be computationally consuming for a large dataset.
Self-learning, in turn, may harm the performance when the summaries are too long, as shown in Section~\ref{sec:self_sup_learning}. Consequently, its application requires a detailed analysis of the properties of the target domain summaries.

\section*{Ethical Considerations}

It is important to note that active learning is a method of biased sampling, which can lead to biased annotated corpora. Therefore, active learning can be used to deliberately increase the bias in the datasets. Our research improves the active learning performance; hence, our contribution would also make it more efficient for introducing more bias as well. We also note that our method uses the pre-trained language models, which usually contain different types of biases by themselves. Since bias affects all applications of pre-trained models, this can also unintentionally facilitate the biased selection of instances for annotation during active learning.

\section*{Acknowledgements}
We thank anonymous reviewers for their insightful
suggestions to improve this paper. The work was supported by a grant for research centers in the field of artificial intelligence (agreement identifier 000000D730321P5Q0002 dated November 2, 2021 No. 70-2021-00142 with ISP RAS). This research was supported in part by computational resources of the HPC facilities at the HSE University \cite{Kostenetskiy_2021}.


\bibliography{anthology,custom}
\bibliographystyle{acl_natbib}

\clearpage
\appendix

\onecolumn
\section{Dataset Statistics and Model Hyperparameters}
\label{app:data_stats}

\input{figures/datasets_stats}
\input{figures/hyperparameters}

\clearpage
\onecolumn
\section{Full Results for Uncertainty-based Methods}
\label{app:unc}

\begin{figure*}[ht!]
\input{figures/aeslc_uncertainty_bart}
\vspace{0.5cm}
\input{figures/aeslc_uncertainty_pegasus}
\vspace{0.5cm}
\input{figures/wikiall_uncertainty_bart}
\vspace{0.5cm}
\input{figures/pubmed_uncertainty_bart}

\end{figure*}

\clearpage
\onecolumn
\section{Full Results for IDDS}
\label{app:idds}

\input{figures/increase_idds}

\vspace{0.2cm}
\input{figures/aeslc_embs_bart}
\input{figures/aeslc_embs_pegasus}
\input{figures/wikiall_embs_bart}
\input{figures/pubmed_embs_bart}

\clearpage
\onecolumn
\section{Full Results for Self-learning}
\label{app:self_learning}

\input{figures/aeslc_pl_bart}
\input{figures/gigaword_pl_bart}
\input{figures/pubmed_pl_bart}

\clearpage
\onecolumn
\section{Diversity Statistics and Query Examples}
\label{app:diversity}

\input{figures/larger_examples}
\vspace{0.7cm}
\input{figures/diversity_table}

\clearpage
\onecolumn
\section{Ablation Studies of IDDS}
\label{app:ablation_embs}

\input{figures/ablation_1_emb_model_aeslc}
\vspace{0.3cm}
\input{figures/ablation_1_emb_model_wikiall}
\vspace{0.3cm}
\input{figures/ablation_2_sim_functions_aeslc}
\vspace{0.3cm}
\input{figures/ablation_2_sim_functions_wikihow}

\input{figures/ablation_2_normalization_aeslc}
\vspace{1.cm}
\input{figures/ablation_2_normalization_pubmed}
\vspace{1.cm}
\input{figures/ablation_3_lambda_aeslc}
\vspace{1.cm}
\input{figures/aeslc_table_lambda}
\input{figures/ablation_4_mmr_aeslc}
\vspace{0.5cm}
\input{figures/ablation_4_aggregation_aeslc}
\vspace{0.5cm}
\input{figures/ablation_4_aggregation_wikihow}

\clearpage
\section{Additional Experiments with Consistency Analysis}
\label{app:consist}

\input{figures/gigaword_consistency_pl.tex}
\input{figures/aeslc_consistency_pl}
\input{figures/wikihow_consistency_model_bart}

\section{Query Duration}
\label{app:query_dur}

\input{figures/time_strategy_tex.tex}

\end{document}

%% file: figures/idds.tex
\begin{figure}[t]
    \vspace{-0.1cm}
	\center{\includegraphics[width=1.\linewidth]{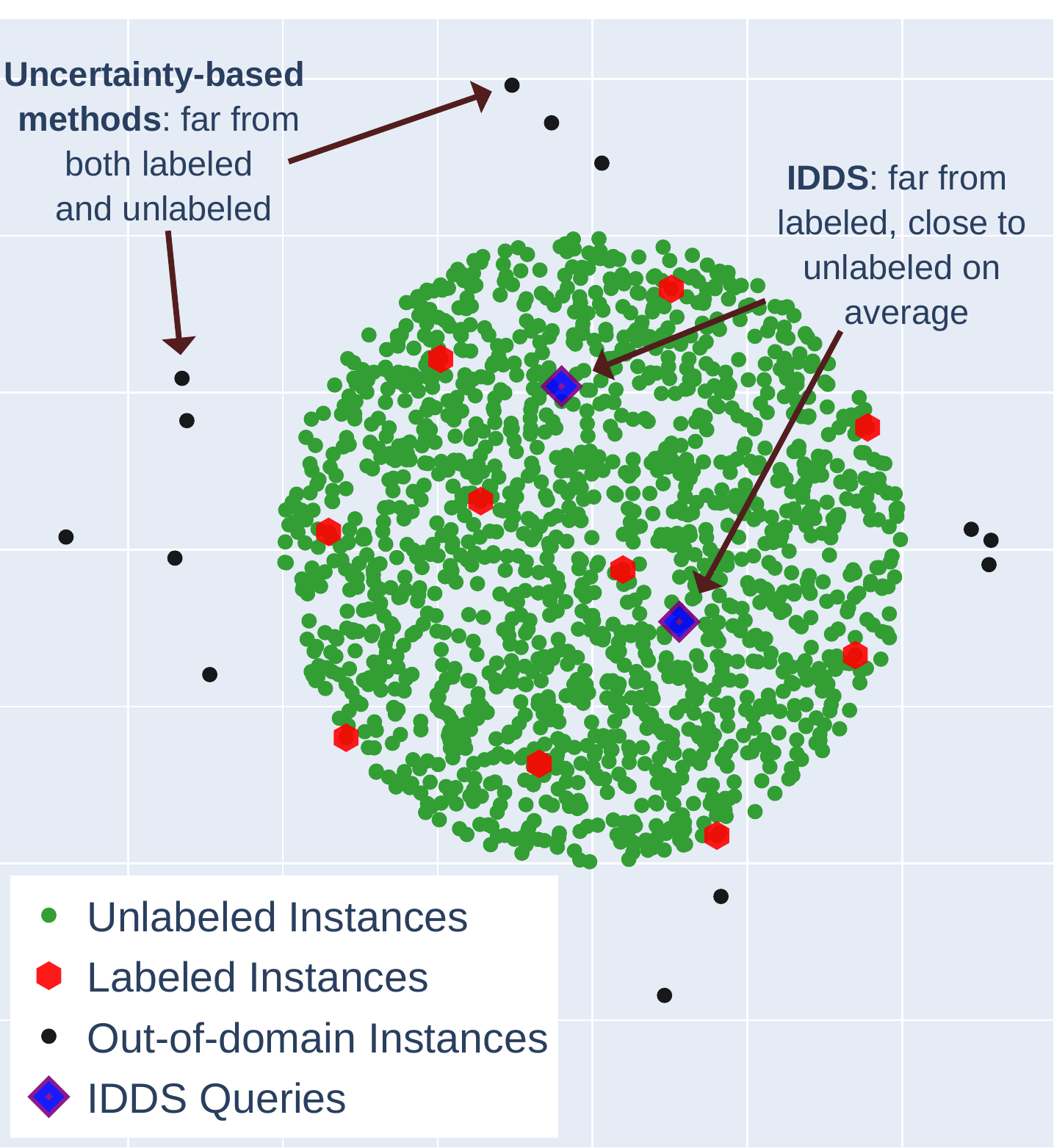}}
	\vspace{-0.7cm}
	\caption{The visualization of the idea behind the IDDS alogrithm on the synthetic data: select instances located far from labeled data while close on average to unlabeled data.}
	\label{fig:idds}
	\vspace{-0.3cm}
	
\end{figure}

%% file: figures/uncertainty_rouge1.tex
\begin{figure*}[h!t]
    \footnotesize
    
    \centering
    \begin{minipage}[h]{0.32\linewidth}
    \center{\includegraphics[width=1\linewidth]{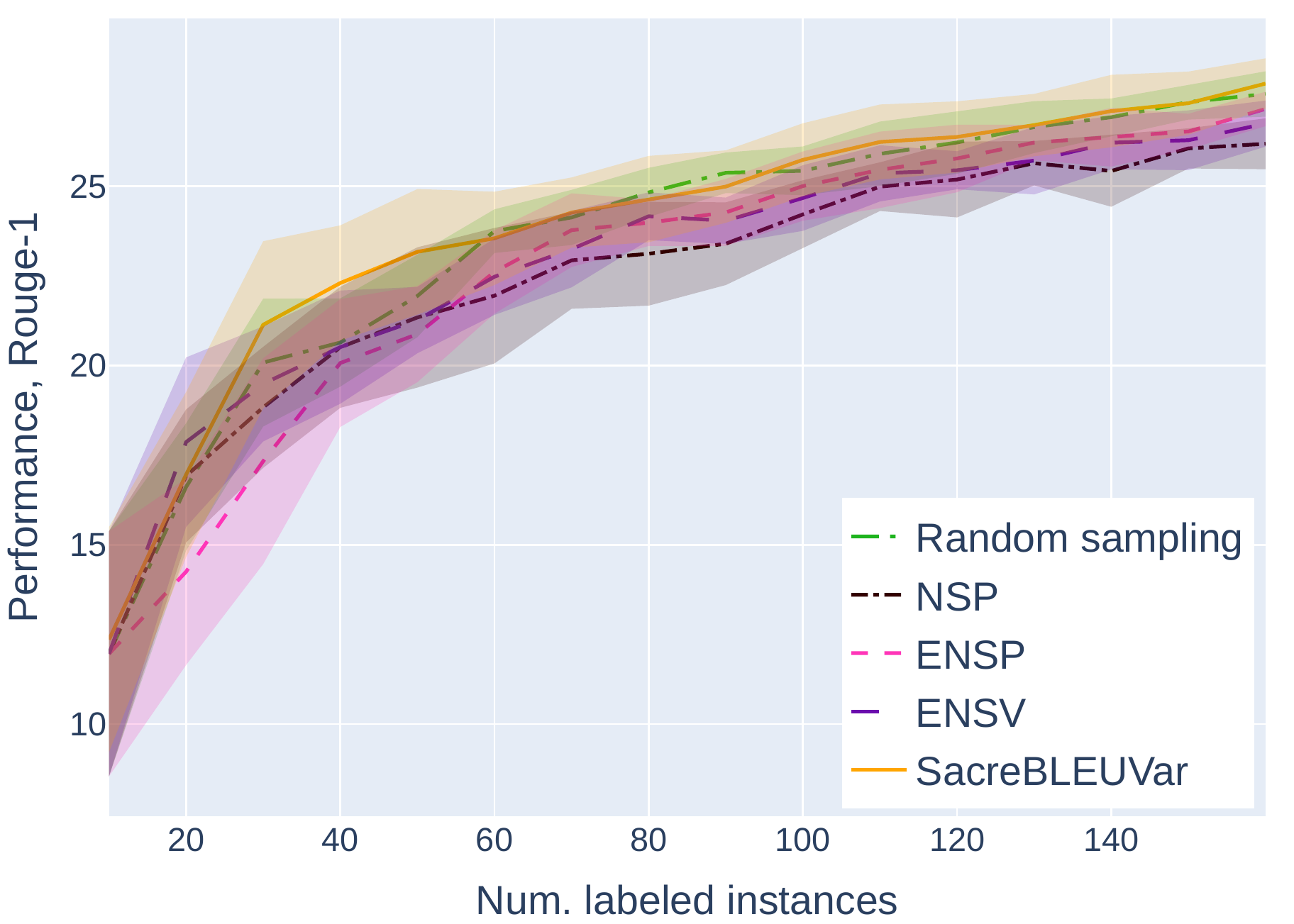} a) AESLC dataset}
    \end{minipage}
    \hspace{0.1cm}
    \centering
    \begin{minipage}[h]{0.32\linewidth}
    \center{\includegraphics[width=1\linewidth]{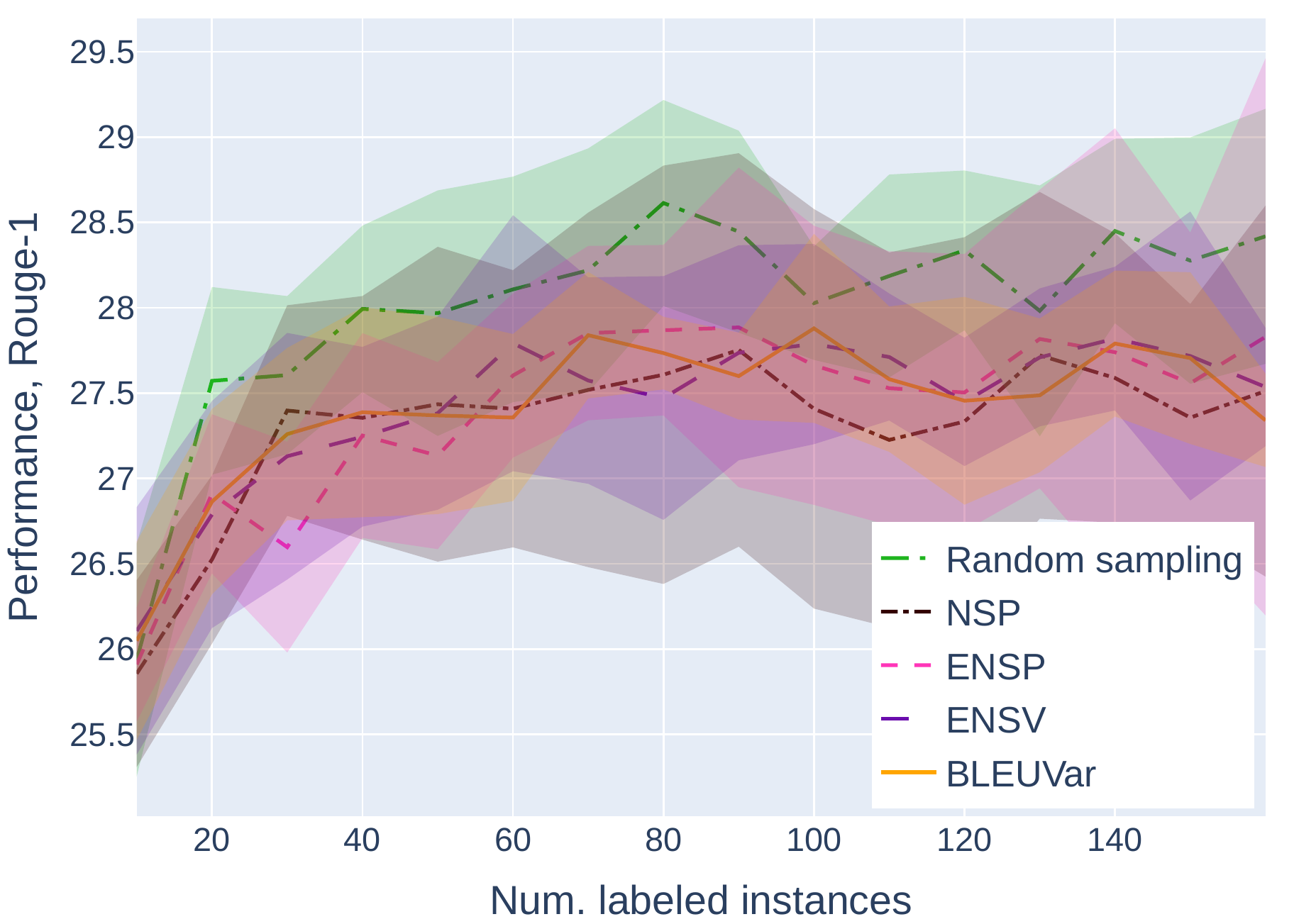} b) WikiHow dataset}
    \end{minipage}
    \hspace{0.1cm}
    \begin{minipage}[h]{0.32\linewidth}
    \center{\includegraphics[width=1\linewidth]{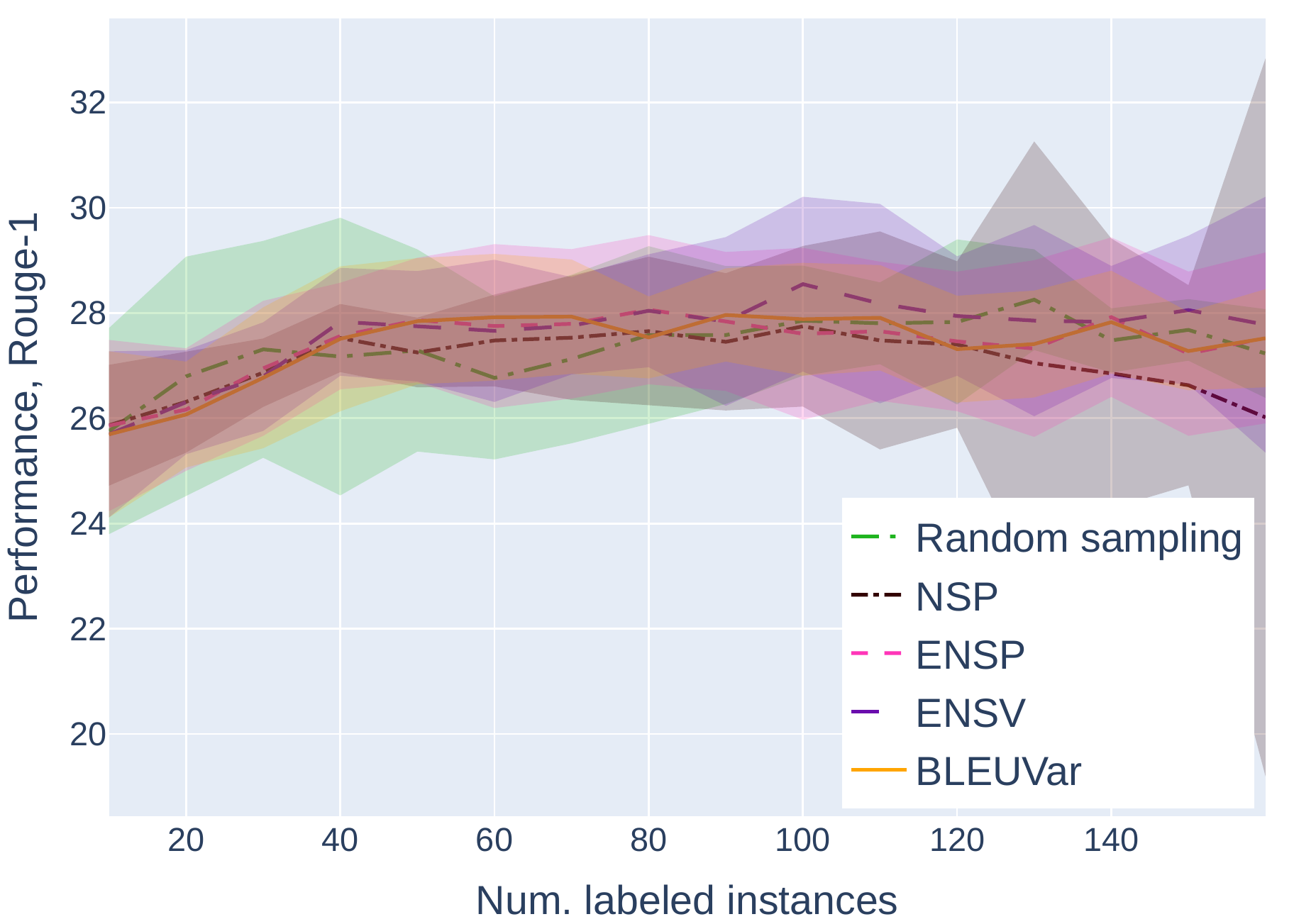} c) PubMed dataset}
    \end{minipage}
    
    \caption{ROUGE-1 scores of BART-base with various uncertainty-based strategies compared with random sampling (baseline) on various datasets. Full results are provided in Figures~\ref{fig:aeslc_uncertainty},~\ref{fig:wikiall_uncertainty},~\ref{fig:pubmed_uncertainty_bart}, respectively.}
    \label{fig:unc_rouge1}
    \vspace{-0.3cm}
\end{figure*}

%% file: figures/idds_rouge1.tex
\begin{figure*}[!ht]
    \footnotesize
    
    \centering
    \begin{minipage}[h]{0.32\linewidth}
    \center{\includegraphics[width=1\linewidth]{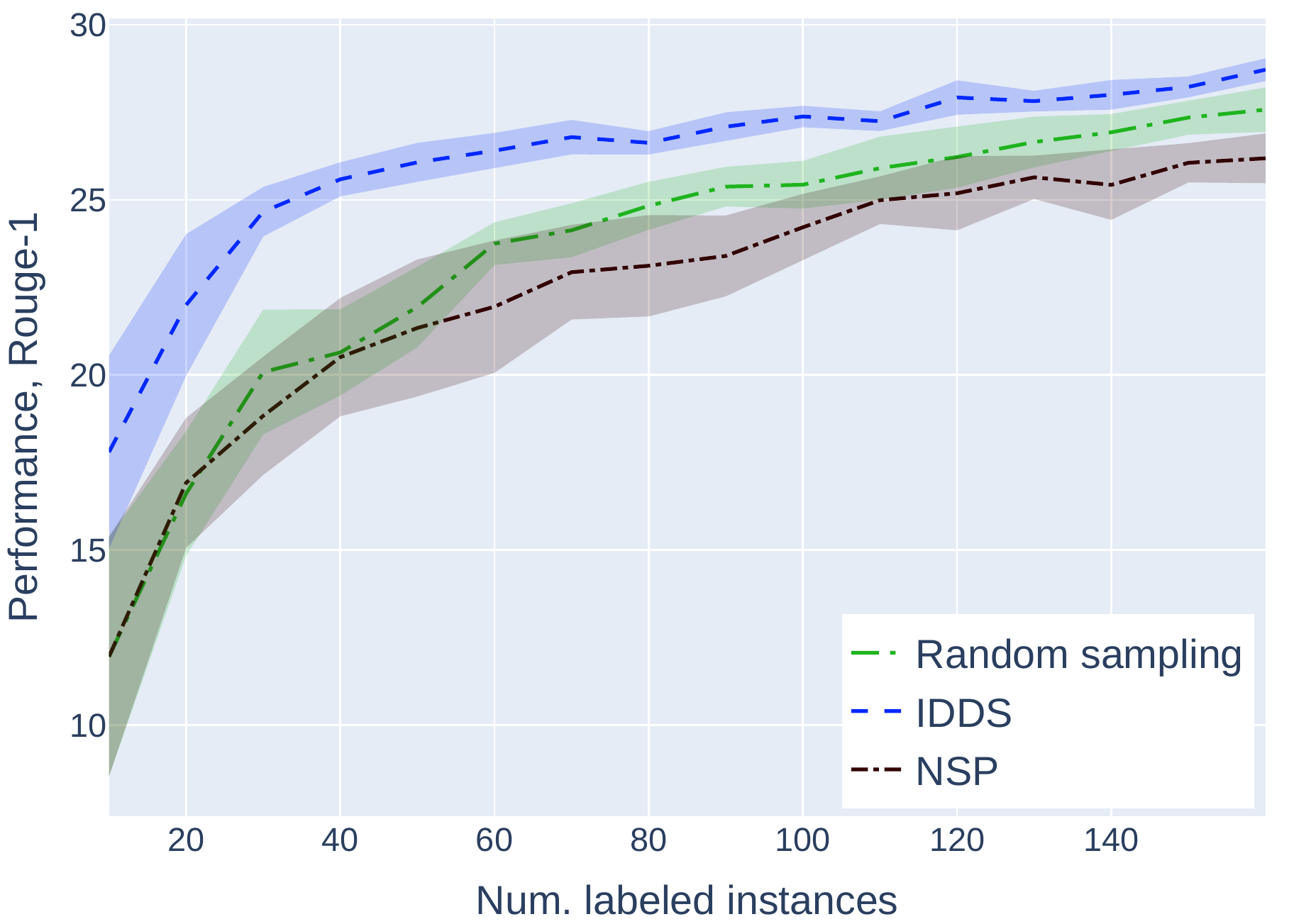} a) AESLC dataset}
    \end{minipage}
    \hspace{0.1cm}
    \centering
    \begin{minipage}[h]{0.32\linewidth}
    \center{\includegraphics[width=1\linewidth]{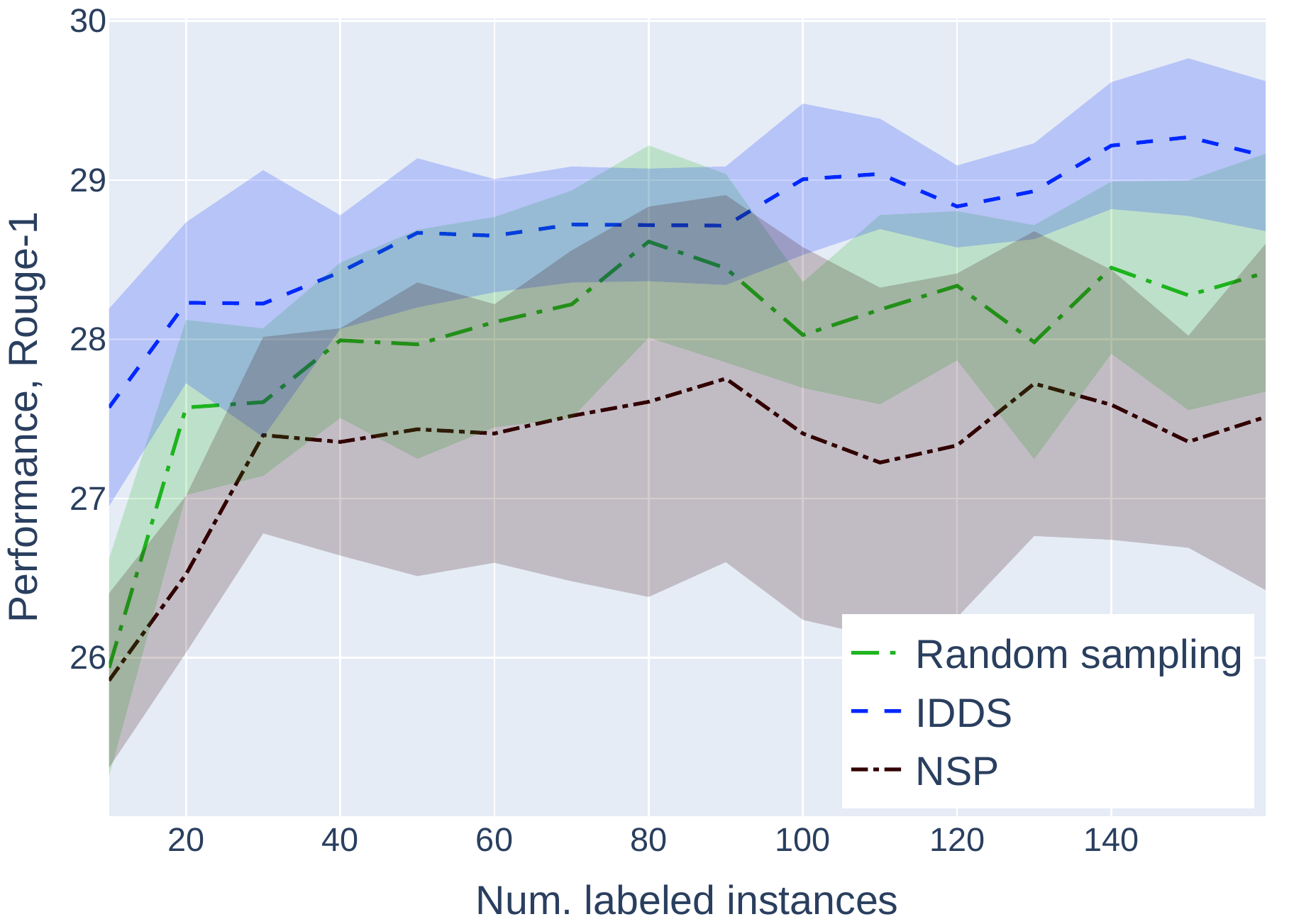} b) WikiHow dataset}
    \end{minipage}
    \hspace{0.1cm}
    \begin{minipage}[h]{0.32\linewidth}
    \center{\includegraphics[width=1\linewidth]{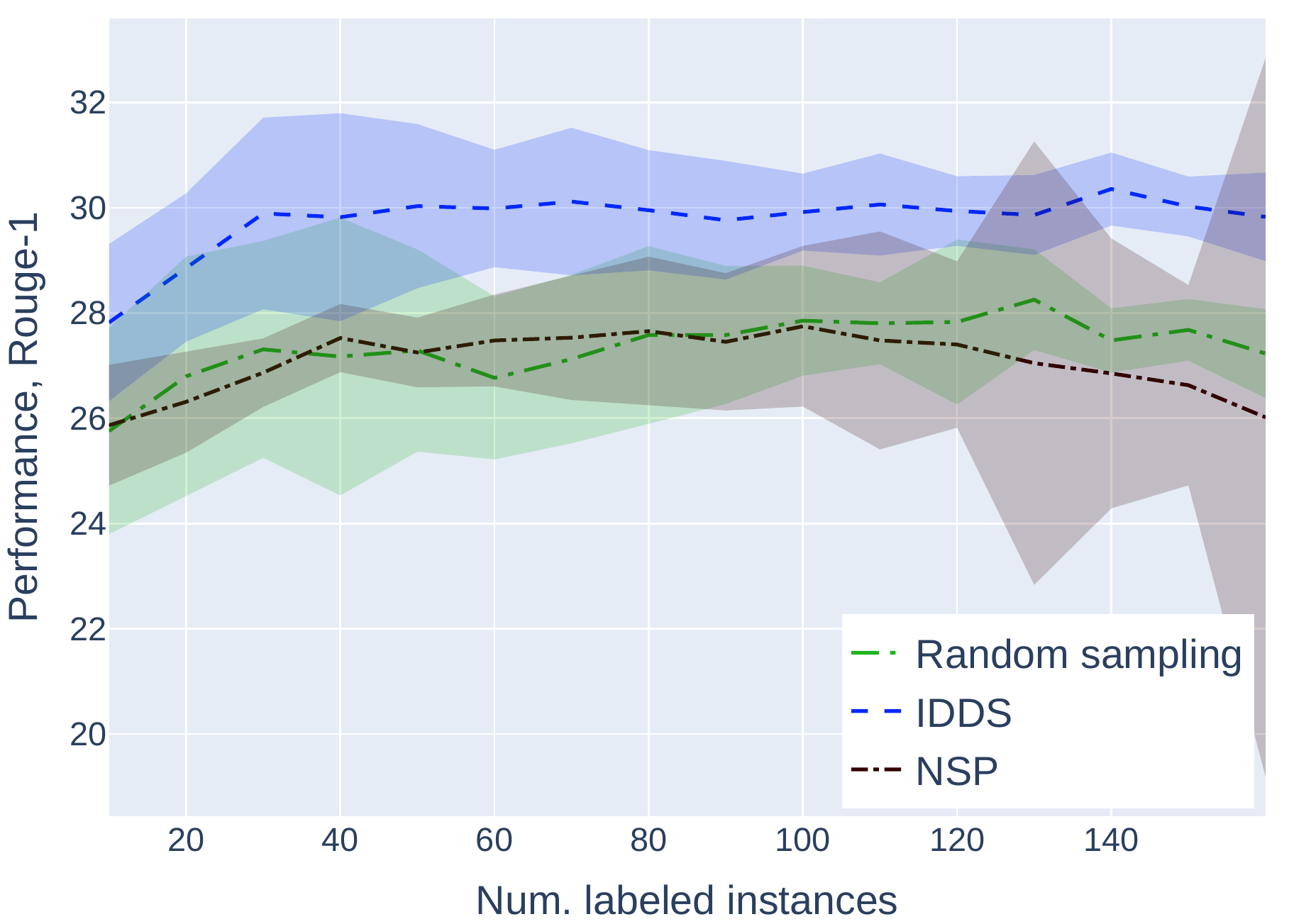} c) PubMed dataset}
    \end{minipage}
    
    \caption{ROUGE-1 scores of BART-base with the IDDS strategy compared with random sampling (baseline) and NSP (uncertainty-based strategy) on various datasets. Full results are provided in Figures~\ref{fig:aeslc_embs}, \ref{fig:wikiall_embs} and~\ref{fig:pubmed_embs}, respectively.}
    \label{fig:idds_rouge1}
    
    \vspace{-0.5cm}
\end{figure*}

%% file: figures/examples.tex
\begin{table*}[ht]
\centering
\footnotesize

\begin{tabular}{|c|l|c|c|}
\hline
\textbf{AL Strat.} & \textbf{Document} & \textbf{Golden Summary} & \textbf{Gener. Summ.} \\ \hline
NSP & \begin{tabular}[c]{@{}l@{}}Aquarius - Horoscope Friday, September 8, 2000 by\\ Astronet.com. Powerful forces are at work to challenge\\ you (...) Don't let hurt feelings prevent you from (...)\end{tabular} & \begin{tabular}[c]{@{}l@{}} \colorred{These things are}\\ \colorred{beginning to} \\ \colorred{scare me...}\end{tabular} & \begin{tabular}[c]{@{}l@{}}Invitation  --\\ Aquarius\end{tabular} \\ \hline
NSP & \begin{tabular}[c]{@{}l@{}}Prod Area and Long Haul k\# \colorblue{Volume}  Rec  Del 3.6746\\ 5000  St 62 (...)  \#6563 PPL (Non NY) should have\\ this contract tomorrow. (...) 3.5318 6500 Leidy  PSE\&G\end{tabular} & \begin{tabular}[c]{@{}l@{}}\colorred{TRCO} \colorblue{capacity} \\ \colorred{for Sep}\end{tabular} & Prod Area \\ \hline
IDDS & \begin{tabular}[c]{@{}l@{}}Greg,  I wanted to forward this letter to you that I received\\ from a \colorblue{good friend of mine} who is interested in discussing\\ (...) \colorgreen{with Enron}. (...) set up a \colorgreen{meeting} (...) Sincerely,\end{tabular} & \begin{tabular}[c]{@{}l@{}}\colorgreen{Meeting with}\\ \colorgreen{Enron} \colorblue{Networks}\end{tabular} & n/a \\ \hline
IDDS & \begin{tabular}[c]{@{}l@{}}Larry,  Could I have the \colorgreen{price} for a 2 day swing \colorgreen{peaker}\\ option at \colorgreen{NGI Chicago}, that can be exercised on any\\ day in \colorblue{February} 2002. Strike is FOM \colorblue{February}, (...)\end{tabular} & \begin{tabular}[c]{@{}l@{}}\colorgreen{Peaker price for}\\ \colorgreen{NGI Chicago}\\ \colorblue{Feb}\end{tabular} & n/a \\ \hline
\end{tabular}

\caption{Examples of instances selected with the NSP and IDDS strategies. Tokens from the source document are highlighted with \colorgreen{green}. Tokens that refer to paraphrasing a part of the document and the corresponding part are highlighted with \colorblue{blue}. Tokens that cannot be derived from the document are highlighted with \colorred{red}.}
\label{tab:examples}
\vspace{-0.4cm}
\end{table*}












%% file: figures/self_supervised_rouge1.tex
\begin{figure*}[!ht]
    \footnotesize
    
    \centering
    \begin{minipage}[ht]{0.32\linewidth}
    \center{\includegraphics[width=1\linewidth]{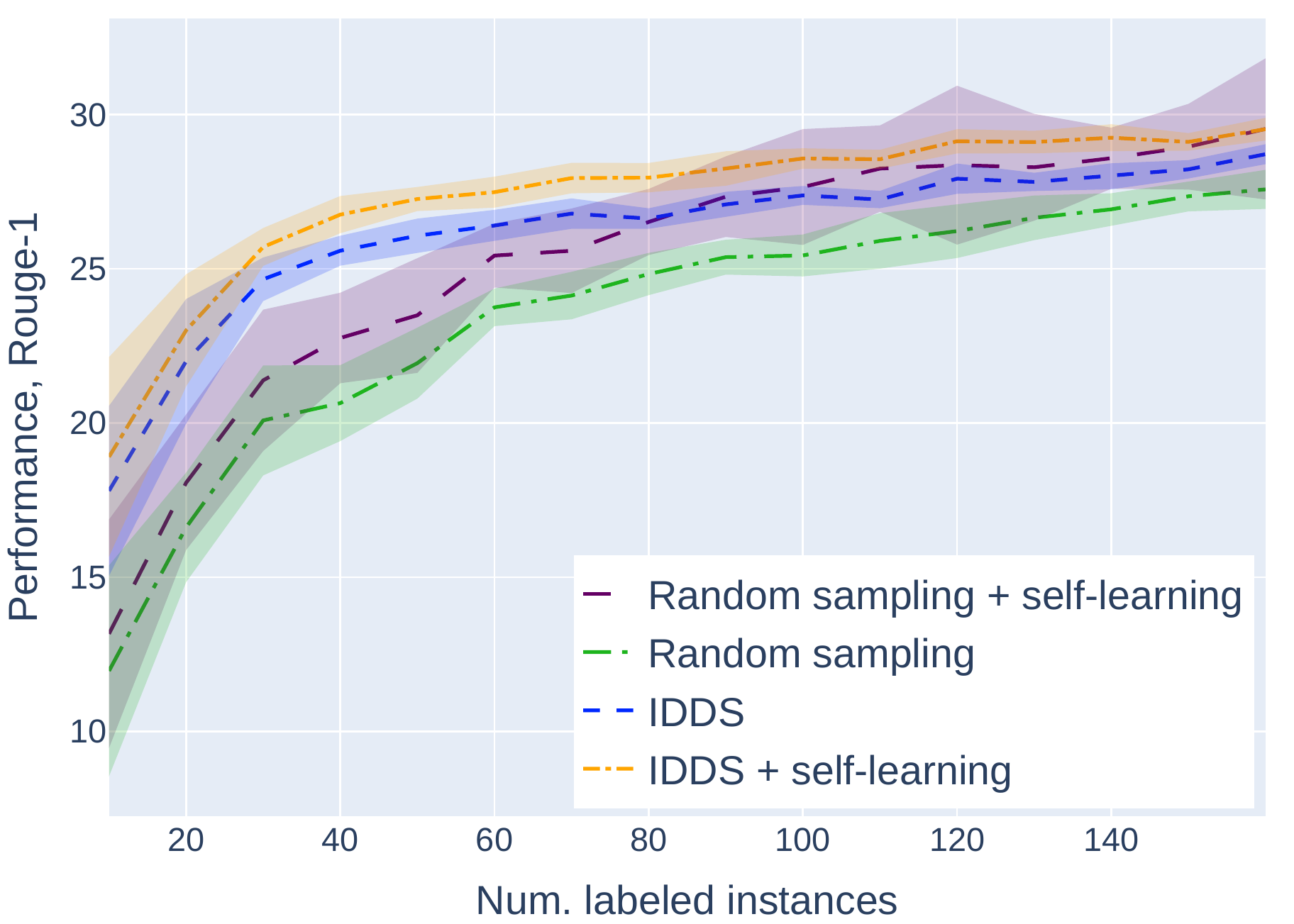} a) AESLC dataset \\ $k_l = 10, k_h = 1$.}
    \end{minipage}
    \hspace{0.1cm}
    \begin{minipage}[ht]{0.32\linewidth}
    \center{\includegraphics[width=1\linewidth]{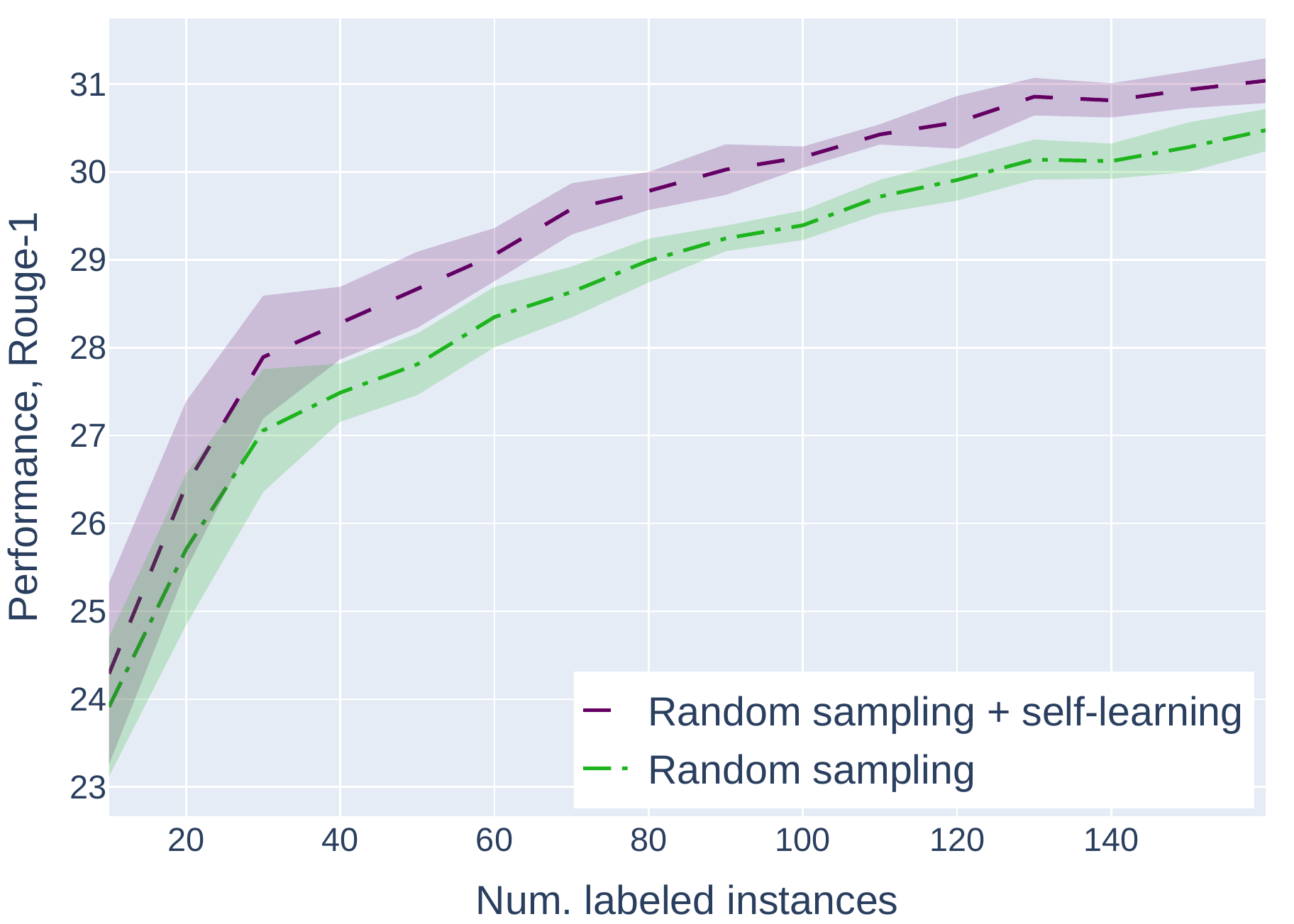} b) Gigaword dataset \\ $k_l = 10, k_h = 1$}
    \end{minipage}
    \hspace{0.1cm}
    \begin{minipage}[ht]{0.32\linewidth}
    \center{\includegraphics[width=1\linewidth]{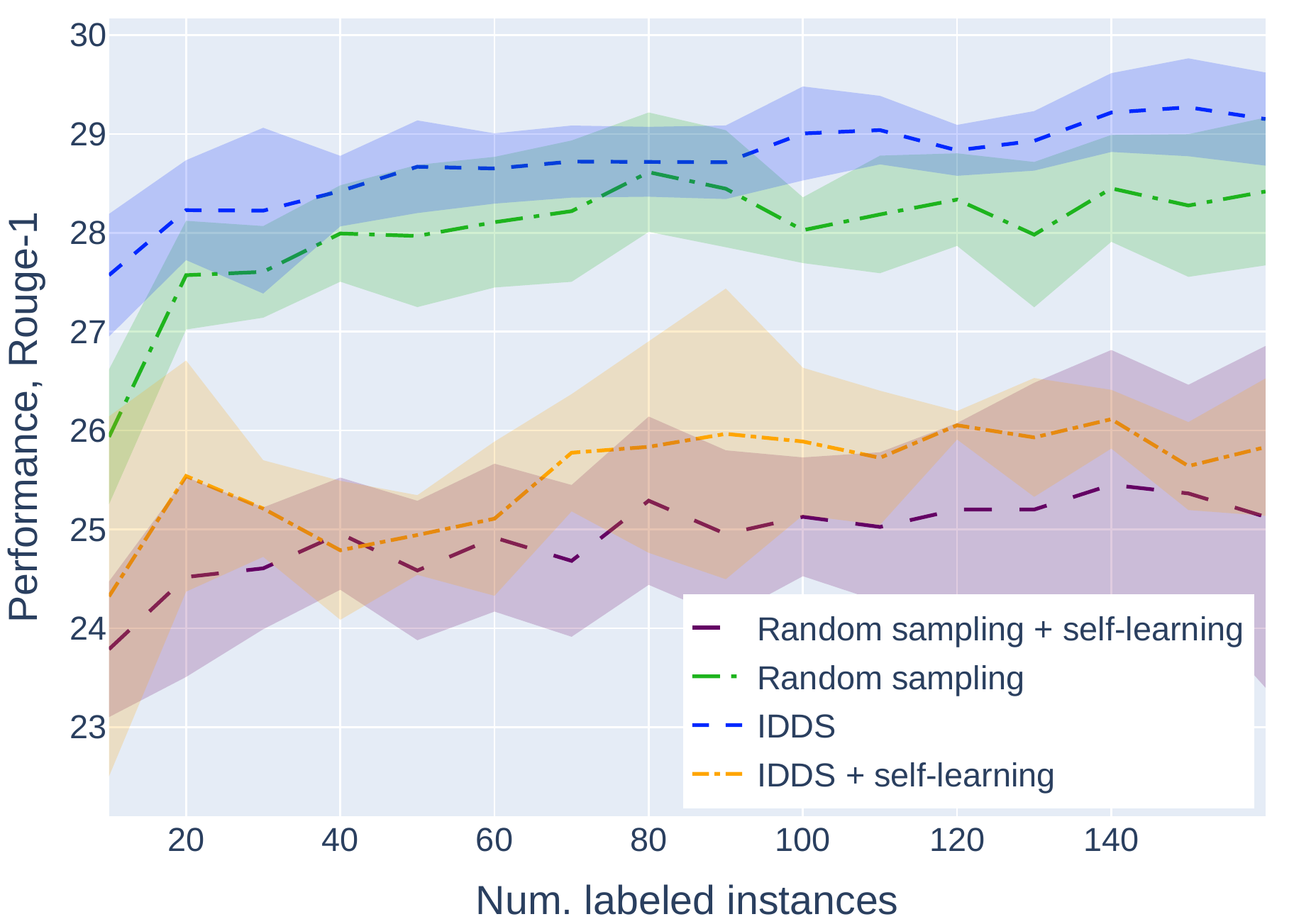} c) WikiHow dataset \\ $k_l = 38, k_h = 2$}
    \end{minipage}
    
    \caption{ROUGE-1 scores of the BART-base model with IDDS and random sampling strategies with and without self-learning on AESLC, Gigaword, and WikiHow. Full results are provided in Figures~\ref{fig:aeslc_pl}, ~\ref{fig:gigaword_pl}, and~\ref{fig:wikihow_pl}, respectively.}
    \label{fig:self_supervised_rouge1}
\end{figure*}

%% file: figures/aeslc_consistency_model_bart.tex
\begin{figure}[ht]
    \vspace{-0.1cm}
	\center{\includegraphics[width=1.\linewidth]{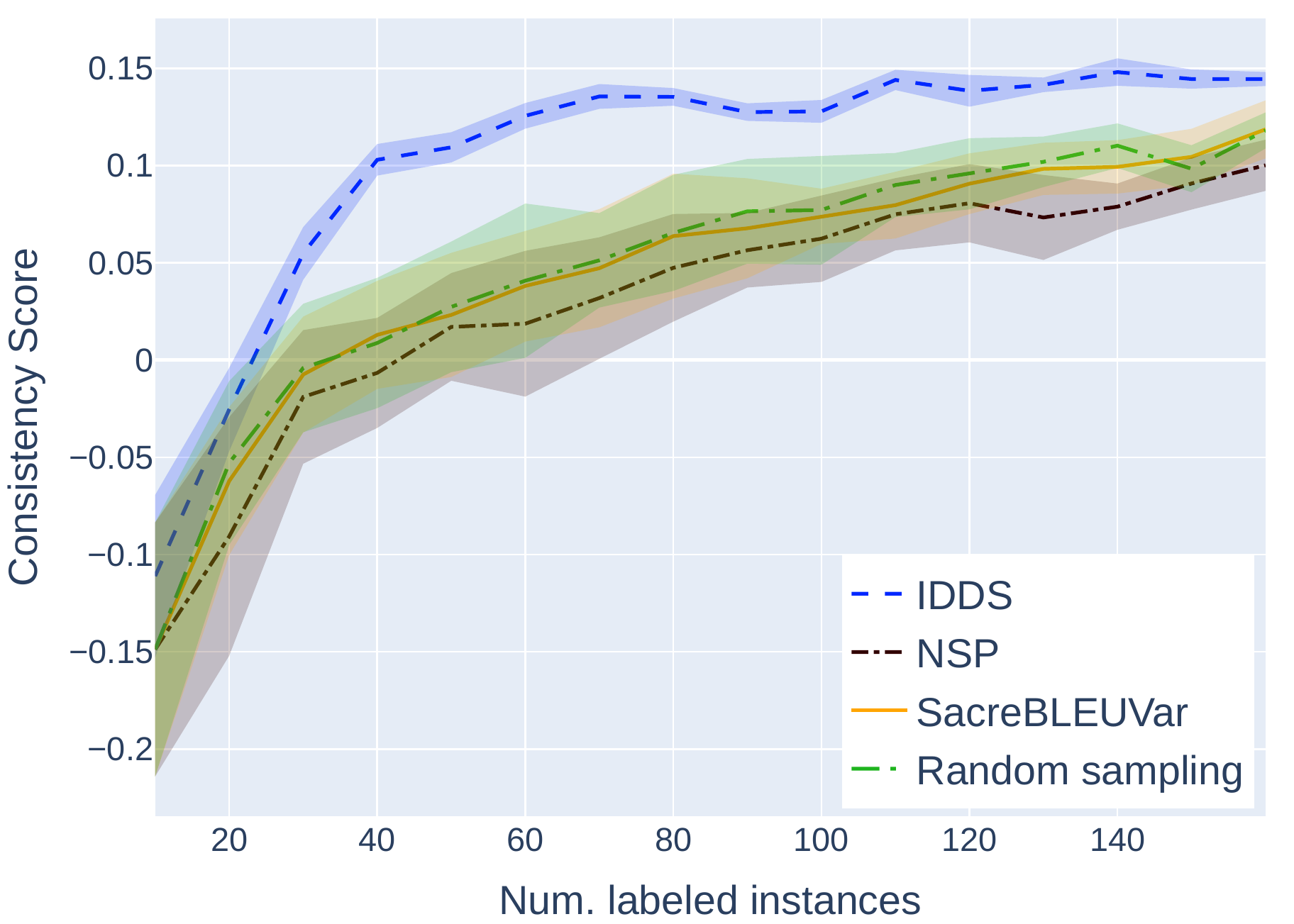}}
	\vspace{-0.3cm}
	\caption{The consistency score calculated via SummaC with BART-base on AESLC with various AL strategies. }
	\label{fig:aeslc_consistency_model_bart}
	
\end{figure}

%% file: figures/datasets_stats.tex
\begin{table}[ht]
\centering
\caption{Dataset statistics. We provide a number of instances for the training and test sets and average lengths of documents / summaries in terms of tokens. All the datasets are English-language. We filter the WikiHow dataset since it contains many noisy instances: we exclude instances with documents that have 10 or less tokens and instances with summaries that have 3 or less tokens.}
\label{tab:datasets_stats}
\begin{tabular}{cccccc}
\textbf{Dataset} & \textbf{Subset} & \textbf{Num. instances} & \textbf{Av. document len.} & \textbf{Av. summary len.} &  \\ \hline
\multirow{2}{*}{AESLC} & Train & 14.4K & 142.4 & 7.8 &  \\
 & Test & 1.9K & 143.8 & 7.9 &  \\ \hline
\multirow{2}{*}{WikiHow} & Train & 184.6K & 377.5 & 77.2 &  \\
 & Test & 1K & 386.9 & 77.0 &  \\ \hline
\multirow{2}{*}{Pubmed} & Train & 119.1K & 495.4 & 263.9 &  \\
 & Test & 6.7K & 509.5 & 268.0 & \\ \hline
\multirow{2}{*}{\begin{tabular}[c]{@{}c@{}}Gigaword \\ (self-learning)\end{tabular}} & Train & 10K & 38.9 & 11.9 &  \\
 & Test & 2K & 37.1 & 12.8 & \\ \hline
\multirow{2}{*}{\begin{tabular}[c]{@{}c@{}}Gigaword \\ (hyperparam. optimiz.)\end{tabular}} & Train & 200 & 40.8 & 13.3 &  \\
 & Test & 2K & 38.6 & 12.5 & \\ \hline
 
\end{tabular}
\end{table}

%% file: figures/hyperparameters.tex
\begin{table}[ht!]
\centering
\caption{Hyperparameter values and checkpoints from the HuggingFace repository~\cite{Wolf2019HuggingFacesTS} of the models. We imitate the low-resource case by randomly selecting 200 instances from Gigaword train dataset as a train sample, and 2,000 instances from the validation set as a test sample for evaluation consistency. For each model, we find the optimal hyperparameters according to evaluation scores on the sampled subset. Generation maximum length is set to the maximum summary length from the available labeled set.
\\ For WikiHow and PubMed datasets, we reduce the batch size and increase gradient accumulation steps by the same amount due to computational bottleneck. \\ Hardware configuration: 2 Intel Xeon Platinum 8168, 2.7 GHz, 24 cores CPU; NVIDIA Tesla v100 GPU, 32 Gb of VRAM.}
\label{tab:hyperparams}

{\begin{tabular}{lll}
\textbf{Hparam} & \textbf{BART} & \textbf{PEGASUS} \\ \hline
Checkpoint & facebook/bart-base   & google/pegasus-large \\
\# Param. & 139M & 570M \\
\hline
Number of epochs & 6 & 4 \\
Batch size & 16 & 2 \\
Gradient accumulation steps & 1 & 8 \\
Min. number of training steps & 350 & 200  \\
Max. sequence length & 1024 & 1024            \\
Optimizer & AdamW & AdamW           \\
Learning rate & 2e-5 & 5e-4 \\
Weight decay & 0.028 & 0.03 \\
Gradient clipping & 0.28 & 0.3 \\
Sheduler & STLR & STLR \\
\% warm-up steps & 10 & 10 \\ 
Num. beams at evaluation & 4 & 4 \\
Generation max. length & Adapt. & Adapt. \\
\hline
\end{tabular}}
\end{table}

%% file: figures/aeslc_uncertainty_bart.tex
    \footnotesize
    \centering
    \begin{minipage}[ht]{0.32\linewidth}
    \vspace{-0.3cm}
    \center{\includegraphics[width=1\linewidth]{figures/aeslc/aeslc_uncertainty_rouge1.pdf} a) ROUGE-1}
    \end{minipage}
    \hspace{0.1cm}
    \begin{minipage}[ht]{0.32\linewidth}
    \center{\includegraphics[width=1\linewidth]{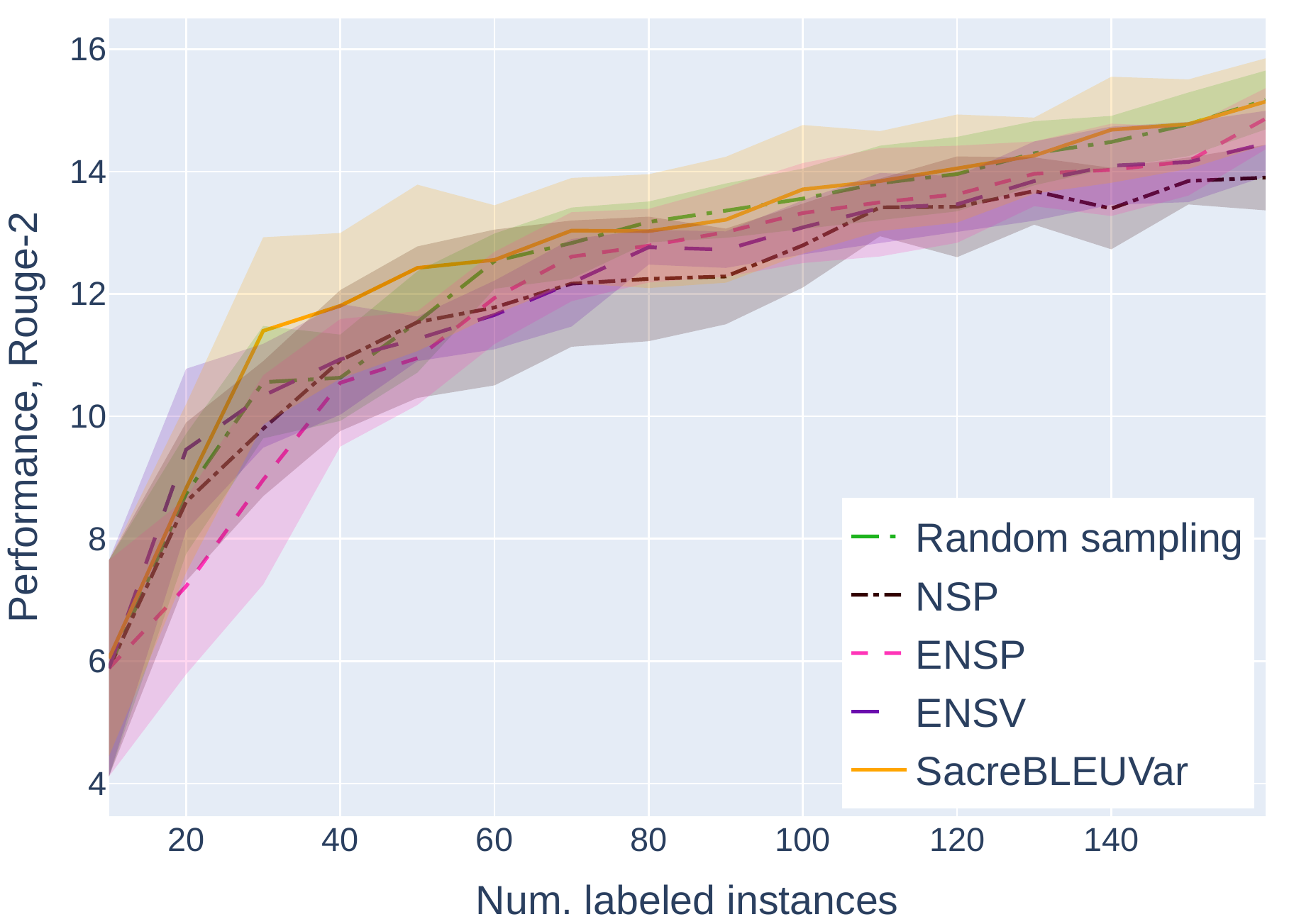} b) ROUGE-2}
    \vspace{0.3cm}
    \end{minipage}
    \hspace{0.1cm}
    \begin{minipage}[ht]{0.32\linewidth}
    \center{\includegraphics[width=1\linewidth]{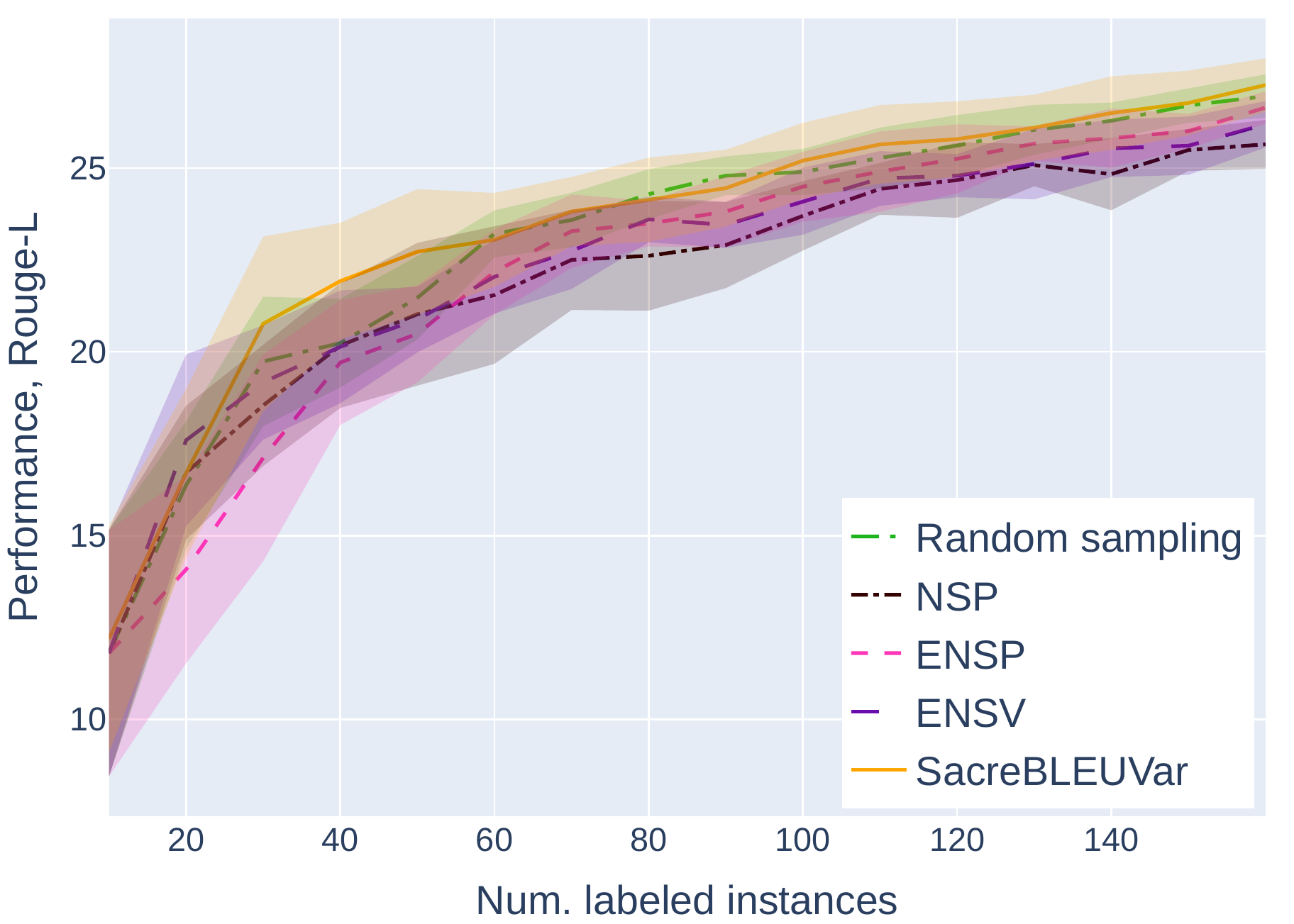} c) ROUGE-L}
    \vspace{0.3cm}
    \end{minipage}
    
    \vspace{-0.4cm}
    \caption{The performance of the BART-base model with various uncertainty-based strategies compared with random sampling (baseline) on AESLC.}
    \label{fig:aeslc_uncertainty}

%% file: figures/aeslc_uncertainty_pegasus.tex
    \footnotesize
    \centering
    \begin{minipage}[ht]{0.32\linewidth}
    \vspace{-0.3cm}
    \center{\includegraphics[width=1\linewidth]{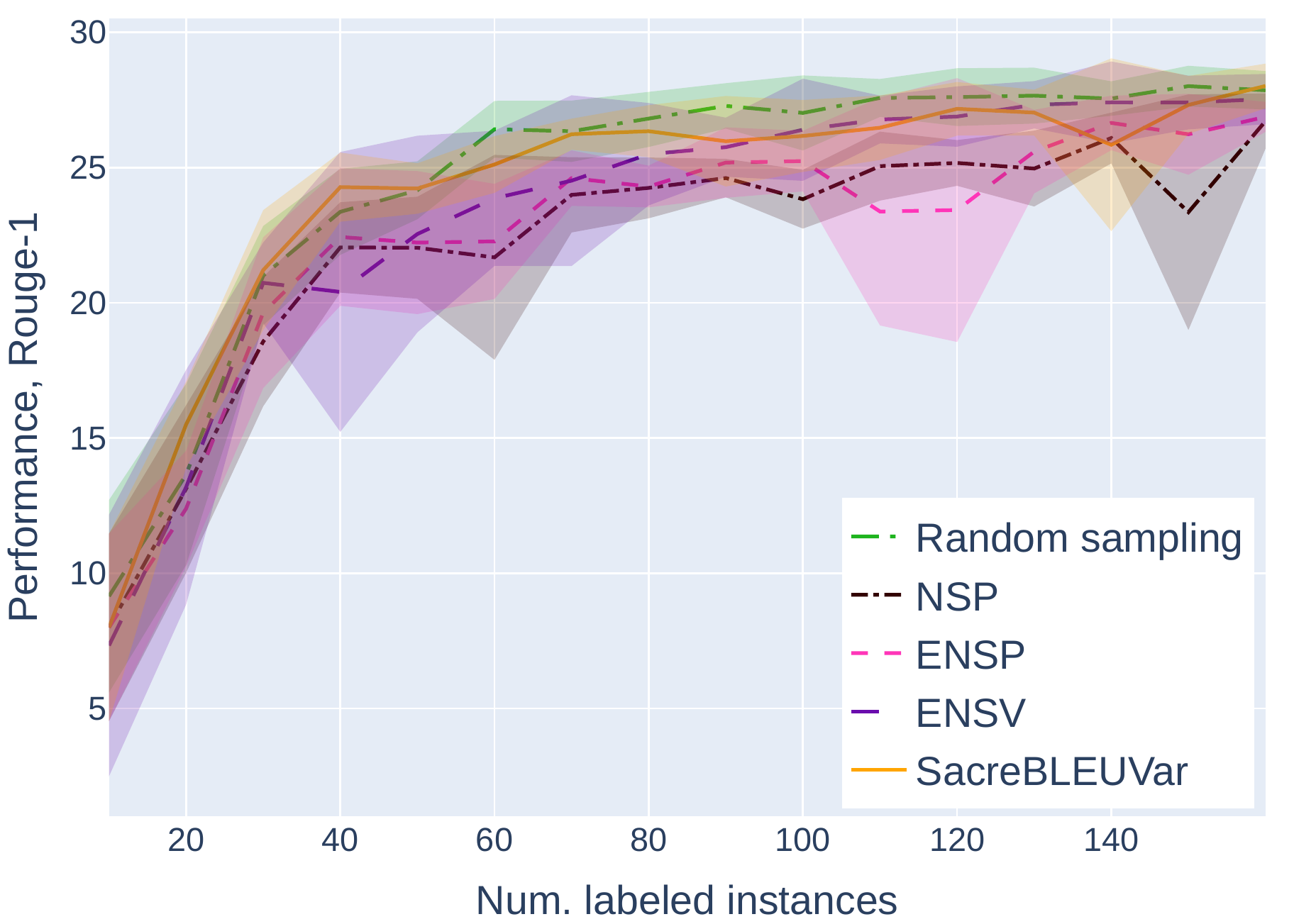} a) ROUGE-1}
    \end{minipage}
    \hspace{0.1cm}
    \begin{minipage}[ht]{0.32\linewidth}
    \center{\includegraphics[width=1\linewidth]{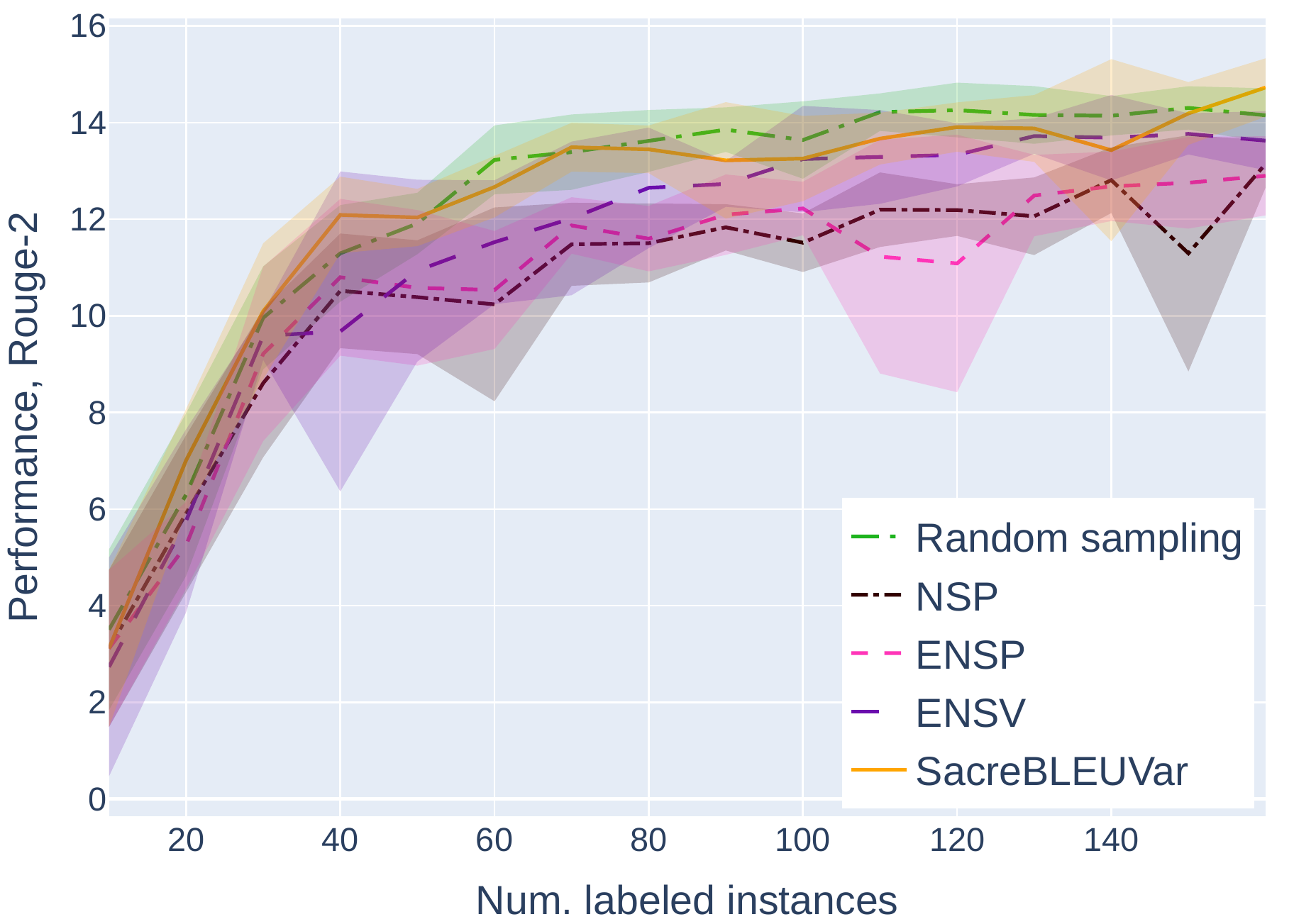} b) ROUGE-2}
    \vspace{0.3cm}
    \end{minipage}
    \hspace{0.1cm}
    \begin{minipage}[ht]{0.32\linewidth}
    \center{\includegraphics[width=1\linewidth]{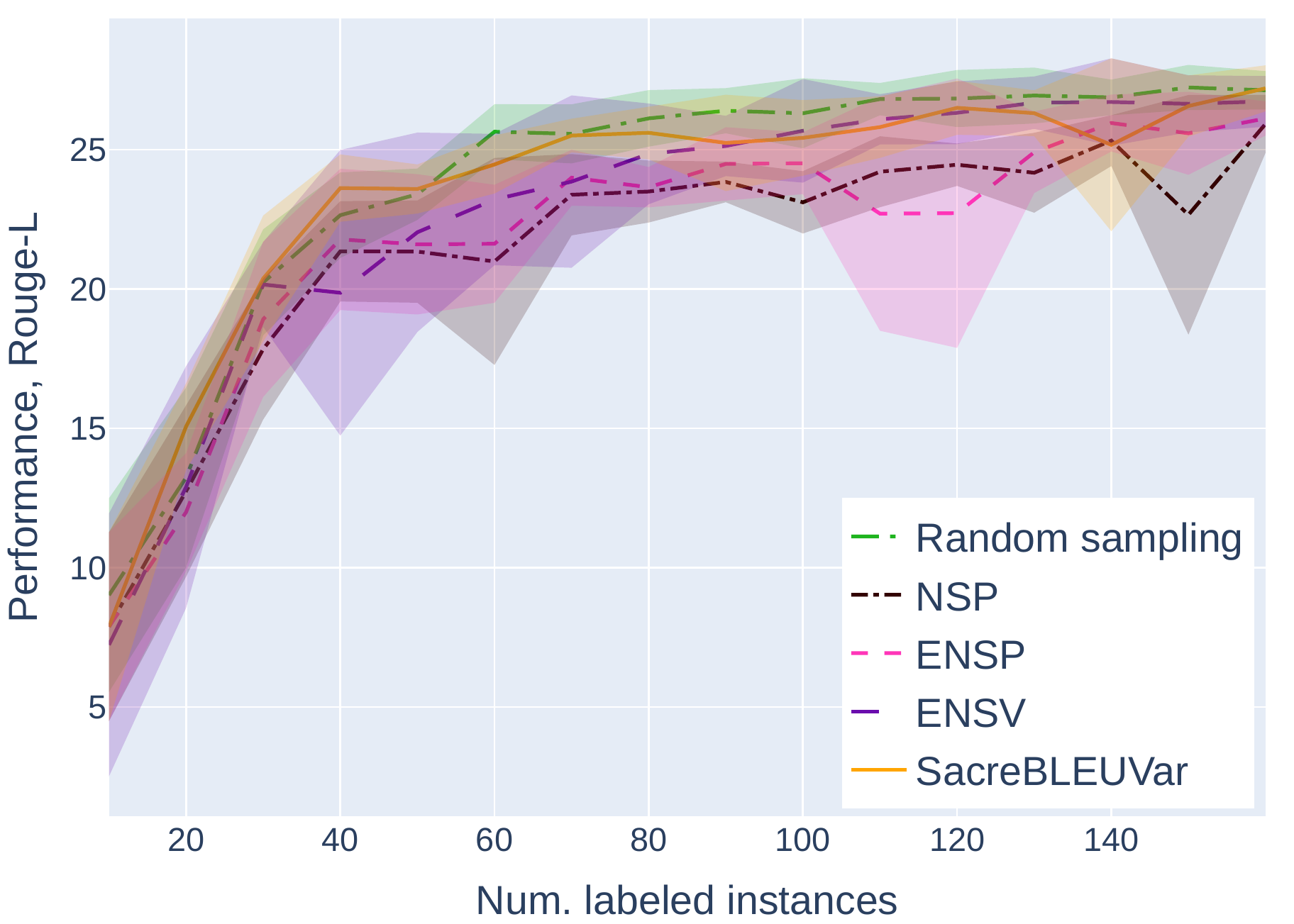} c) ROUGE-L}
    \vspace{0.3cm}
    \end{minipage}
    
    \vspace{-0.4cm}
    \caption{The performance of the PEGASUS-large model with various uncertainty-based strategies compared with random sampling (baseline) on AESLC.}
    \label{fig:aeslc_uncertainty_pegasus}

%% file: figures/wikiall_uncertainty_bart.tex
    \footnotesize
    \centering
    \begin{minipage}[ht]{0.32\linewidth}
    \vspace{-0.3cm}
    \center{\includegraphics[width=1\linewidth]{figures/wikihow_all/wikiall_bart_uncertainty_rouge1.pdf} a) ROUGE-1}
    \end{minipage}
    \hspace{0.1cm}
    \begin{minipage}[ht]{0.32\linewidth}
    \center{\includegraphics[width=1\linewidth]{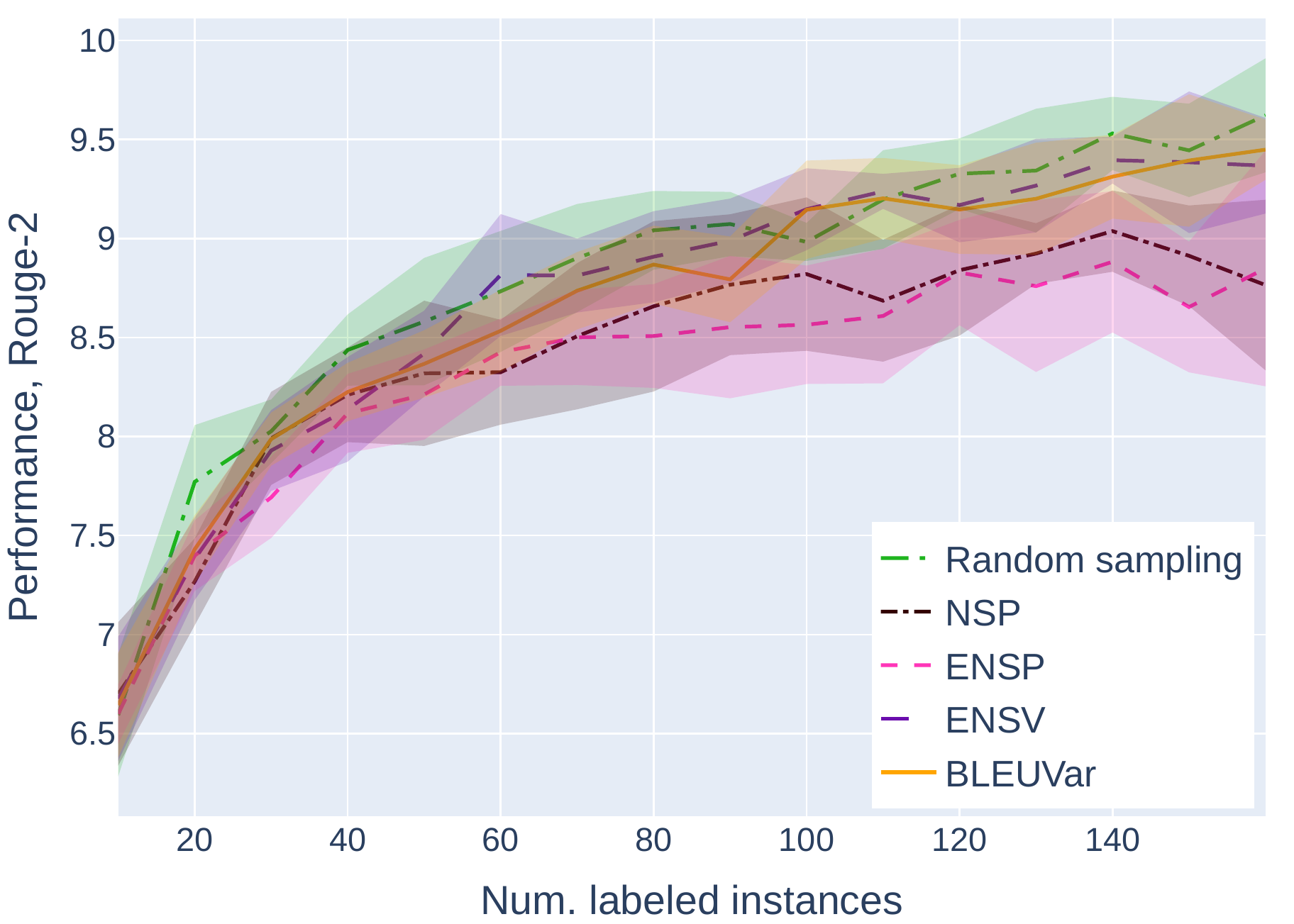} b) ROUGE-2}
    \vspace{0.3cm}
    \end{minipage}
    \hspace{0.1cm}
    \begin{minipage}[ht]{0.32\linewidth}
    \center{\includegraphics[width=1\linewidth]{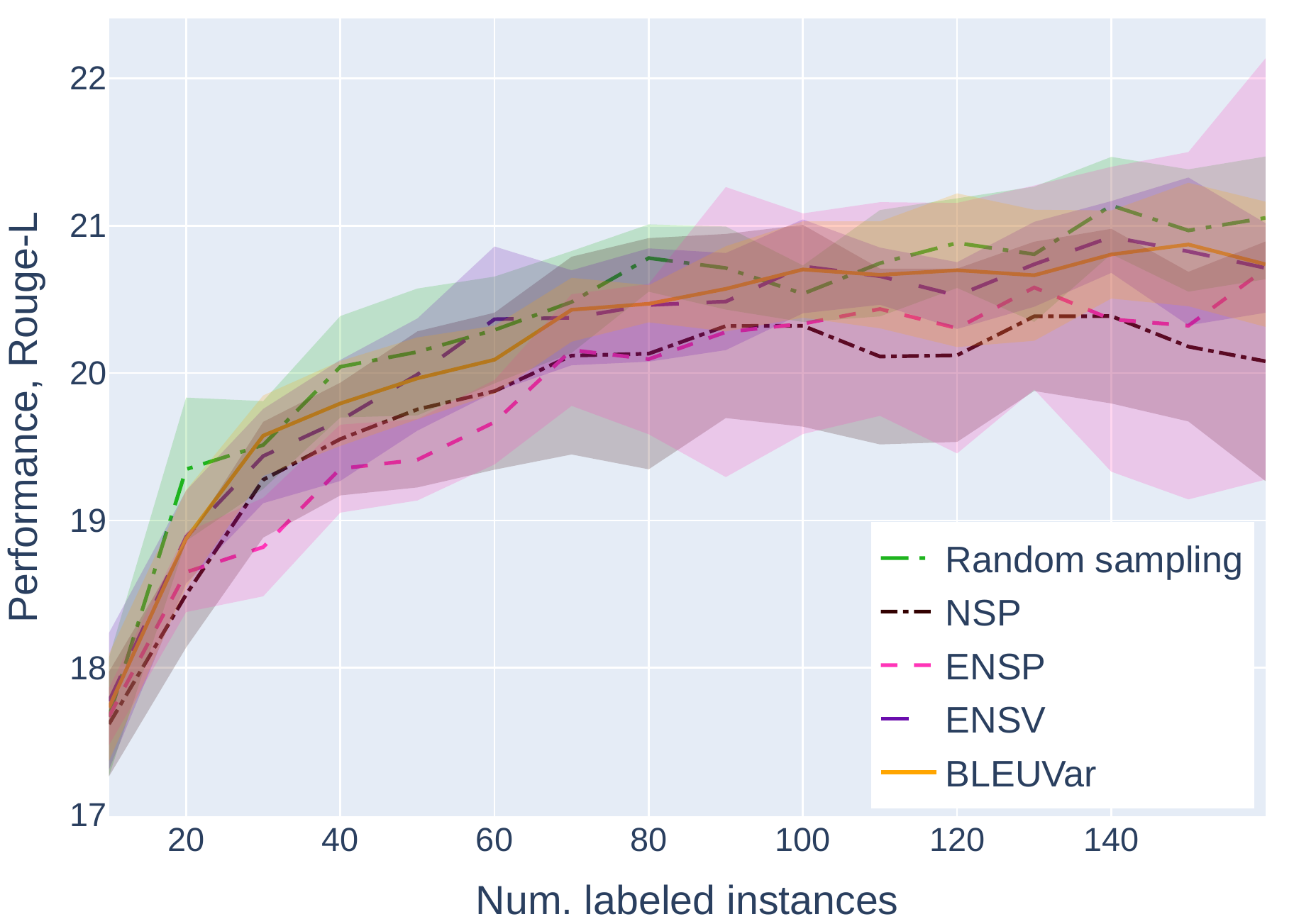} c) ROUGE-L}
    \vspace{0.3cm}
    \end{minipage}
    
    \vspace{-0.4cm}
    \caption{The performance of the BART-base model with various uncertainty-based strategies compared with random sampling (baseline) on WikiHow.}
    \label{fig:wikiall_uncertainty}

%% file: figures/pubmed_uncertainty_bart.tex
    \footnotesize
    \centering
    \begin{minipage}[ht]{0.32\linewidth}
    \vspace{-0.3cm}
    \center{\includegraphics[width=1\linewidth]{figures/pubmed/pubmed_bart_uncertainty_rouge1.pdf} a) ROUGE-1}
    \end{minipage}
    \hspace{0.1cm}
    \begin{minipage}[ht]{0.32\linewidth}
    \center{\includegraphics[width=1\linewidth]{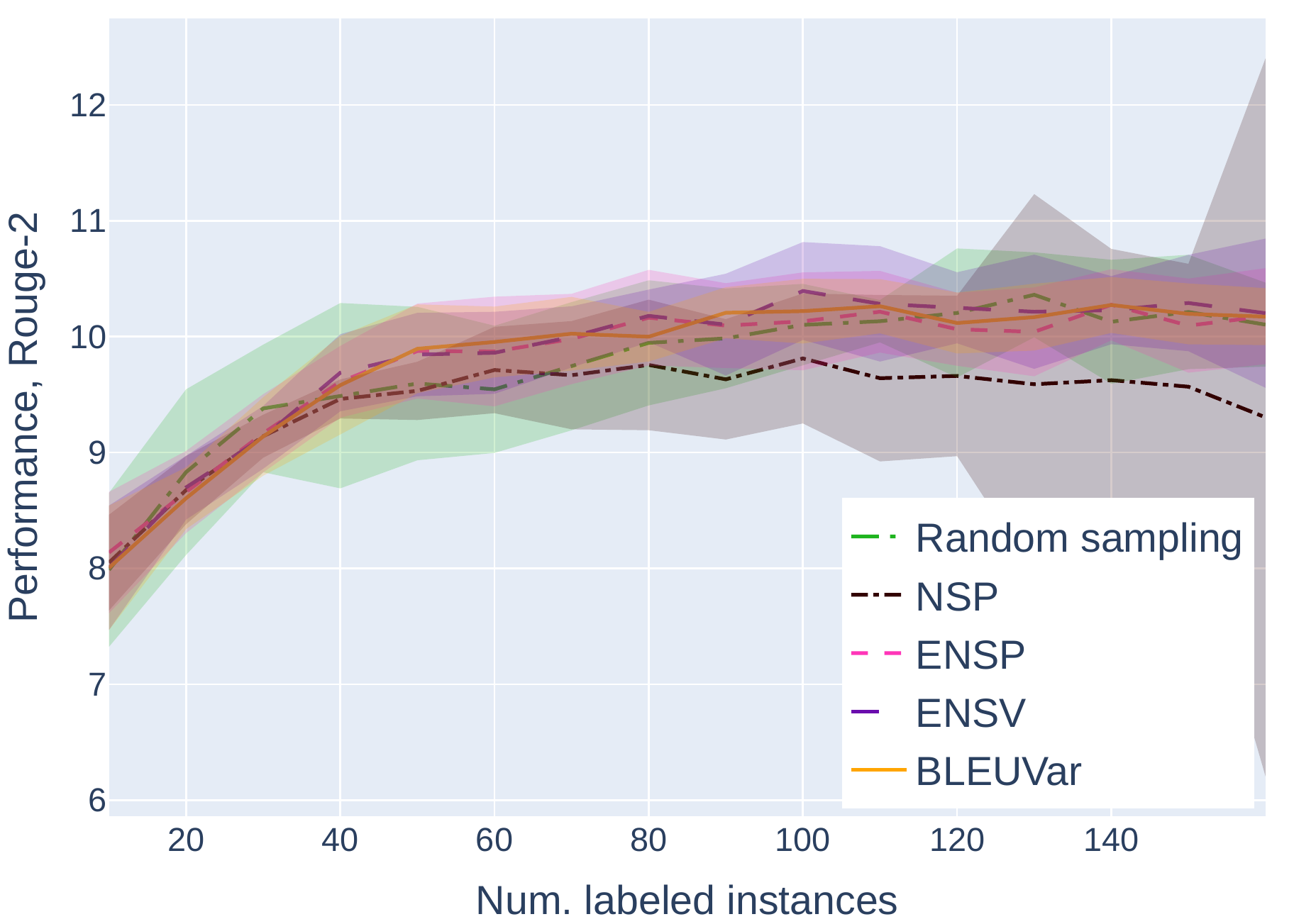} b) ROUGE-2}
    \vspace{0.3cm}
    \end{minipage}
    \hspace{0.1cm}
    \begin{minipage}[ht]{0.32\linewidth}
    \center{\includegraphics[width=1\linewidth]{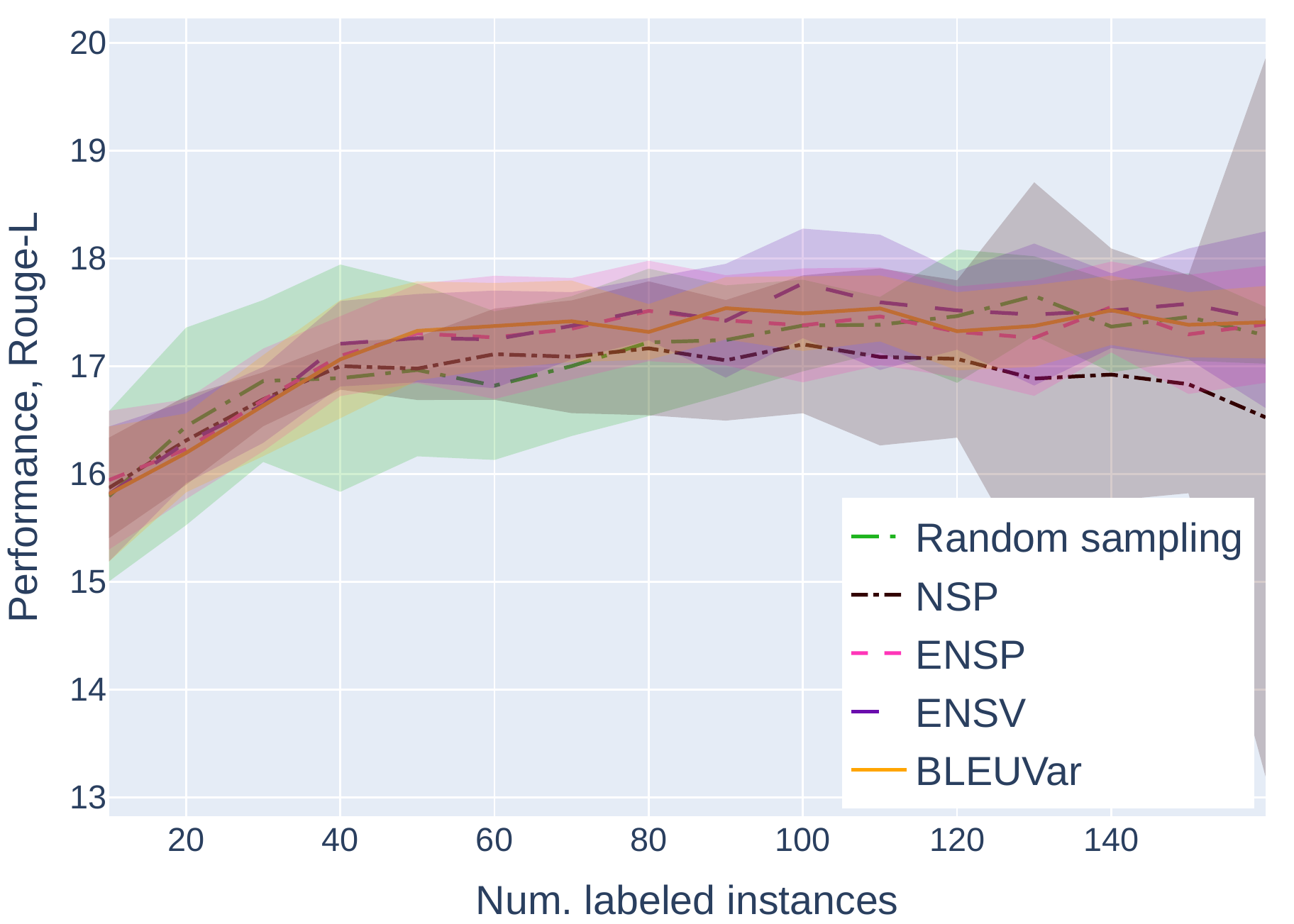} c) ROUGE-L}
    \vspace{0.3cm}
    \end{minipage}
    
    \vspace{-0.4cm}
    \caption{The performance of the BART-base model with various uncertainty-based strategies compared with random sampling (baseline) on PubMed.}
    \label{fig:pubmed_uncertainty_bart}

%% file: figures/increase_idds.tex
\begin{table*}[h]

\centering
\begin{tabular}{ccccc}
\hline
\multicolumn{1}{c} {\textbf{Iter. 0}} & \textbf{Iter. 5} & \textbf{Iter. 10} & \textbf{Iter. 15} & \textbf{Average}
\\
\textbf{R-1/R-2/R-L} & \textbf{R-1/R-2/R-L} & \textbf{R-1/R-2/R-L} & \textbf{R-1/R-2/R-L} &
\textbf{R-1/R-2/R-L}
\\
\hline
\multicolumn{5}{c}{\textbf{AESLC + BART-base}} \\
\hline
48.8 / 52.5 / 48.4 & 11.2 / 14.9 / 11.4 & 5.2 / 5.4 / 5.0 & 4.1 / 2.6 / 3.8 & 10.2 / 11.9 / 10.0 \\
\hline
\multicolumn{5}{c}{\textbf{AESLC + PEGASUS-large}}
\\
\hline
-24.8 / -19.7 / -24.5 & 6.9 / 7.3 / 7.4 &  1.6 / 0.4 / 2.0 &  4.8 / 3.5 / 4.7 &  7.6 / 6.7 / 8.0  \\
\hline
\multicolumn{5}{c}{\textbf{WikiHow + BART-base}} \\
\hline
6.3 / 12.5 / 5.4 & 1.9 / 2.7 / 1.3 & 3.0 / 4.2 / 2.5 & 2.6 / 2.9 / 1.8 & 2.3 / 3.2 / 1.5 
\\
\hline
\multicolumn{5}{c}{\textbf{PubMed + BART-base}}
\\
\hline
8.0 / 10.4 / 5.8 & 12.0 / 11.7 / 8.0 &  8.1 / 6.4 / 4.9 & 9.5 / 6.7 / 5.1 &  8.9 / 7.7 / 5.5 \\
\hline
\end{tabular}

\caption{Percentage increase in ROUGE F-scores of IDDS over the baseline on different AL iterations. \textbf{Average} refers to the average increase throughout the whole AL cycle.}
\label{tab:increase_idds}

\end{table*}





%% file: figures/aeslc_embs_bart.tex
\begin{figure*}[h]
    \footnotesize
    \centering
    \begin{minipage}[ht]{0.32\linewidth}
    \vspace{-0.3cm}
    \center{\includegraphics[width=1\linewidth]{figures/aeslc/aeslc_embs_rouge1.pdf} a) ROUGE-1}
    \end{minipage}
    \hspace{0.1cm}
    \begin{minipage}[ht]{0.32\linewidth}
    \center{\includegraphics[width=1\linewidth]{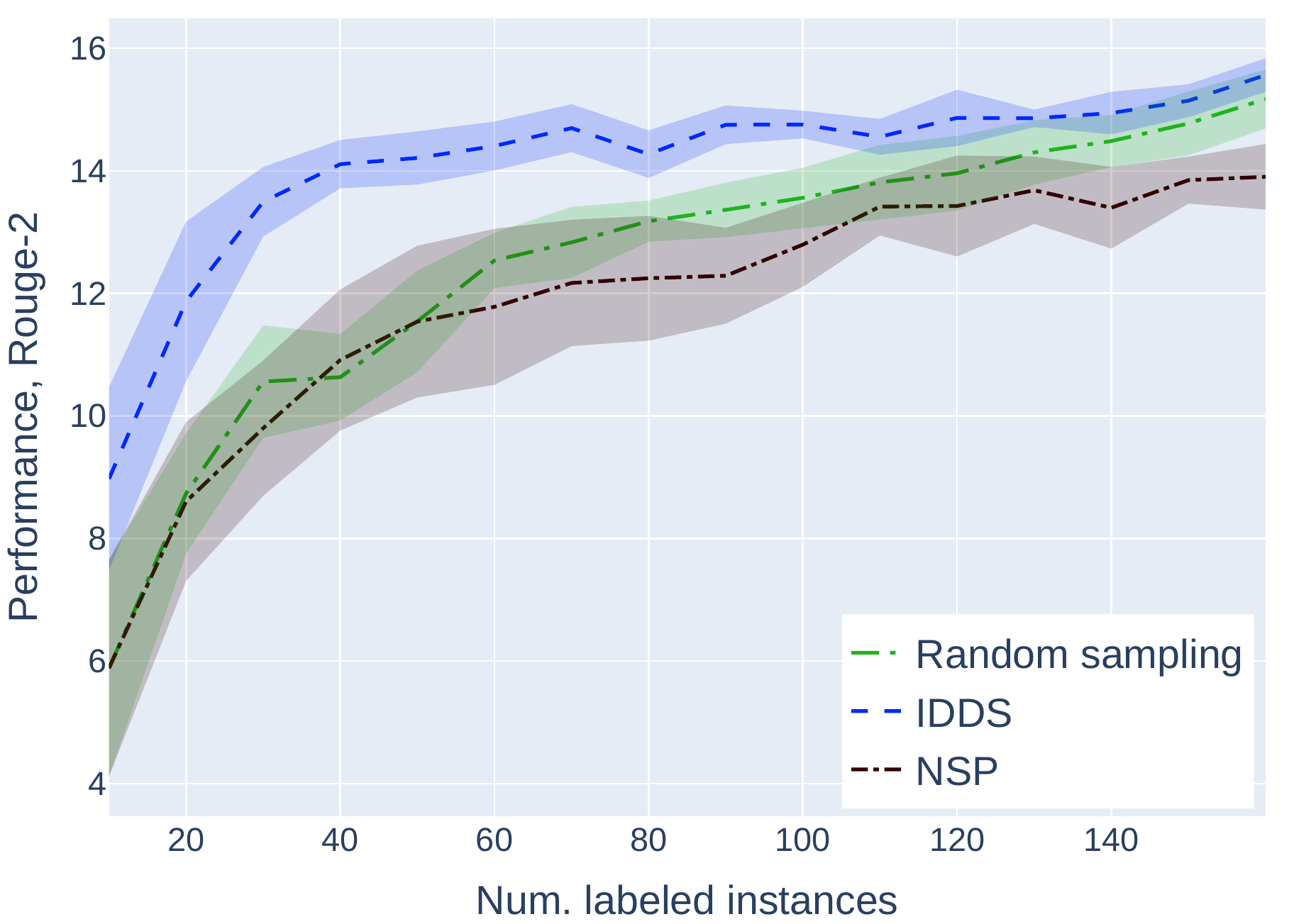} b) ROUGE-2}
    \vspace{0.3cm}
    \end{minipage}
    \hspace{0.1cm}
    \begin{minipage}[ht]{0.32\linewidth}
    \center{\includegraphics[width=1\linewidth]{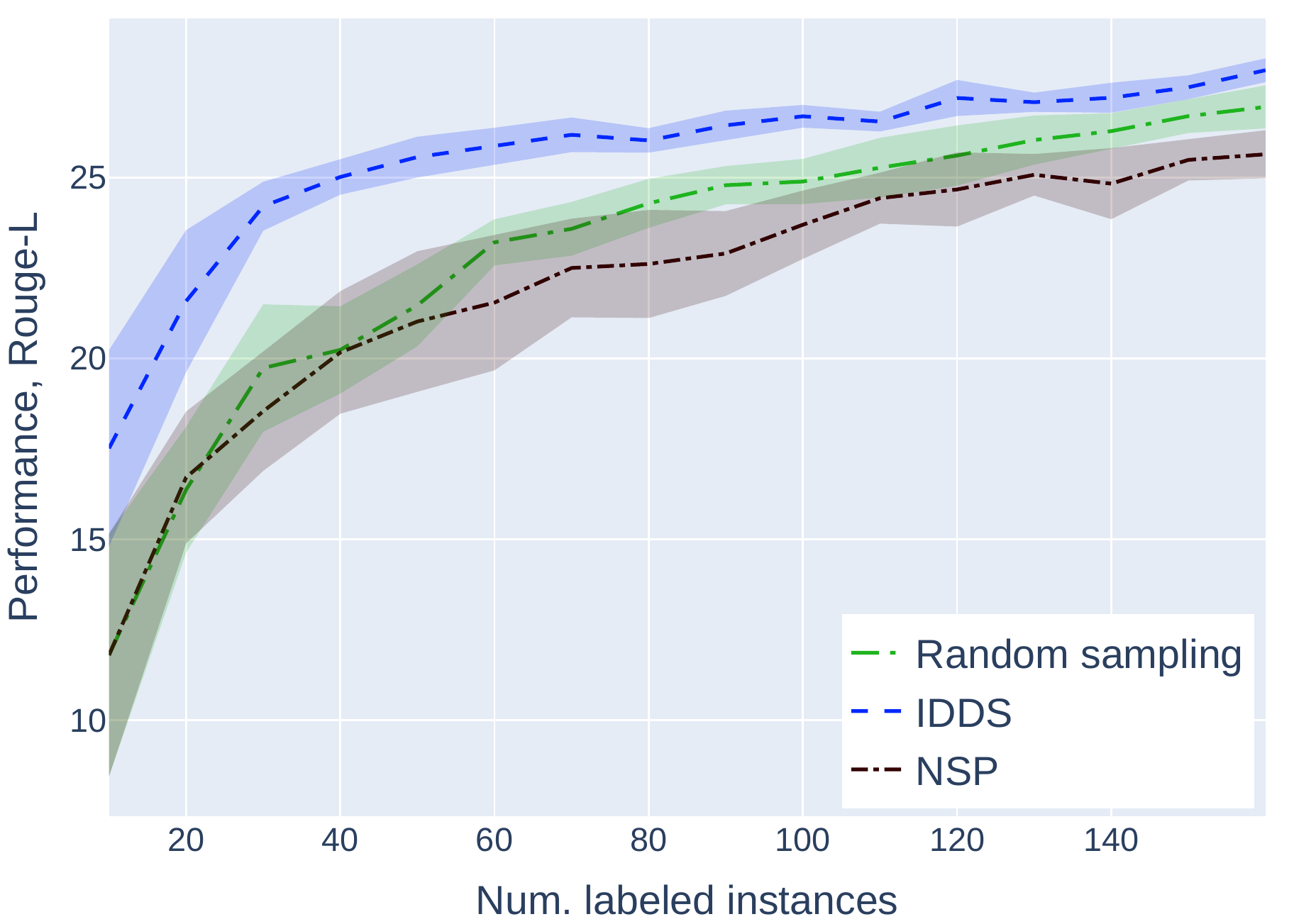} c) ROUGE-L}
    \vspace{0.3cm}
    \end{minipage}
    
    \vspace{-0.2cm}
    \caption{The performance of the BART-base model with the IDDS strategy compared with random sampling (baseline) and NSP (uncertainty-based strategy) on AESLC.}
    \label{fig:aeslc_embs}
\end{figure*}

%% file: figures/aeslc_embs_pegasus.tex
\begin{figure*}[h]
    \footnotesize
    \centering
    \begin{minipage}[ht]{0.32\linewidth}
    \vspace{-0.3cm}
    \center{\includegraphics[width=1\linewidth]{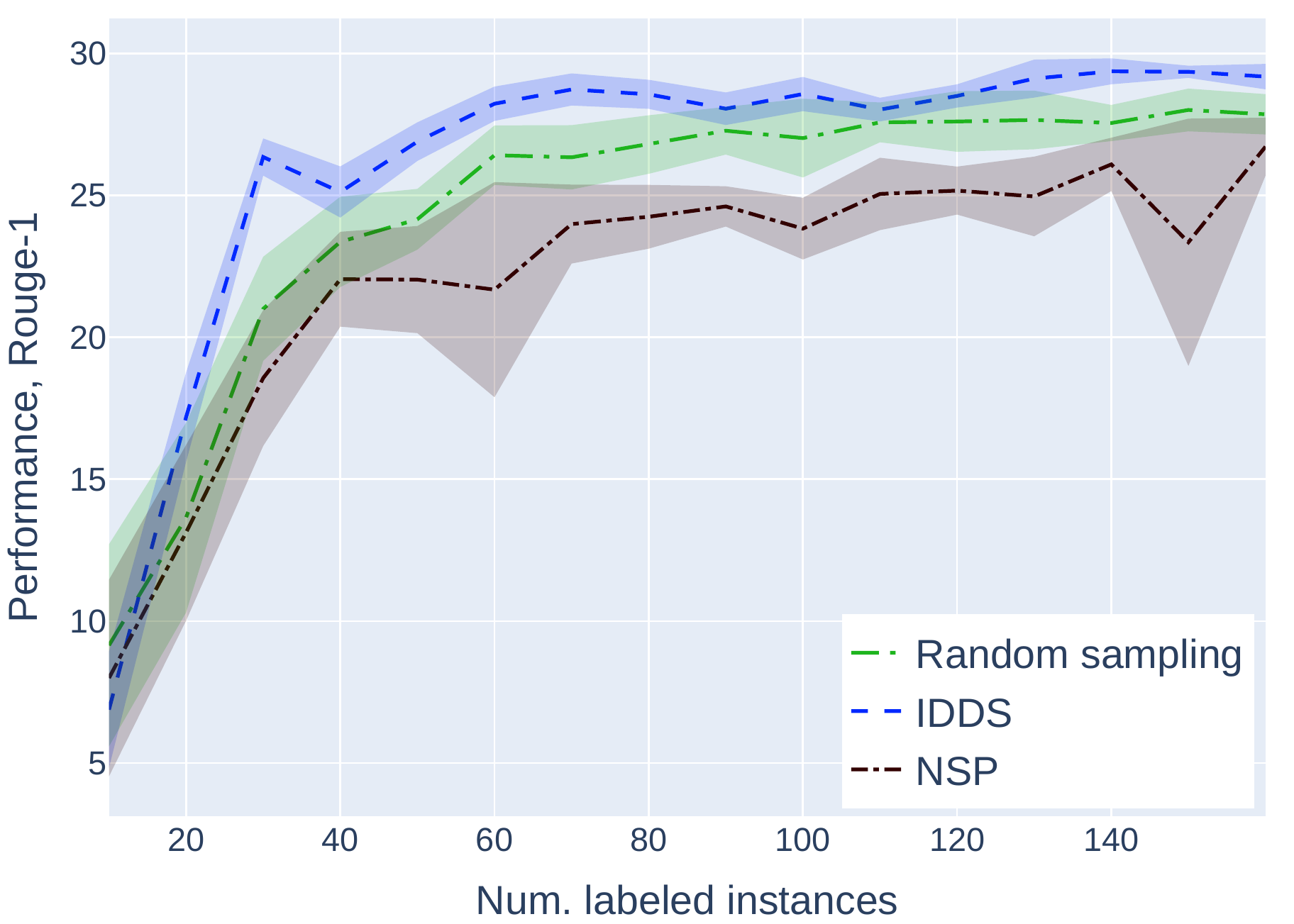} a) ROUGE-1}
    \end{minipage}
    \hspace{0.1cm}
    \begin{minipage}[ht]{0.32\linewidth}
    \center{\includegraphics[width=1\linewidth]{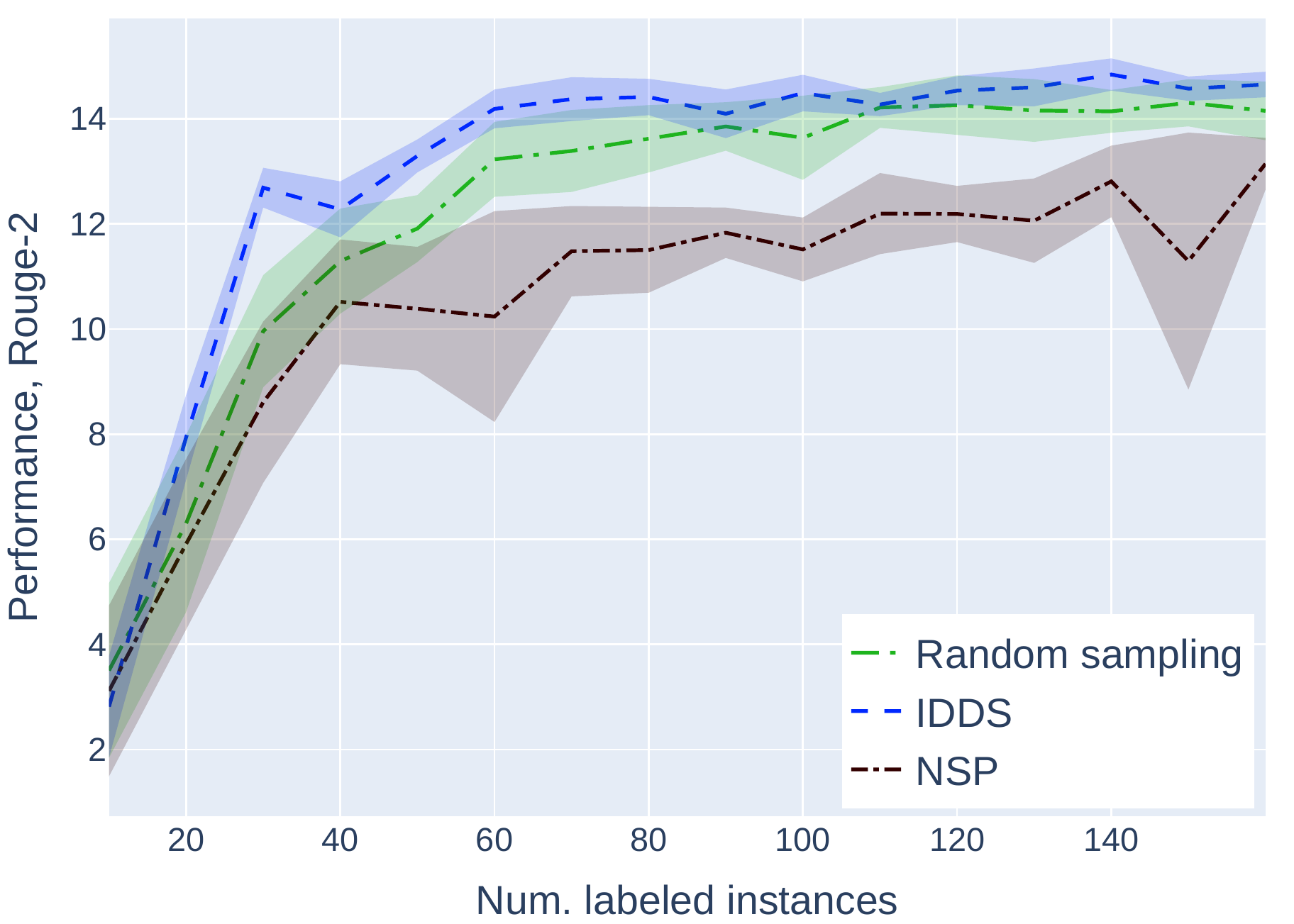} b) ROUGE-2}
    \vspace{0.3cm}
    \end{minipage}
    \hspace{0.1cm}
    \begin{minipage}[ht]{0.32\linewidth}
    \center{\includegraphics[width=1\linewidth]{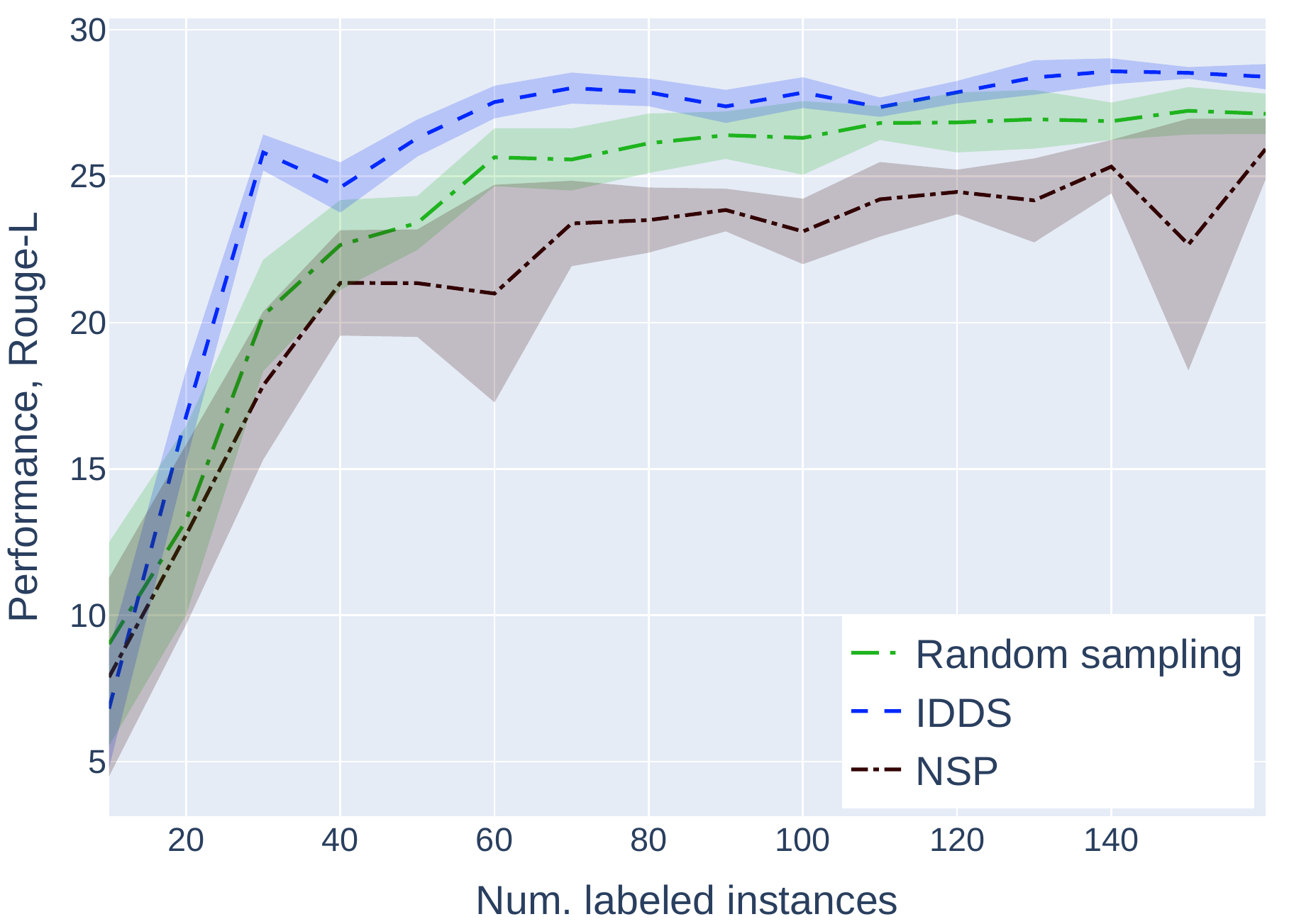} c) ROUGE-L}
    \vspace{0.3cm}
    \end{minipage}
    
    \vspace{-0.2cm}
    \caption{The performance of the PEGASUS-large model with the IDDS strategy compared with random sampling (baseline) and NSP (uncertainty-based strategy) on AESLC.}
    \label{fig:aeslc_embs_pegasus}
\end{figure*}

%% file: figures/wikiall_embs_bart.tex
\begin{figure*}[ht]
    \footnotesize
    \centering
    \begin{minipage}[ht]{0.32\linewidth}
    \vspace{-0.3cm}
    \center{\includegraphics[width=1\linewidth]{figures/wikihow_all/wikiall_embs_rouge1.pdf} a) ROUGE-1}
    \end{minipage}
    \hspace{0.1cm}
    \begin{minipage}[ht]{0.32\linewidth}
    \center{\includegraphics[width=1\linewidth]{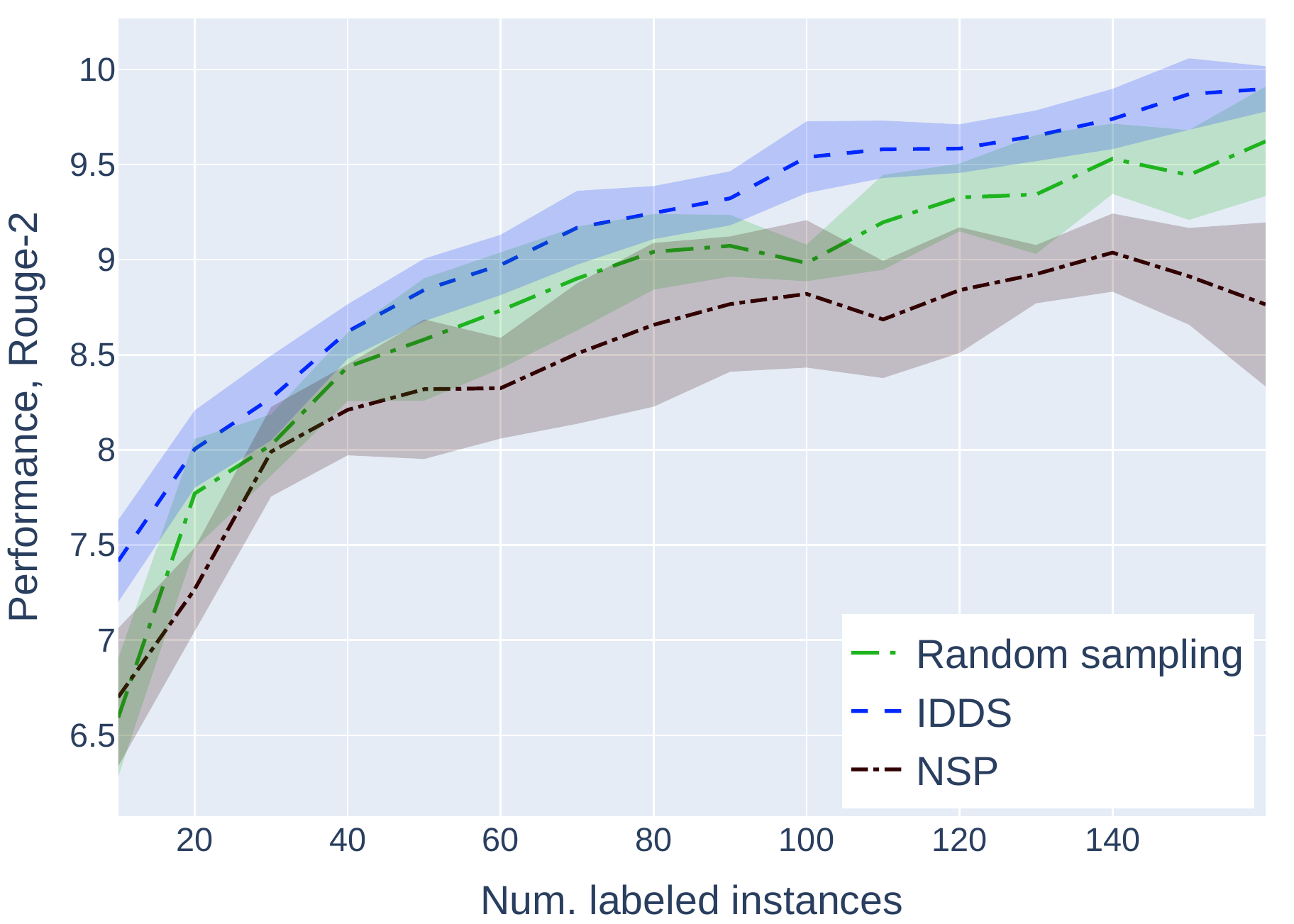} b) ROUGE-2}
    \vspace{0.3cm}
    \end{minipage}
    \hspace{0.1cm}
    \begin{minipage}[ht]{0.32\linewidth}
    \center{\includegraphics[width=1\linewidth]{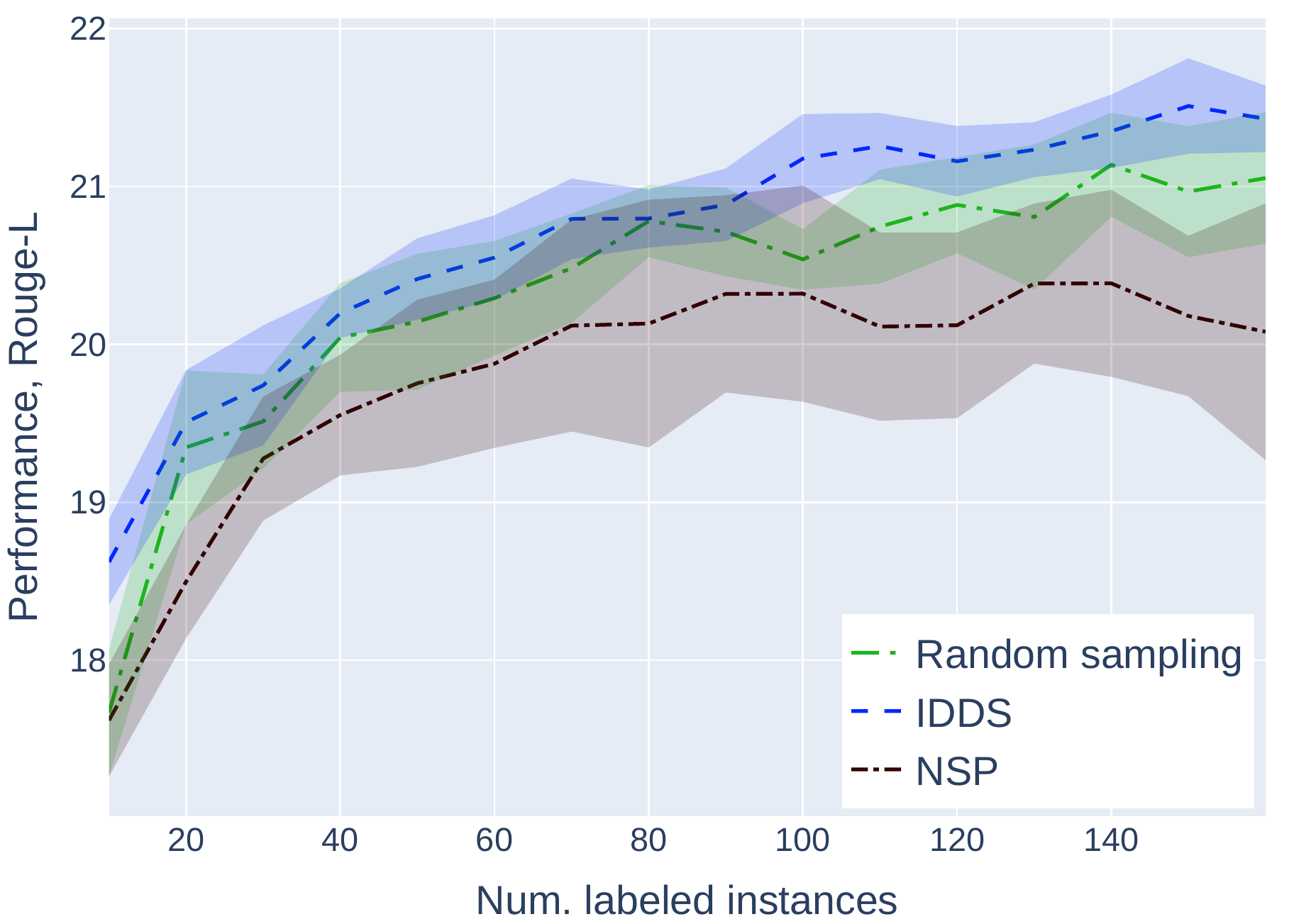} c) ROUGE-L}
    \vspace{0.3cm}
    \end{minipage}
    
    \vspace{-0.2cm}
    \caption{The performance of the BART-base model with the IDDS strategy compared with random sampling (baseline) and NSP (uncertainty-based strategy) and NSP (uncertainty-based strategy) on WikiHow.}
    \label{fig:wikiall_embs}
\end{figure*}

%% file: figures/pubmed_embs_bart.tex
\begin{figure*}[ht]
    \footnotesize
    \centering
    \begin{minipage}[ht]{0.32\linewidth}
    \vspace{-0.3cm}
    \center{\includegraphics[width=1\linewidth]{figures/pubmed/pubmed_embs_rouge1.pdf} a) ROUGE-1}
    \end{minipage}
    \hspace{0.1cm}
    \begin{minipage}[ht]{0.32\linewidth}
    \center{\includegraphics[width=1\linewidth]{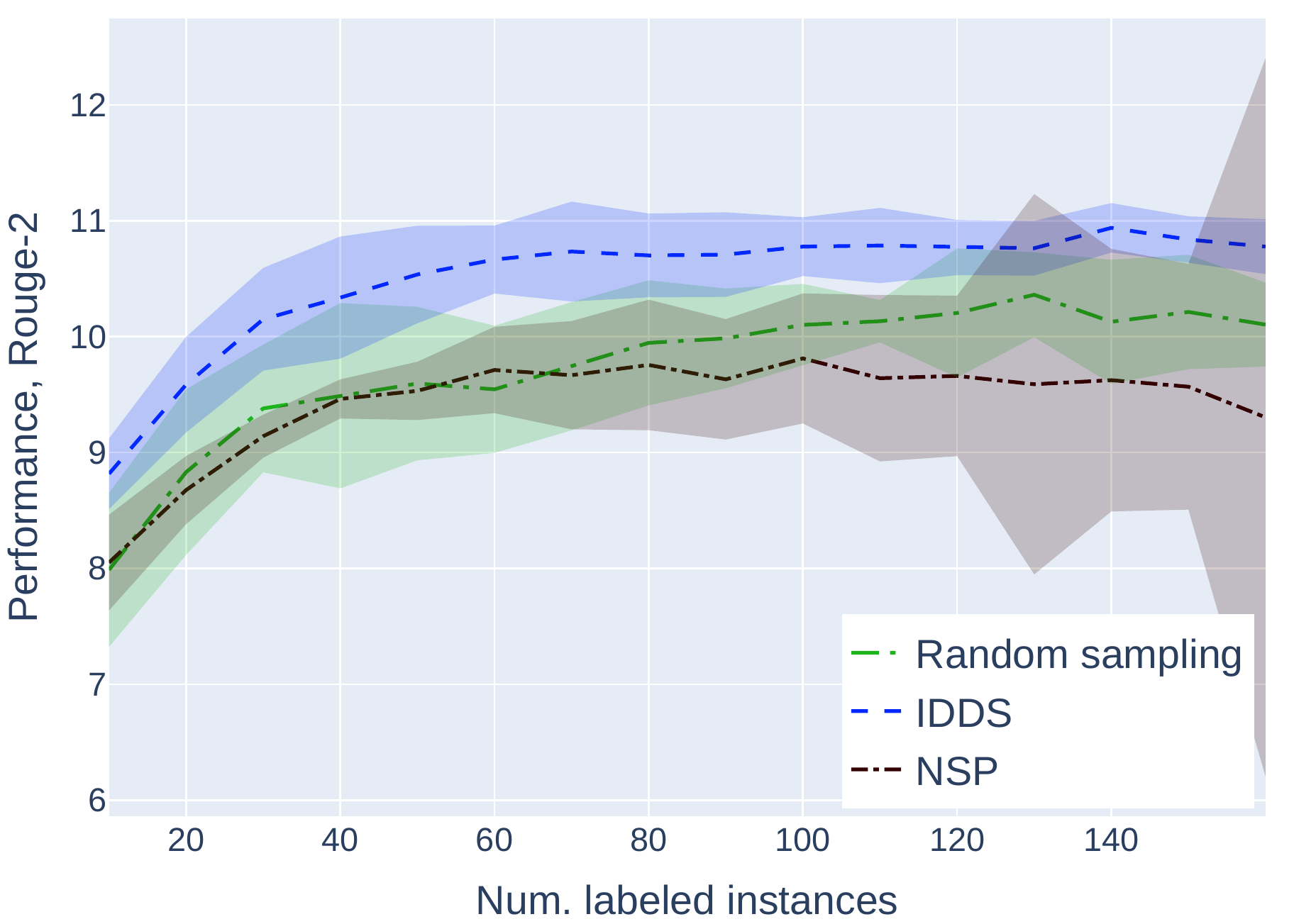} b) ROUGE-2}
    \vspace{0.3cm}
    \end{minipage}
    \hspace{0.1cm}
    \begin{minipage}[ht]{0.32\linewidth}
    \center{\includegraphics[width=1\linewidth]{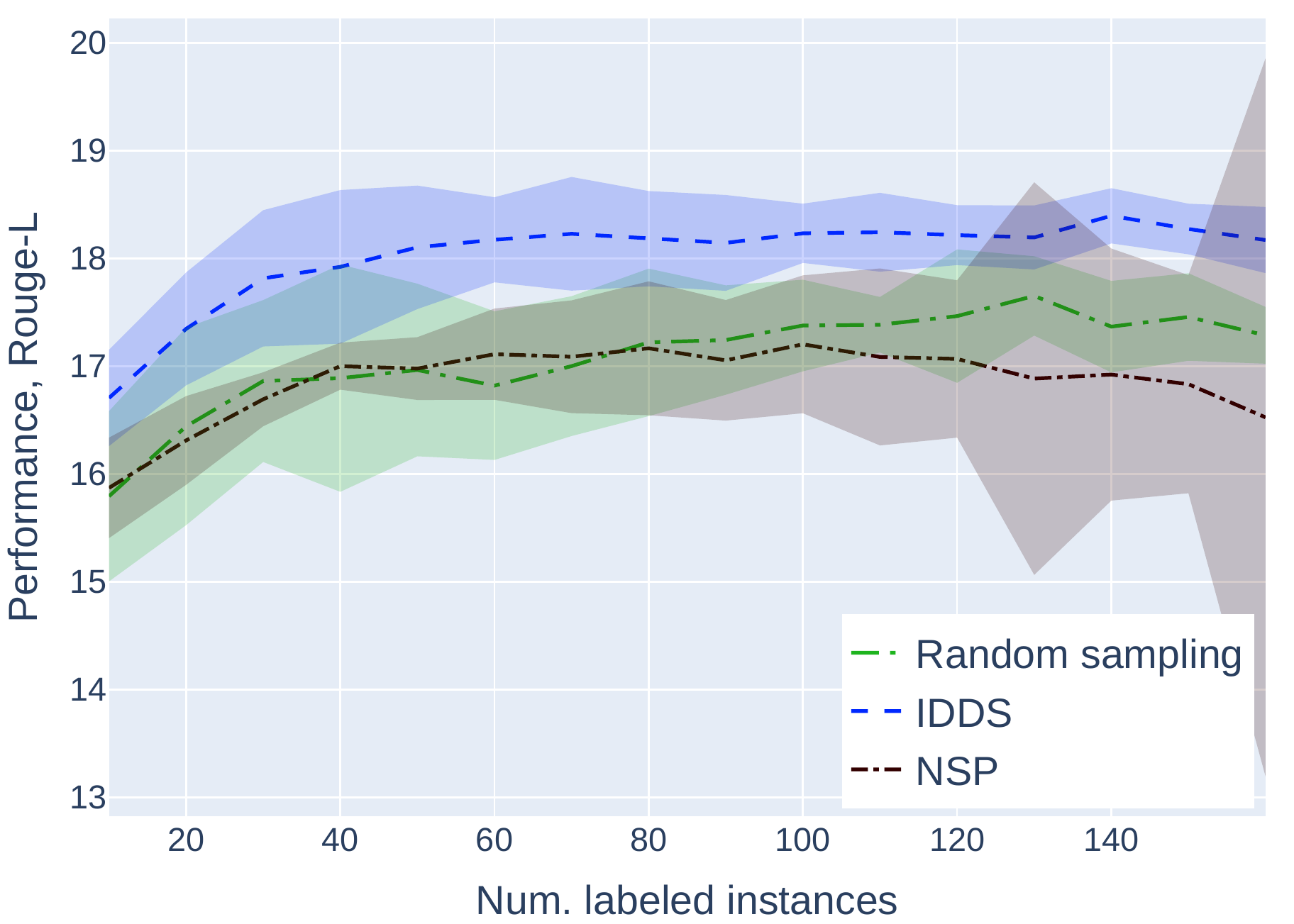} c) ROUGE-L}
    \vspace{0.3cm}
    \end{minipage}
    
    \vspace{-0.2cm}
    \caption{The performance of the BART-base model with the IDDS strategy compared with random sampling (baseline) and NSP (uncertainty-based strategy) on PubMed.}
    \label{fig:pubmed_embs}
\end{figure*}

%% file: figures/aeslc_pl_bart.tex
\begin{figure*}[ht]
    \footnotesize
    \centering
    \begin{minipage}[ht]{0.32\linewidth}
    \vspace{-0.3cm}
    \center{\includegraphics[width=1\linewidth]{figures/aeslc/aeslc_pl_rouge1.pdf} a) ROUGE-1}
    \end{minipage}
    \hspace{0.1cm}
    \begin{minipage}[ht]{0.32\linewidth}
    \center{\includegraphics[width=1\linewidth]{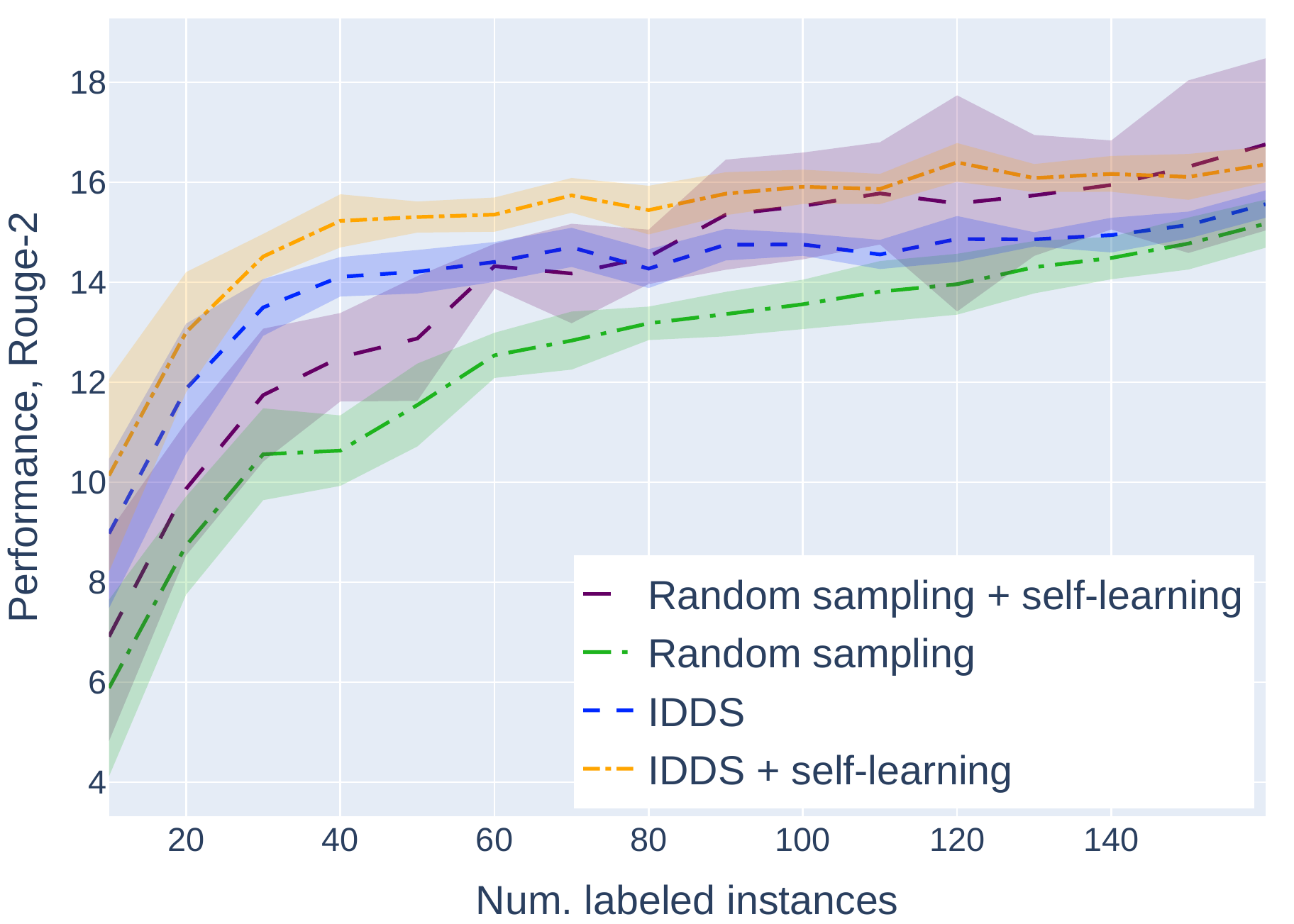} b) ROUGE-2}
    \vspace{0.3cm}
    \end{minipage}
    \hspace{0.1cm}
    \begin{minipage}[ht]{0.32\linewidth}
    \center{\includegraphics[width=1\linewidth]{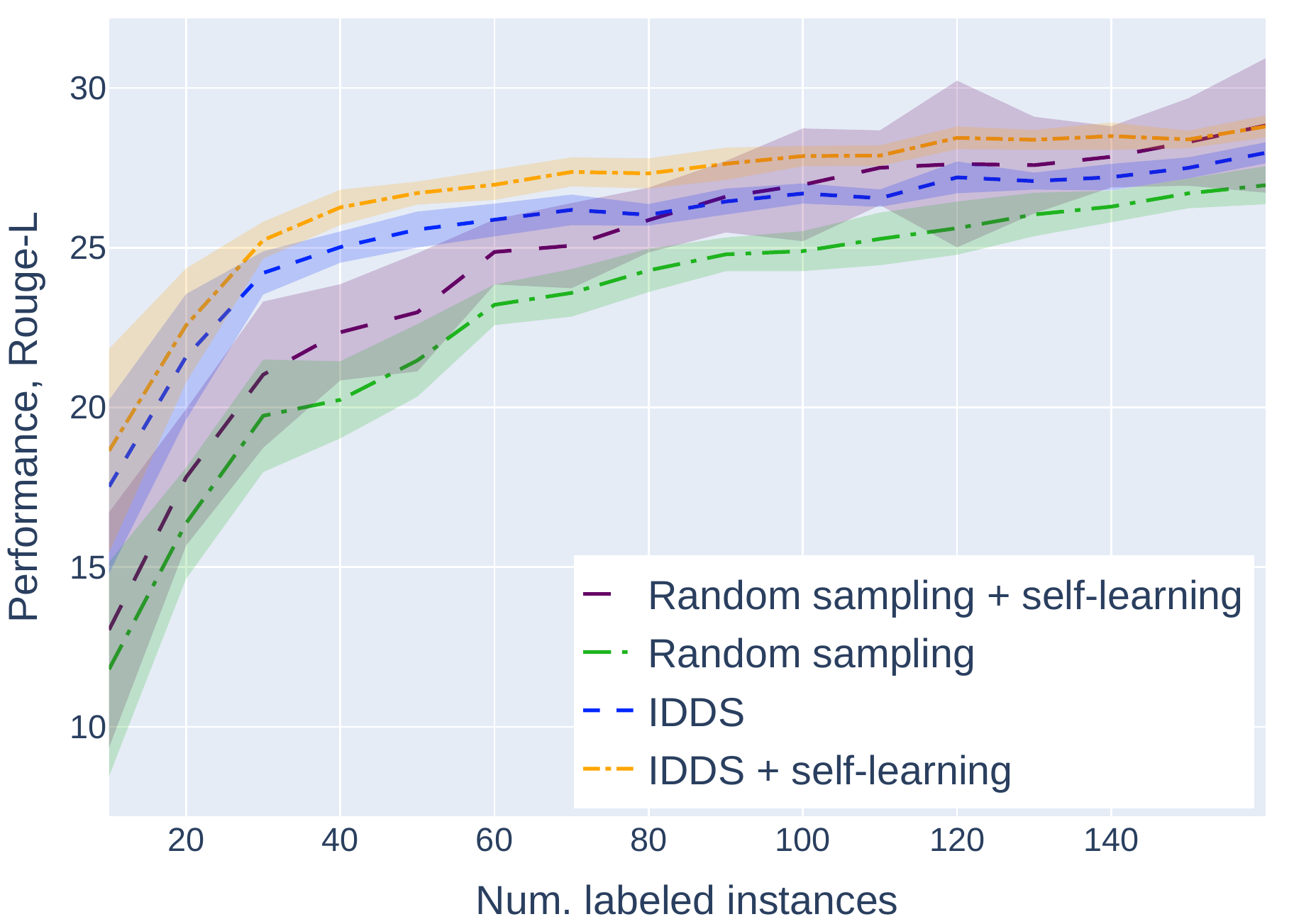} c) ROUGE-L}
    \vspace{0.3cm}
    \end{minipage}
    
    \vspace{-0.2cm}
    \caption{The performance of the BART-base model with the IDDS and  random sampling strategies with and without pseudo-labeling of the unlabeled data on AESLC ($k_l = 0.1, k_h = 0.01$).}
    \label{fig:aeslc_pl}
\end{figure*}

%% file: figures/gigaword_pl_bart.tex
\begin{figure*}[ht]
    \footnotesize
    \centering
    \begin{minipage}[ht]{0.32\linewidth}
    \vspace{-0.3cm}
    \center{\includegraphics[width=1\linewidth]{figures/gigaword/gigaword_pl_rouge1.pdf} a) ROUGE-1}
    \end{minipage}
    \hspace{0.1cm}
    \begin{minipage}[ht]{0.32\linewidth}
    \center{\includegraphics[width=1\linewidth]{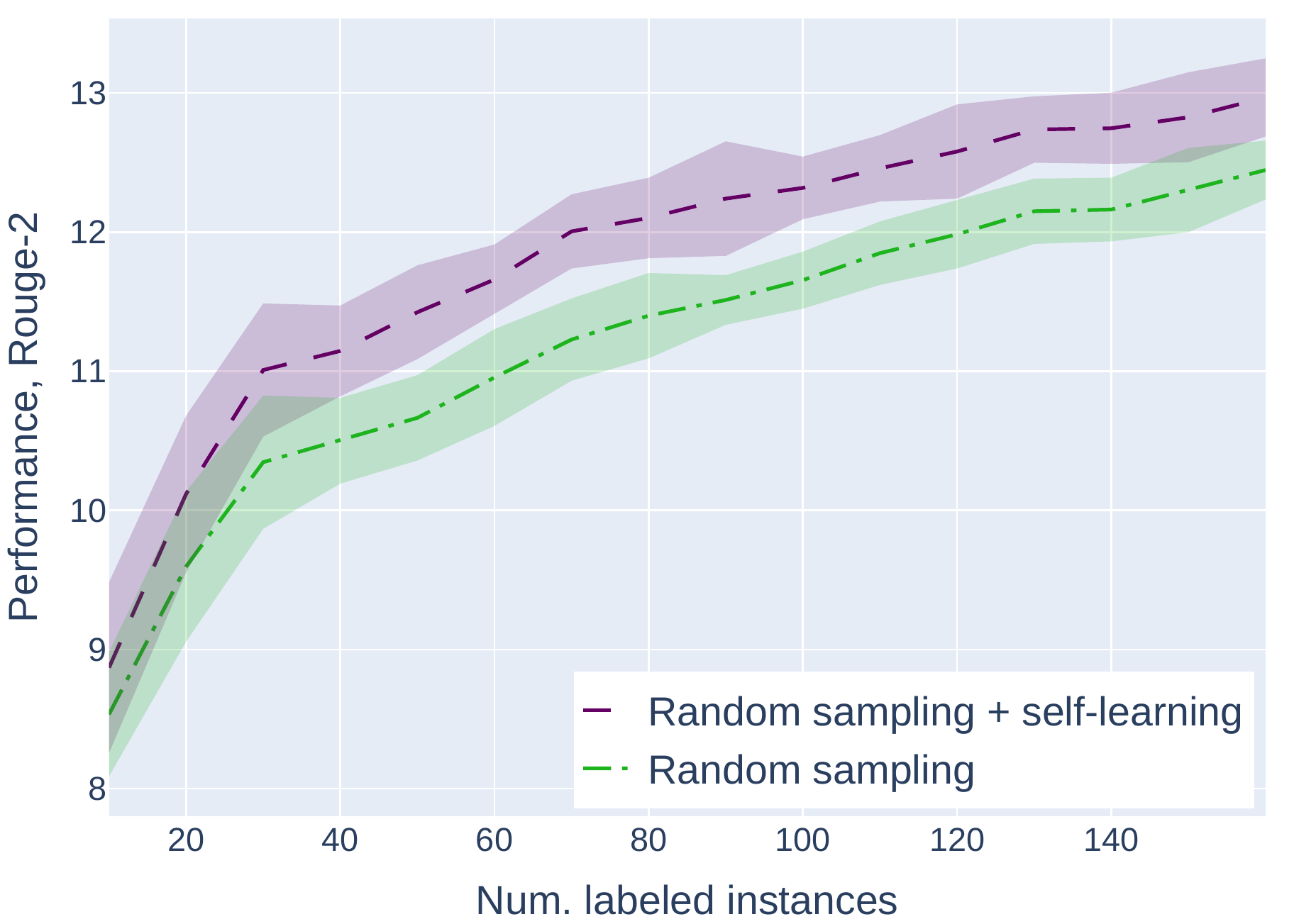} b) ROUGE-2}
    \vspace{0.3cm}
    \end{minipage}
    \hspace{0.1cm}
    \begin{minipage}[ht]{0.32\linewidth}
    \center{\includegraphics[width=1\linewidth]{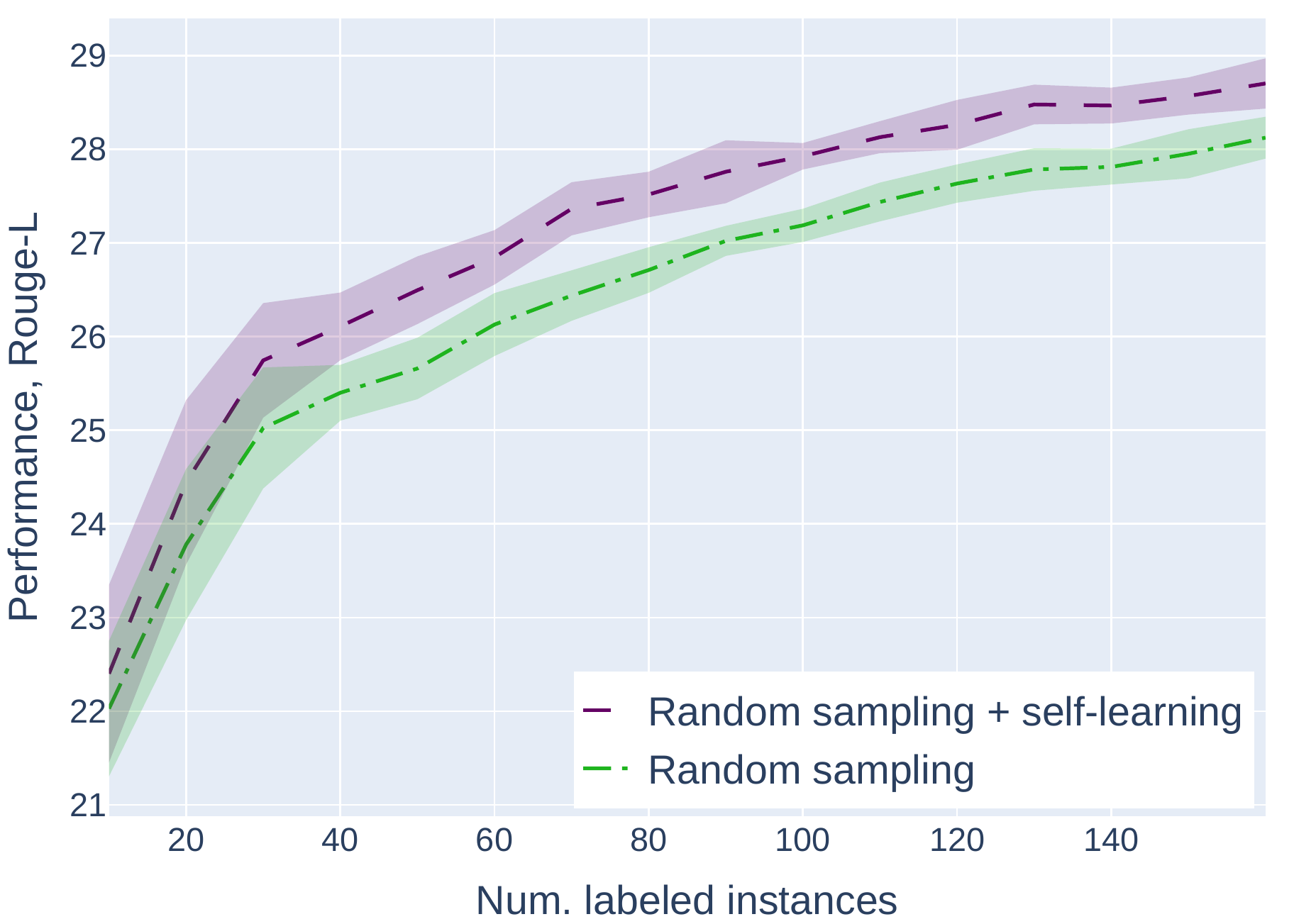} c) ROUGE-L}
    \vspace{0.3cm}
    \end{minipage}
    
    \vspace{-0.2cm}
    \caption{The performance of the BART-base model without AL (random sampling) with and without pseudo-labeling of the unlabeled data on the randomly sampled subset of Gigaword ($k_l = 0.1, k_h = 0.01$).}
    \label{fig:gigaword_pl}
\end{figure*}

%% file: figures/pubmed_pl_bart.tex
\begin{figure*}[ht]
    \footnotesize
    \centering
    \begin{minipage}[ht]{0.32\linewidth}
    \vspace{-0.3cm}
    \center{\includegraphics[width=1\linewidth]{figures/wikihow_all/wikiall_pl_rouge1.pdf} a) ROUGE-1}
    \end{minipage}
    \hspace{0.1cm}
    \begin{minipage}[ht]{0.32\linewidth}
    \center{\includegraphics[width=1\linewidth]{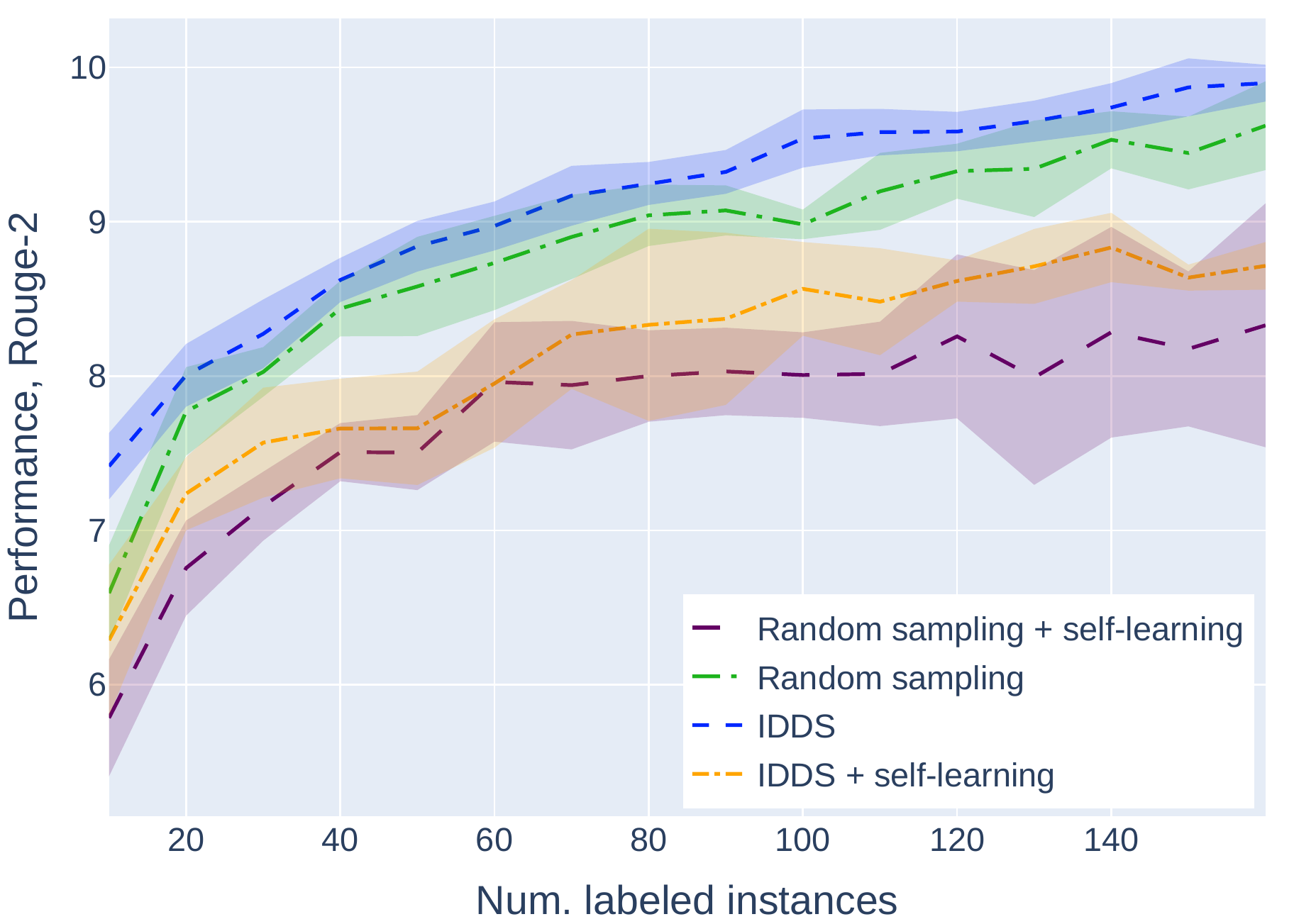} b) ROUGE-2}
    \vspace{0.3cm}
    \end{minipage}
    \hspace{0.1cm}
    \begin{minipage}[ht]{0.32\linewidth}
    \center{\includegraphics[width=1\linewidth]{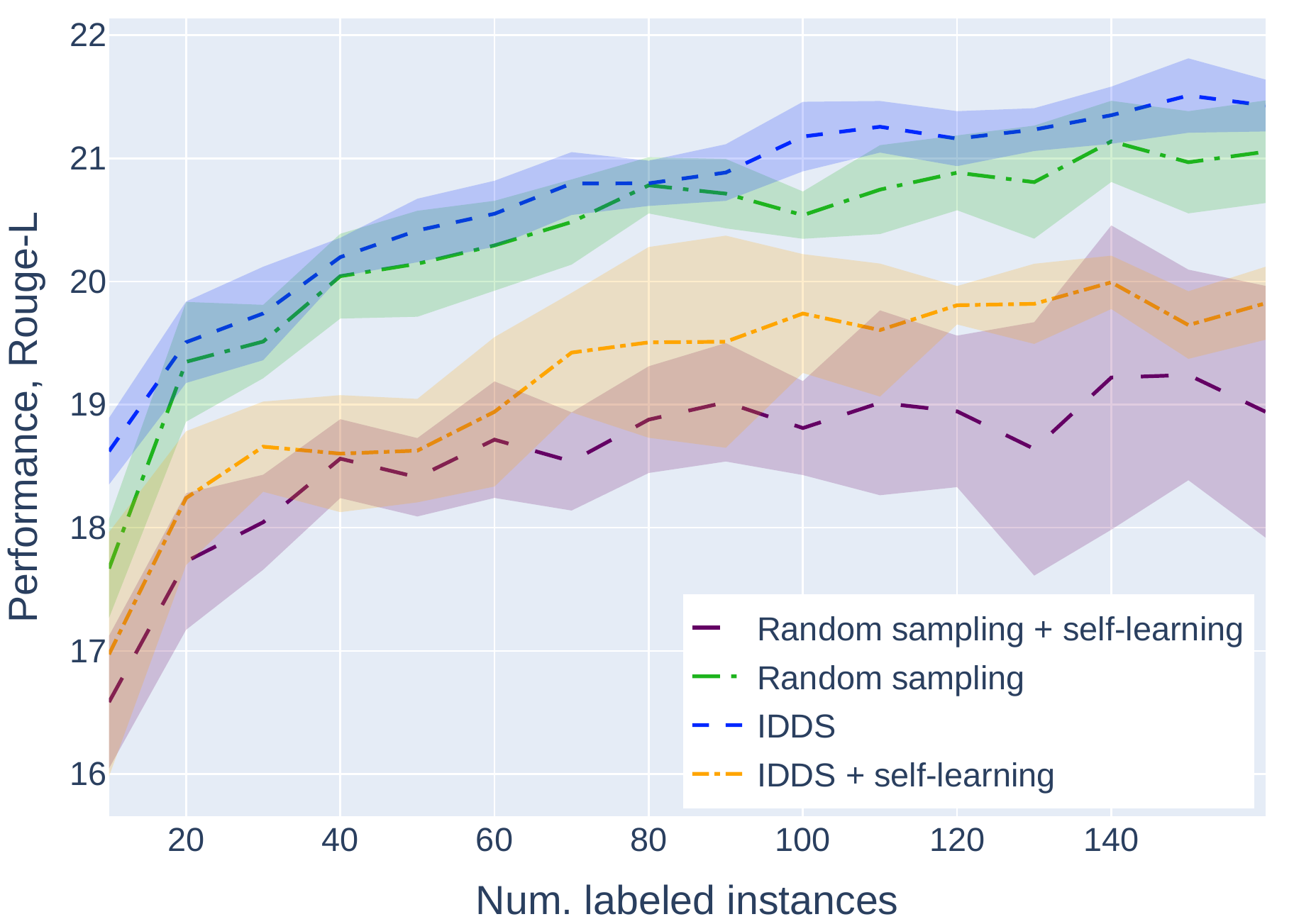} c) ROUGE-L}
    \vspace{0.3cm}
    \end{minipage}
    
    \vspace{-0.2cm}
    \caption{The performance of the BART-base model with the IDDS and random sampling strategies with and without self-supervised learning on WikiHow ($k_l = 0.38, k_h = 0.02)$.}
    \label{fig:wikihow_pl}
    \vspace{-0.4cm}
\end{figure*}

%% file: figures/larger_examples.tex
\begin{table*}[h]
\centering
\footnotesize

\begin{tabular}{|l|l|l|c|}
\hline
\textbf{AL Strat.} & \textbf{Document} & \textbf{Golden summary} & \textbf{Gen. summary} \\ \hline

IDDS & \begin{tabular}[c]{@{}l@{}}
"Here's the latest info. regarding \colorgreen{Bloomberg's} ability to \\ accept \colorgreen{deals} with  \colorgreen{Pinnacle} West (formerly \colorblue{Arizona} \\ \colorblue{Public Service} Co.) (...) \end{tabular} & \begin{tabular}[c]{@{}l@{}} \colorgreen{Bloomberg-} \\ \colorgreen{Pinnacle}/\colorblue{APS} \colorgreen{deals}  \end{tabular} & n/a \\ \hline

IDDS & \begin{tabular}[c]{@{}l@{}}Hi. Nice to see you in Houston. I'm giving a\\
\colorgreen{presentation} on \colorgreen{gas} issues on  Tuesday.\\ I've got a draft of (...)\end{tabular} & \begin{tabular}[c]{@{}l@{}} \colorgreen{Gas} \colorgreen{Presentation} \end{tabular} & n/a \\ \hline

IDDS & \begin{tabular}[c]{@{}l@{}}Kelley,  I am writing to you to (...) Can you give  \\ me the name and contact information for the person within \\
your company that would work with us to put a \\\colorgreen{Confidentiality}  \colorgreen{Agreement} in place (...) \end{tabular} & \begin{tabular}[c]{@{}l@{}} \colorgreen{confidentiality} \colorgreen{agreement} \end{tabular} & n/a \\ \hline

NSP & \begin{tabular}[c]{@{}l@{}}\colorgreen{tantivy} (tan-TIV-ee) adverb     At full gallop; at full speed. \\ noun     A fast gallop; rush.adjective     Swift.interjection     A \\ hunting 
cry by a hunter riding a horse at full speed(...) \end{tabular} & \begin{tabular}[c]{@{}l@{}} \colorred{A.Word.A.Day--\colorgreen{tantivy}} \end{tabular} & \begin{tabular}[c]{@{}l@{}} tricky \\ (tan-TIV-ee) \\adjective \end{tabular} \\ \hline

NSP & \begin{tabular}[c]{@{}l@{}}Prod Area and Long Haul k\# \colorblue{Volume}  Rec \\ Del 3.6746 5000  St 62  Con Ed 3.4358 \\15000  St 65 Con Ed 3.5049 10000  St 
(...) \end{tabular} & \begin{tabular}[c]{@{}l@{}} \colorred{TRCO} \colorblue{capacity} \colorred{for Sep
} \end{tabular} &\begin{tabular}[c]{@{}l@{}}  Prod Area \\and Long \\Haul k\# \\Volume \end{tabular} \\ \hline

NSP & \begin{tabular}[c]{@{}l@{}}This is a list of RisktRAC book-ids corresponding to\\ what has been created in ERMS. Let me know if the\\ book-id  naming is ok with you. \\ Regards \end{tabular} & \begin{tabular}[c]{@{}l@{}}  \colorred{Book2.xls
} \end{tabular} & RisktRAC \\ \hline

ENSP & \begin{tabular}[c]{@{}l@{}}Fred,  I suggest a phone call among the team today \\ to  make sure we are all on the  same wave length. \\ What is your schedule? \\ Thanks \end{tabular} & \begin{tabular}[c]{@{}l@{}}  \colorred{PSEG
} \end{tabular} & \begin{tabular}[c]{@{}l@{}} Firm \\ schedule  \end{tabular}\\ \hline

ENSP & \begin{tabular}[c]{@{}l@{}}Stephanie -  When you get a chance, could you finalize the \\ attached (also found in Tana's O drive). I am not sure where \\ the originals need to go after signed by Enron, but I have a \\ request for that information currently out to \colorgreen{Hess}. Thanks. \end{tabular} & \begin{tabular}[c]{@{}l@{}}  \colorgreen{Hess} \colorred{NDA
} \end{tabular} & \begin{tabular}[c]{@{}l@{}}  Enron \\O Drive \end{tabular}\\ \hline

ENSP & \begin{tabular}[c]{@{}l@{}}Current Notes User: To ensure that you experience  \\ a successful migration from Notes to Outlook, it is \\ necessary  to gather individual user information  prior\\ to your date of migration. Please take a few  \\ minutes to  completely fill
out the following survey (...) \end{tabular} & \begin{tabular}[c]{@{}l@{}}   \colorred{2- SURVEY} \\
\colorred{/INFORMATION}\\
\colorred{EMAIL 5-17-01
} \end{tabular} & \begin{tabular}[c]{@{}l@{}}Office 2000 \\Migration \\Survey\end{tabular}\\ \hline

\begin{tabular}[c]{@{}l@{}}Sacre- \\ BleuVAR \end{tabular} & \begin{tabular}[c]{@{}l@{}}Sheri, We are going to NO for JazzFest at the end of April.\\ April 27th-29th to be  exact. \\Let me know if you're going.\\DG \end{tabular} & \begin{tabular}[c]{@{}l@{}}   \colorred{southwest.com} weekly \\ specials \end{tabular} & JazzFest \\ \hline

\begin{tabular}[c]{@{}l@{}}Sacre- \\ BleuVAR \end{tabular} & \begin{tabular}[c]{@{}l@{}}This warning is sent automatically to inform you that your \\ mailbox is approaching the maximum size limit. Your \\ mailbox size is currently 78515  KB. Mailbox size limits (...) \end{tabular} & \begin{tabular}[c]{@{}l@{}}   \colorred{WARNING:  Your}
 \\
 \colorred{mailbox is approaching} \\ \colorred{the size limit} \end{tabular} & \begin{tabular}[c]{@{}l@{}} Mailbox \\size limit \end{tabular}\\ \hline

\end{tabular}

\caption{Examples of the instances queried with different AL strategies. Tokens overlapping with the source document are highlighted with \colorgreen{green}. Tokens that refer to paraphrasing the part of the document and the corresponding part are highlighted with \colorblue{blue}. Tokens that cannot be derived from the document are highlighted with \colorred{red}. Tokens, the usage of which depends on the peculiarities of the dataset, are not highlighted. Summaries for IDDS are not presented, because IDDS does not require model inference.} 
\label{tab:larger_examples}

\end{table*}












%% file: figures/diversity_table.tex
\begin{table}[ht!]
\centering
\begin{tabular}{llllll}
\toprule
AL Iter. &           SP &           ESP & SacreBleuVAR &     Random &     IDDS \\
\midrule
1    &   33.3\% / 0\% &  30.0\% / 4.4\% &      0\% / 0\% &    0\% / 0\% &  0\% / 0\% \\
6    &   15.6\% / 0\% &     0\% / 1.1\% &    0\% / 2.2\% &    0\% / 0\% &  0\% / 0\% \\
15   &    3.3\% / 0\% &       0\% / 0\% &      0\% / 0\% &  0\% / 2.2\% &  0\% / 0\% \\
Mean &  7.8\% / 1.0\% &   2.1\% / 0.8\% &  0.1\% / 0.3\% &  0\% / 0.7\% &  \textbf{0\% / 0\%} \\
\bottomrule
\end{tabular}

\caption{Share of fully / partly overlapping summaries among batches of instances, queried with various AL strategies during AL using BART-base model on AESLC. We consider two summaries to be partly overlapping if their ROUGE-1 score > 0.66. The results are averaged across 9 seeds for all the strategies except for IDDS, which has constant seed-independent queries.}
\label{tab:diversity}
\end{table}

%% file: figures/ablation_1_emb_model_aeslc.tex
\begin{figure*}[h]
    \footnotesize
    \centering
    \begin{minipage}[ht]{0.32\linewidth}
    \vspace{-0.3cm}
    \center{\includegraphics[width=1\linewidth]{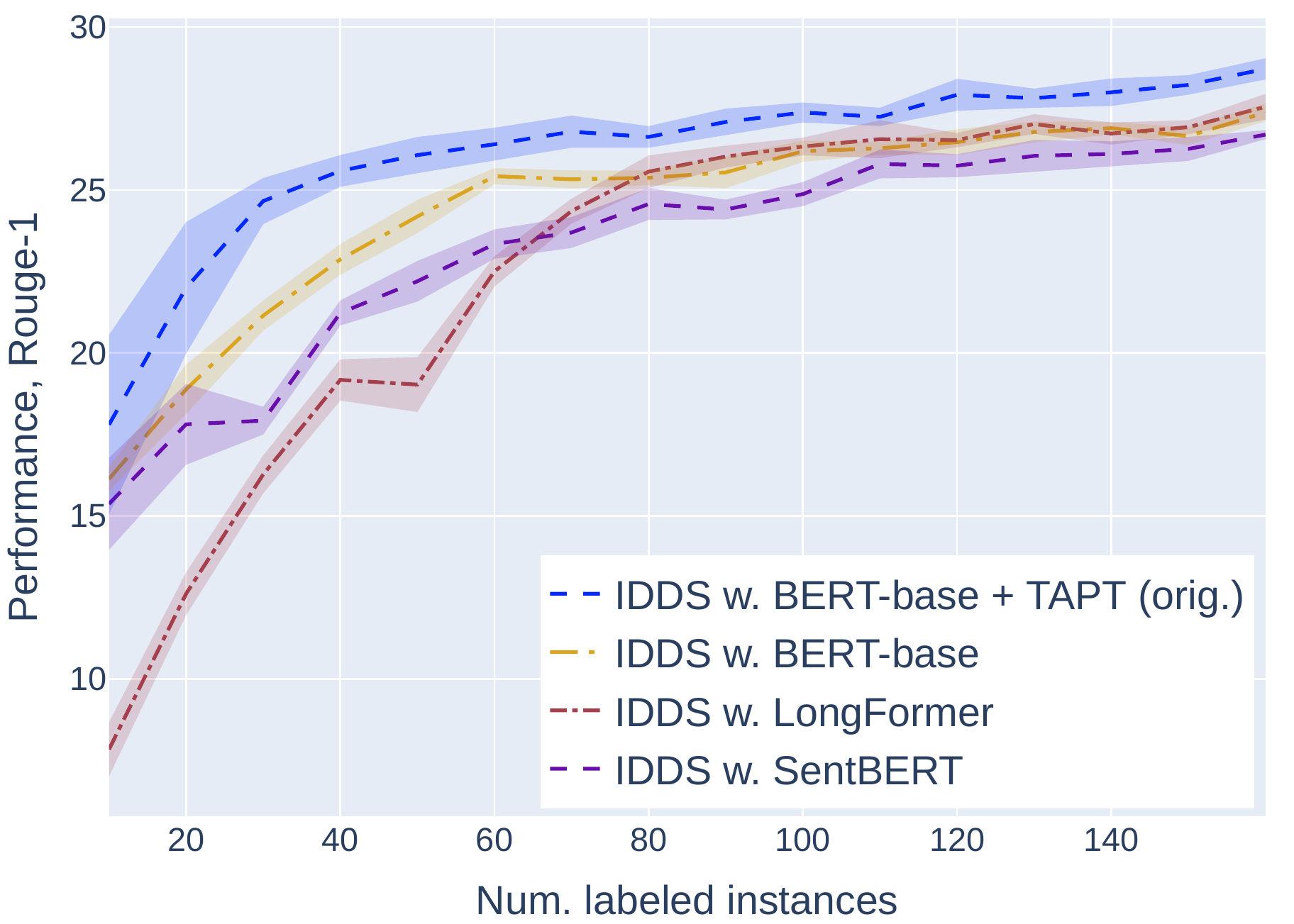} a) ROUGE-1}
    \end{minipage}
    \hspace{0.1cm}
    \begin{minipage}[ht]{0.32\linewidth}
    \center{\includegraphics[width=1\linewidth]{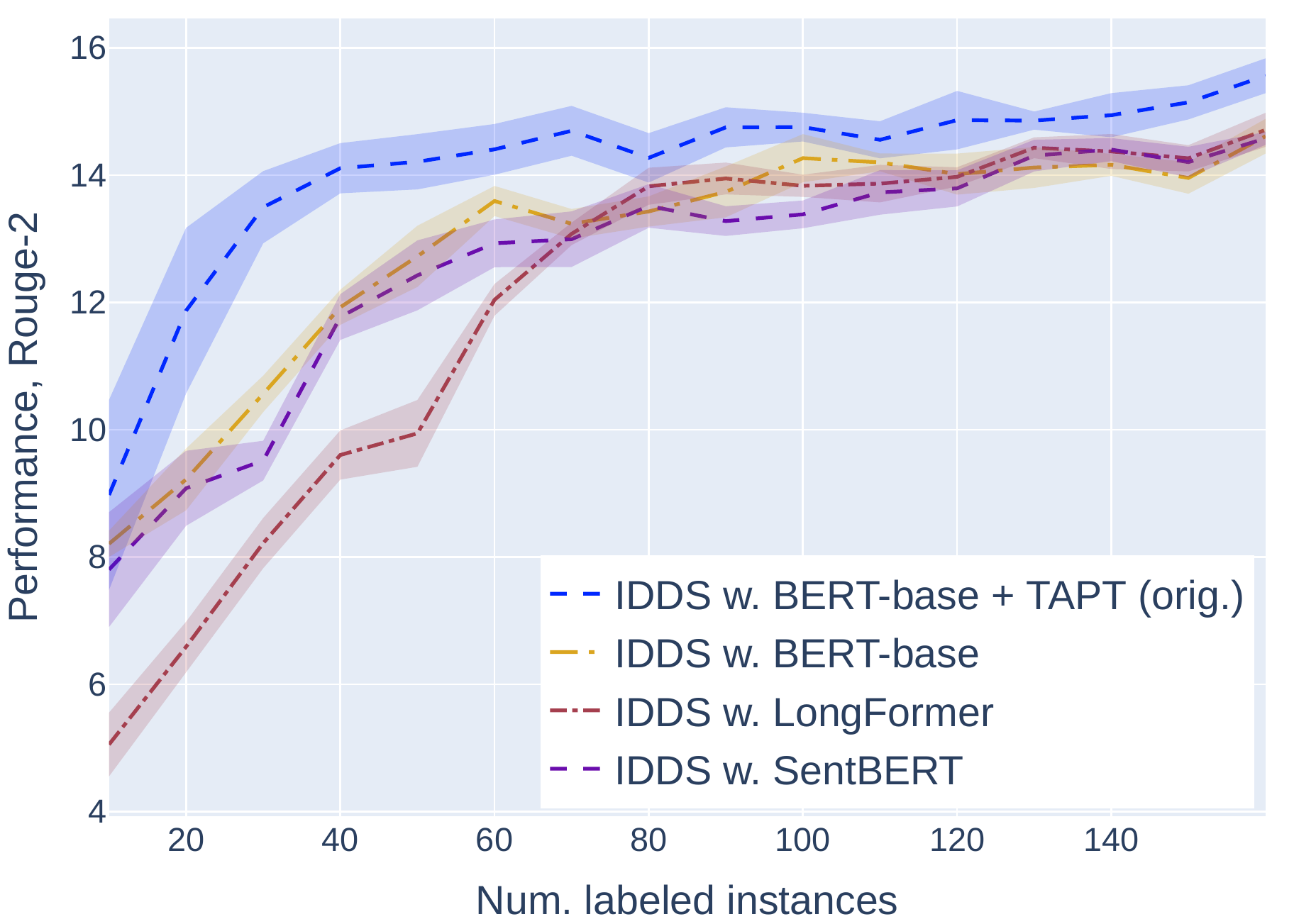} b) ROUGE-2}
    \vspace{0.3cm}
    \end{minipage}
    \hspace{0.1cm}
    \begin{minipage}[ht]{0.32\linewidth}
    \center{\includegraphics[width=1\linewidth]{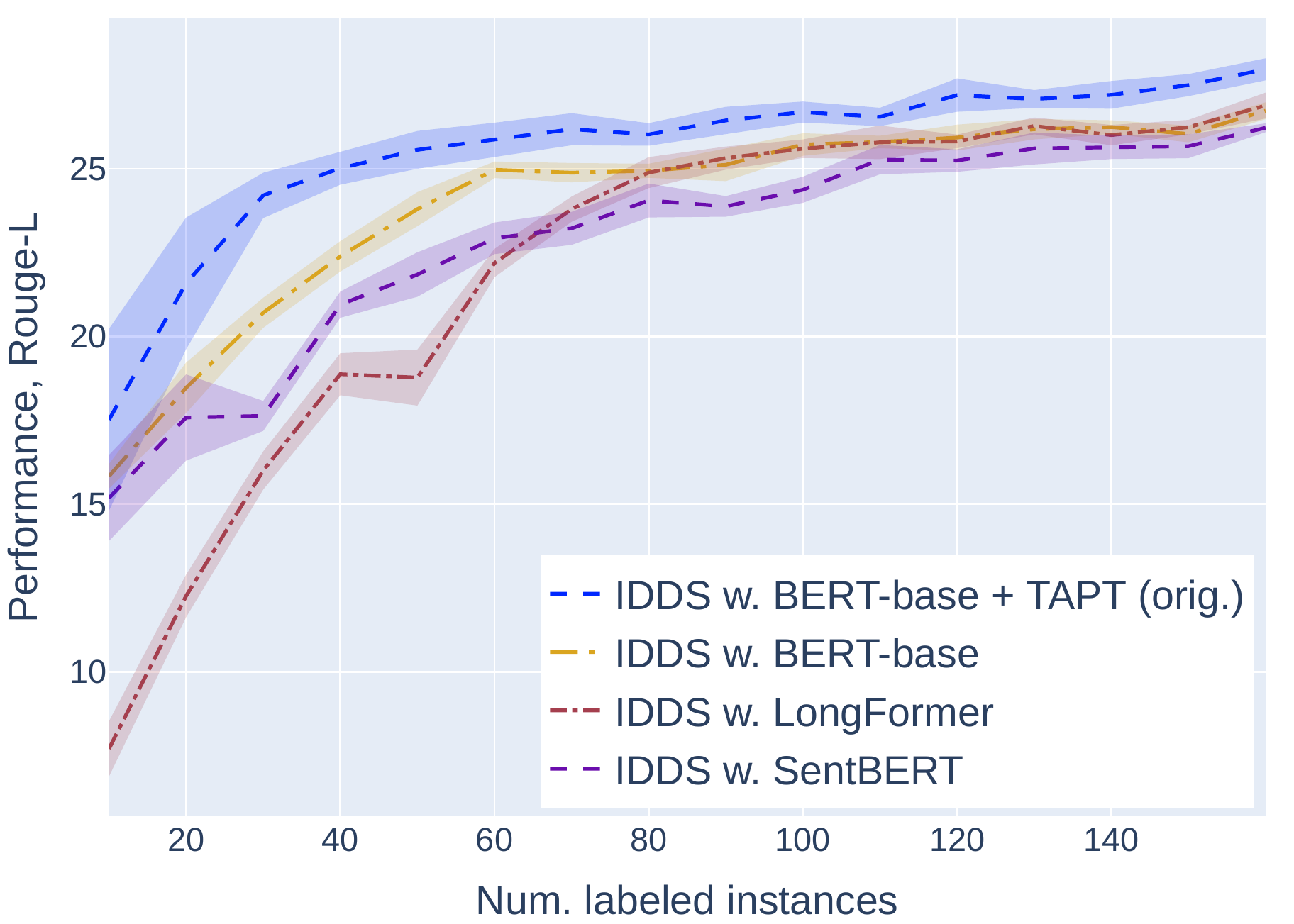} c) ROUGE-L}
    \vspace{0.3cm}
    \end{minipage}
    
    \vspace{-0.2cm}
    \caption{Ablation study of the document embeddings model \& the necessity of performing TAPT for it in the IDDS strategy with BART-base on AESLC.}
    \label{fig:as_emb_model_aeslc}
    \vspace{-0.4cm}
\end{figure*}

%% file: figures/ablation_1_emb_model_wikiall.tex
\begin{figure*}[h]
    \footnotesize
    \centering
    \begin{minipage}[ht]{0.32\linewidth}
    \vspace{-0.3cm}
    \center{\includegraphics[width=1\linewidth]{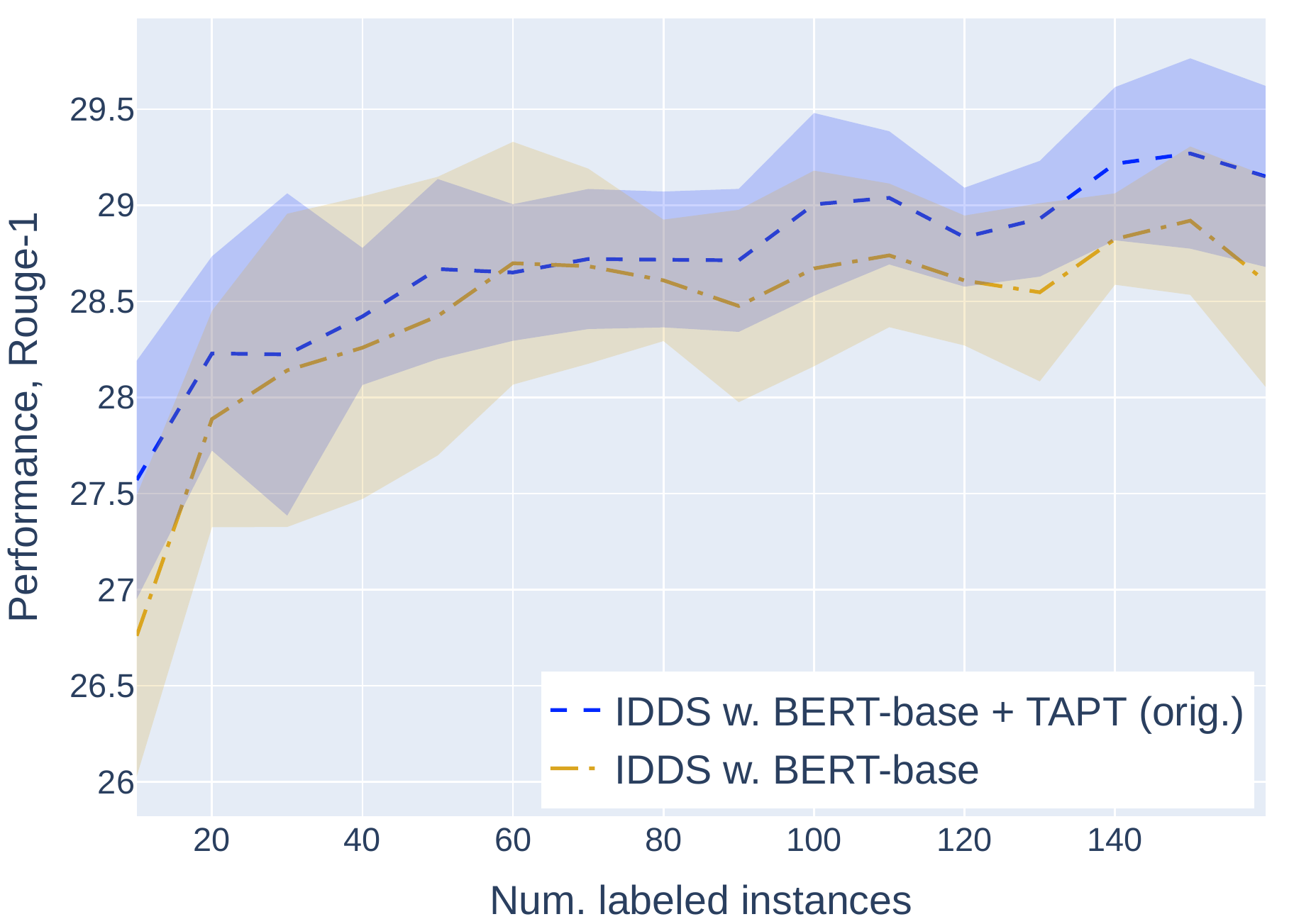} a) ROUGE-1}
    \end{minipage}
    \hspace{0.1cm}
    \begin{minipage}[ht]{0.32\linewidth}
    \center{\includegraphics[width=1\linewidth]{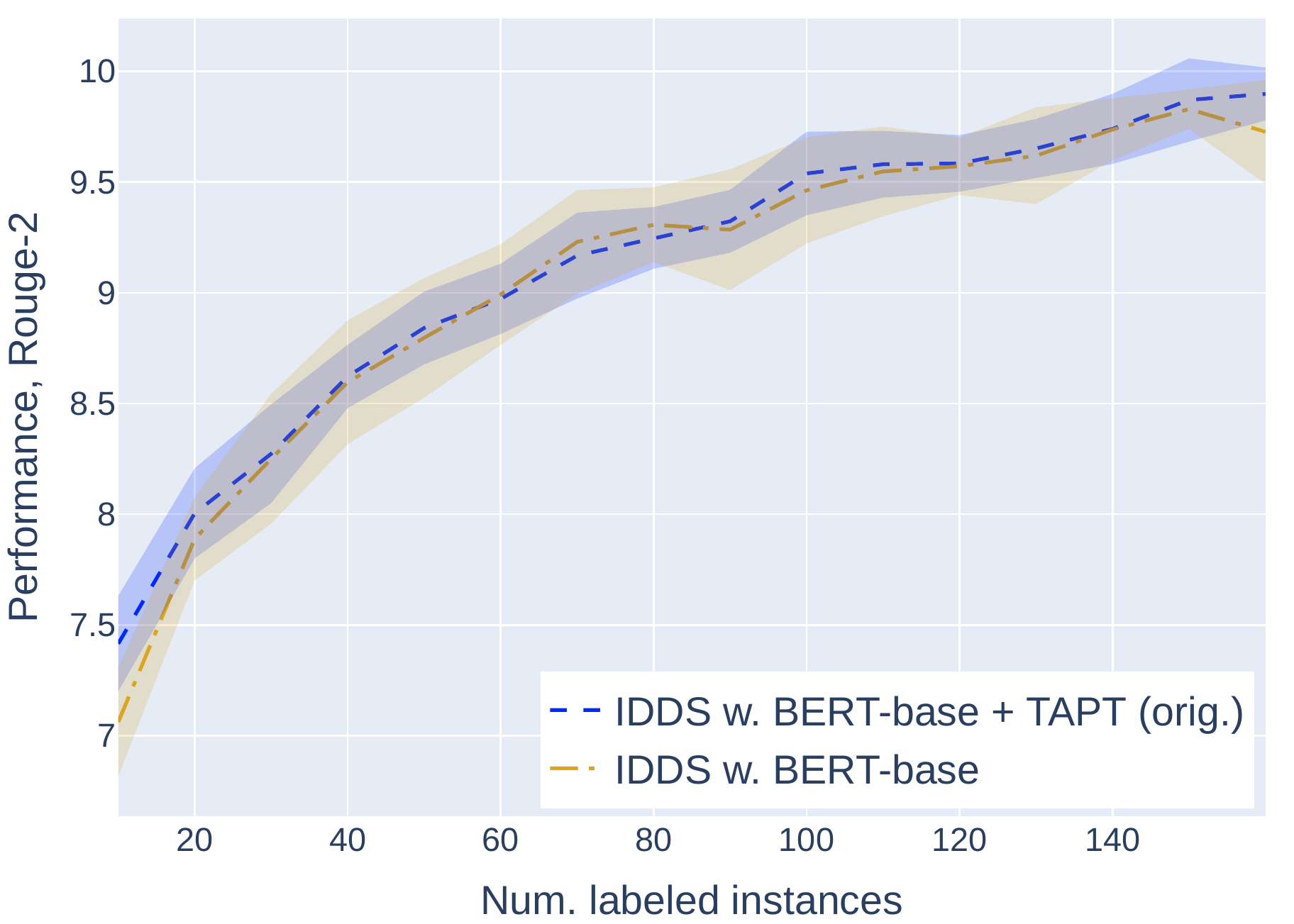} b) ROUGE-2}
    \vspace{0.3cm}
    \end{minipage}
    \hspace{0.1cm}
    \begin{minipage}[ht]{0.32\linewidth}
    \center{\includegraphics[width=1\linewidth]{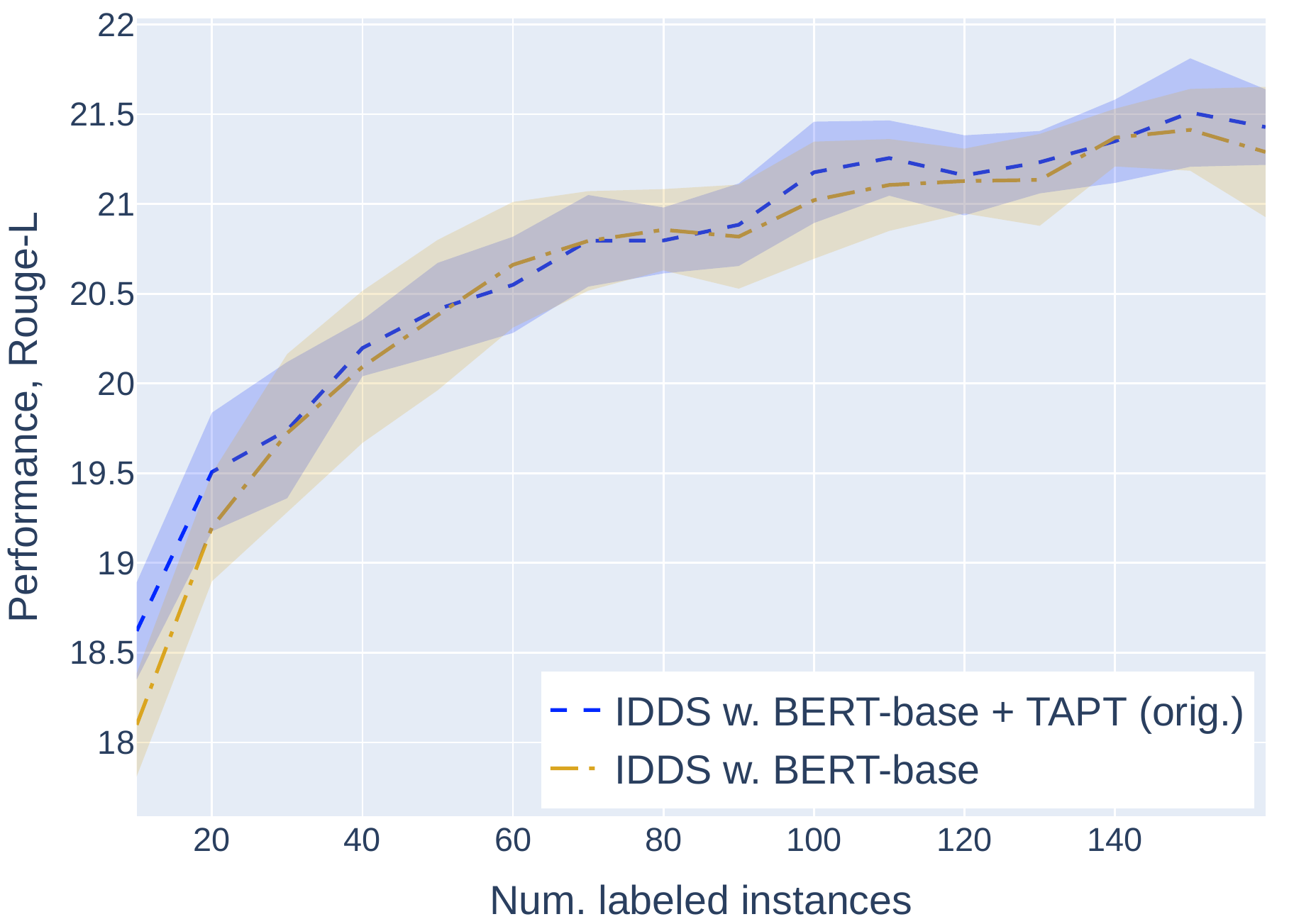} c) ROUGE-L}
    \vspace{0.3cm}
    \end{minipage}

    \vspace{-0.2cm}
    \caption{Ablation study of the necessity of performing TAPT for the model, which generates embeddings in the IDDS strategy with BART-base on WikiHow.}
    \label{fig:as_emb_model_wikiall}
    \vspace{-0.4cm}
\end{figure*}

%% file: figures/ablation_2_sim_functions_aeslc.tex
\begin{figure*}[h!]
    \footnotesize
    \centering
    \begin{minipage}[ht]{0.32\linewidth}
    \vspace{-0.3cm}
    \center{\includegraphics[width=1\linewidth]{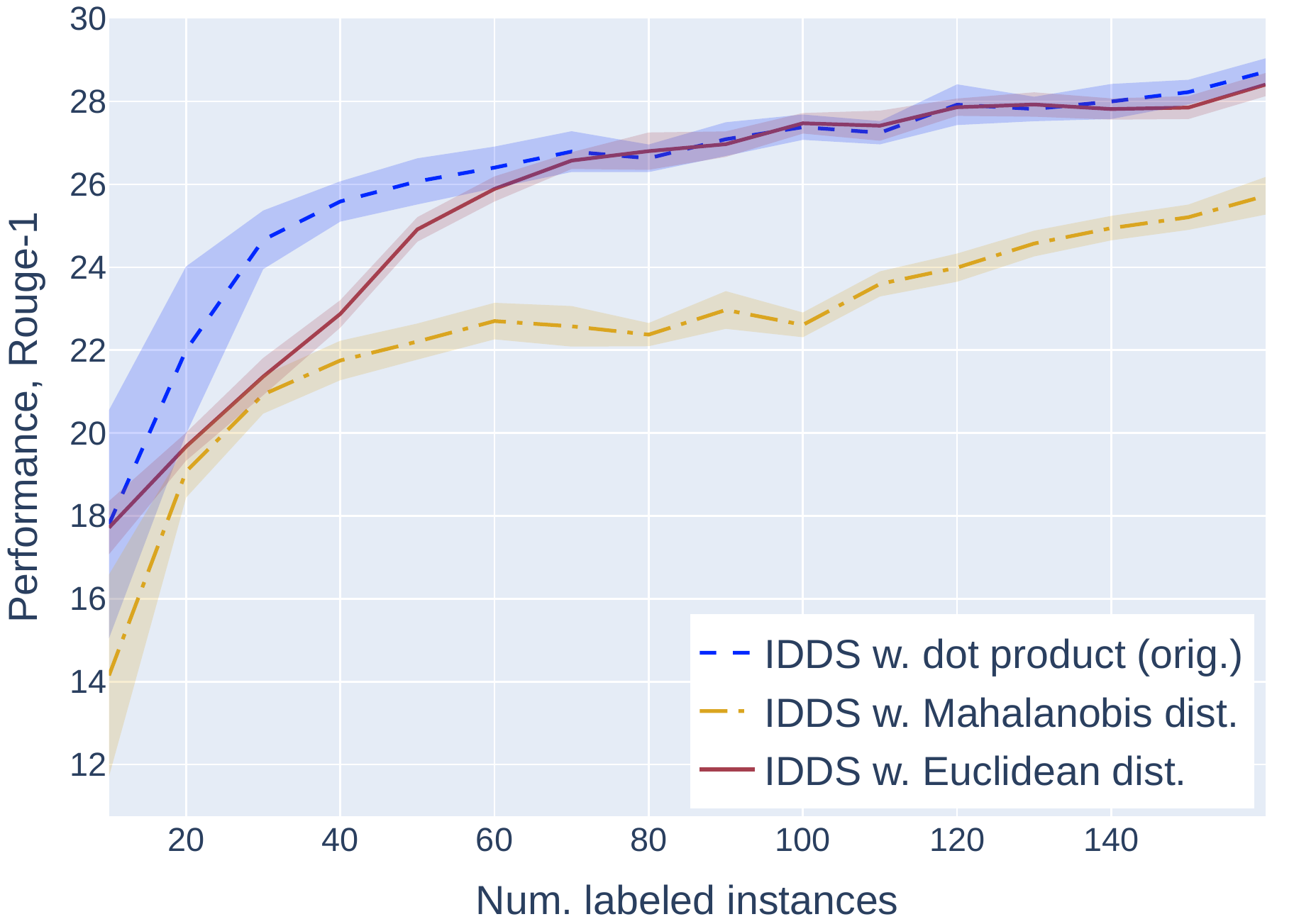} a) ROUGE-1}
    \end{minipage}
    \hspace{0.1cm}
    \begin{minipage}[ht]{0.32\linewidth}
    \center{\includegraphics[width=1\linewidth]{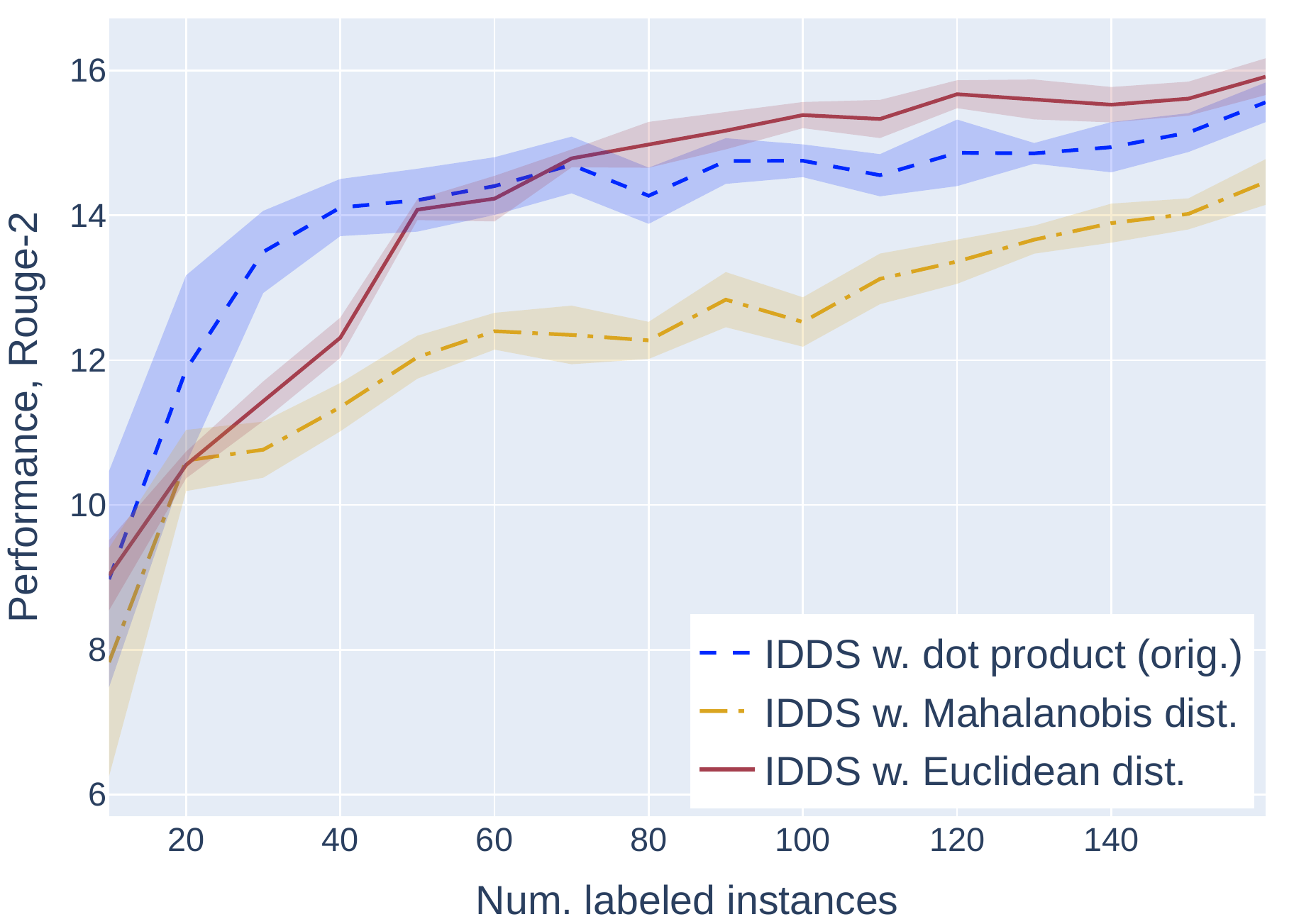} b) ROUGE-2}
    \vspace{0.3cm}
    \end{minipage}
    \hspace{0.1cm}
    \begin{minipage}[ht]{0.32\linewidth}
    \center{\includegraphics[width=1\linewidth]{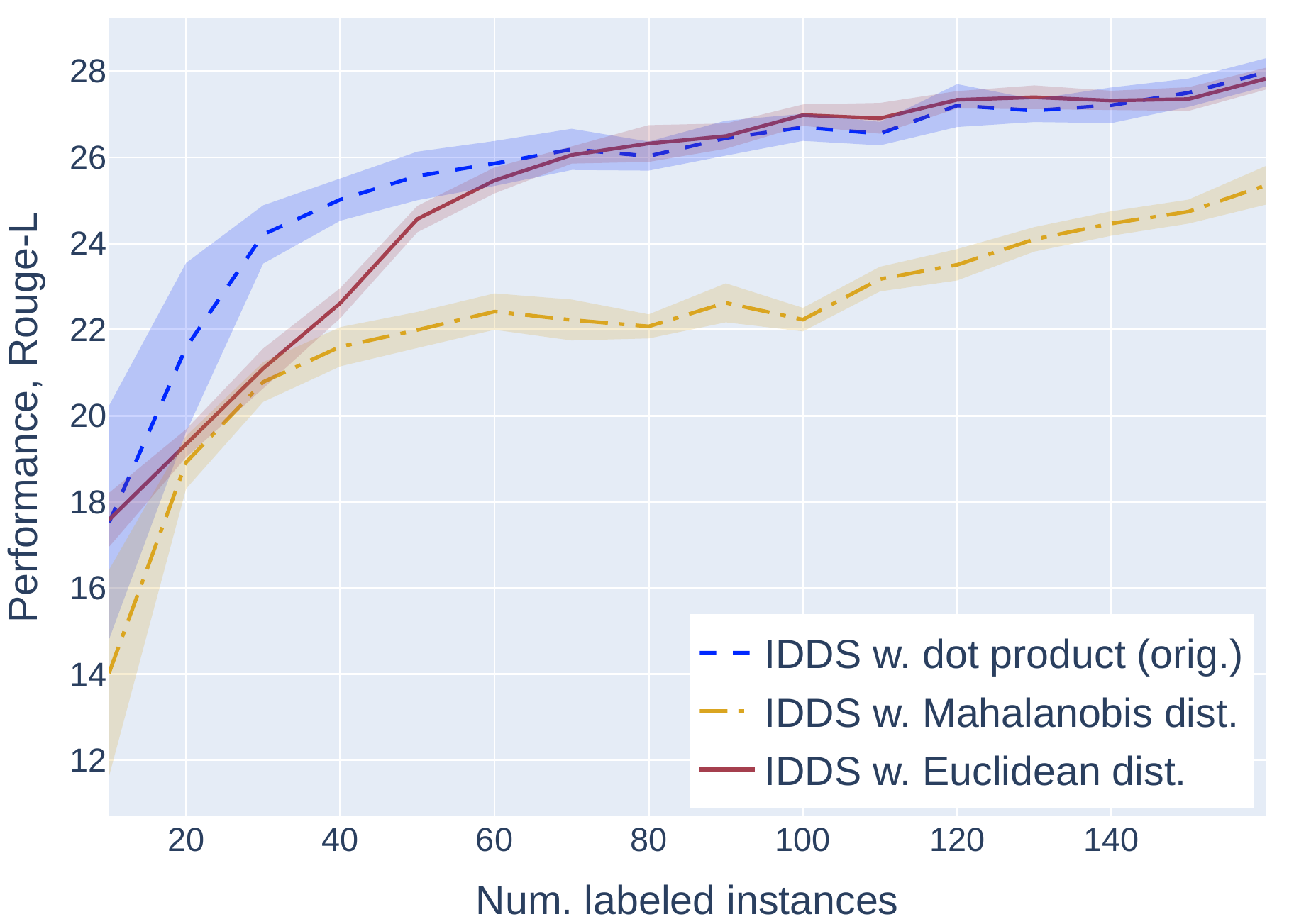} c) ROUGE-L}
    \vspace{0.3cm}
    \end{minipage}

    \vspace{-0.2cm}
    \caption{The performance of IDDS with different similarity functions with BART-base on AESLC.}
    \label{fig:as_sim_aeslc}
    \vspace{-0.4cm}
\end{figure*}

%% file: figures/ablation_2_sim_functions_wikihow.tex
\begin{figure*}[h!]
    \footnotesize
    \centering
    \begin{minipage}[ht]{0.32\linewidth}
    \vspace{-0.3cm}
    \center{\includegraphics[width=1\linewidth]{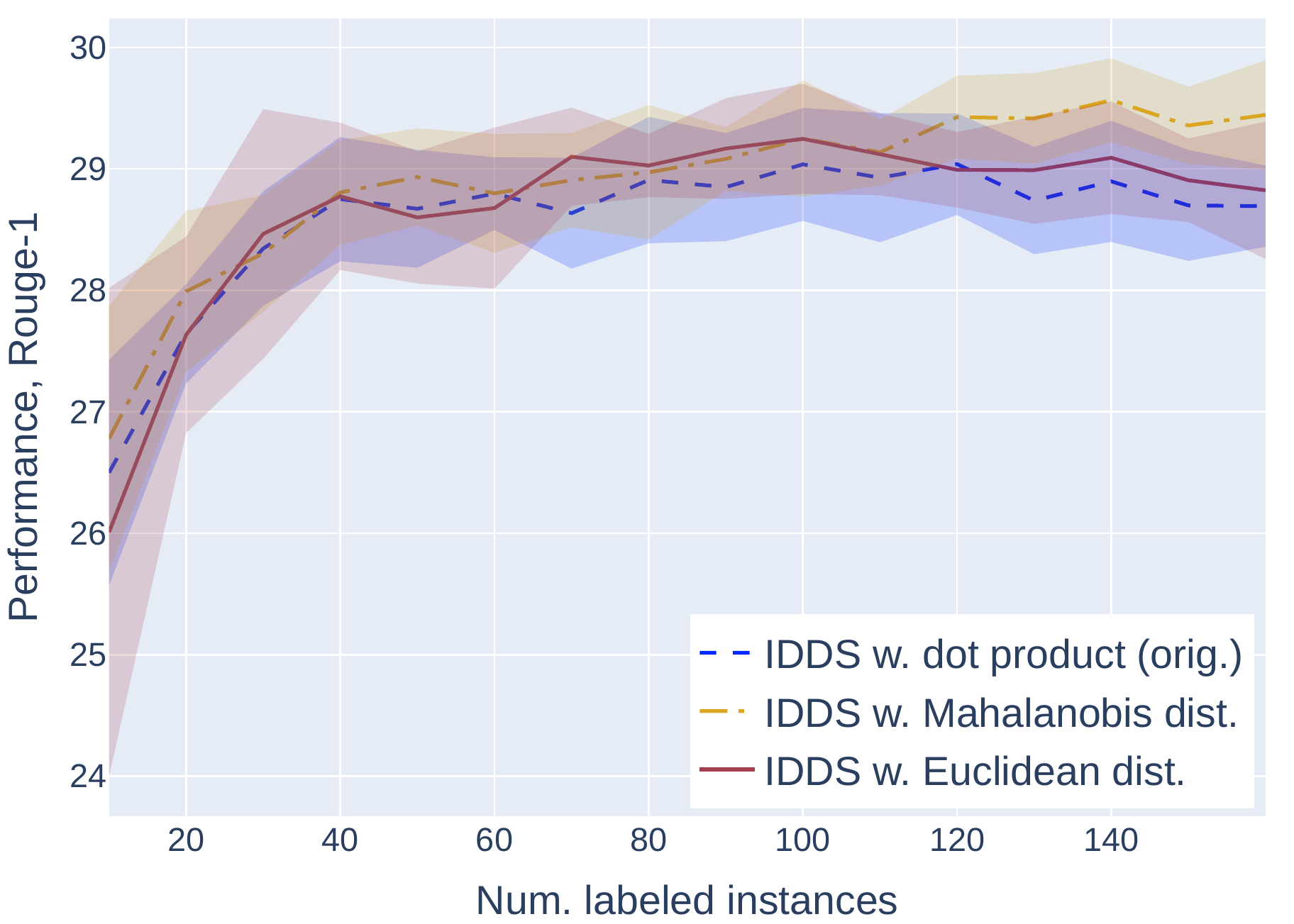} a) ROUGE-1}
    \end{minipage}
    \hspace{0.1cm}
    \begin{minipage}[ht]{0.32\linewidth}
    \center{\includegraphics[width=1\linewidth]{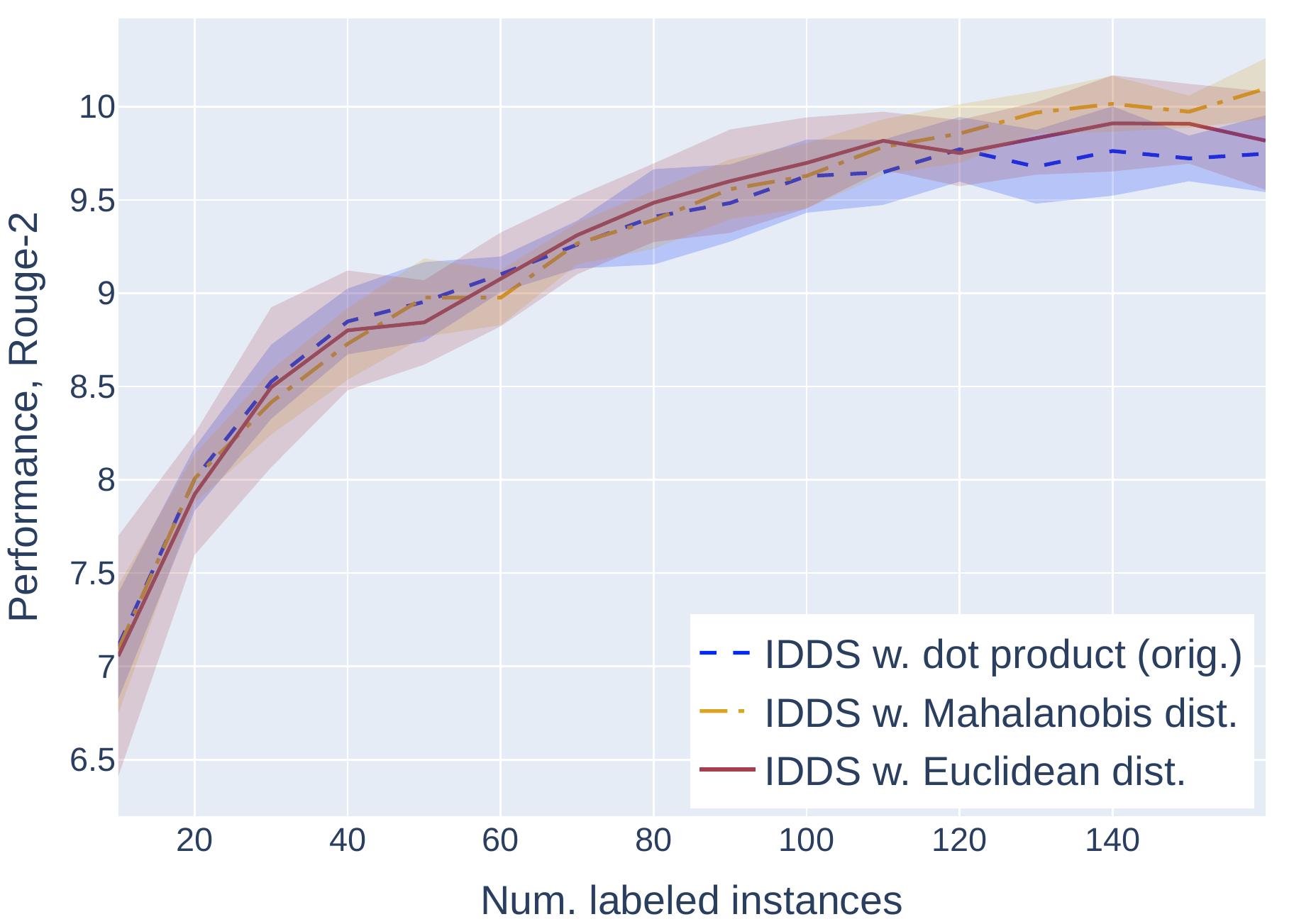} b) ROUGE-2}
    \vspace{0.3cm}
    \end{minipage}
    \hspace{0.1cm}
    \begin{minipage}[ht]{0.32\linewidth}
    \center{\includegraphics[width=1\linewidth]{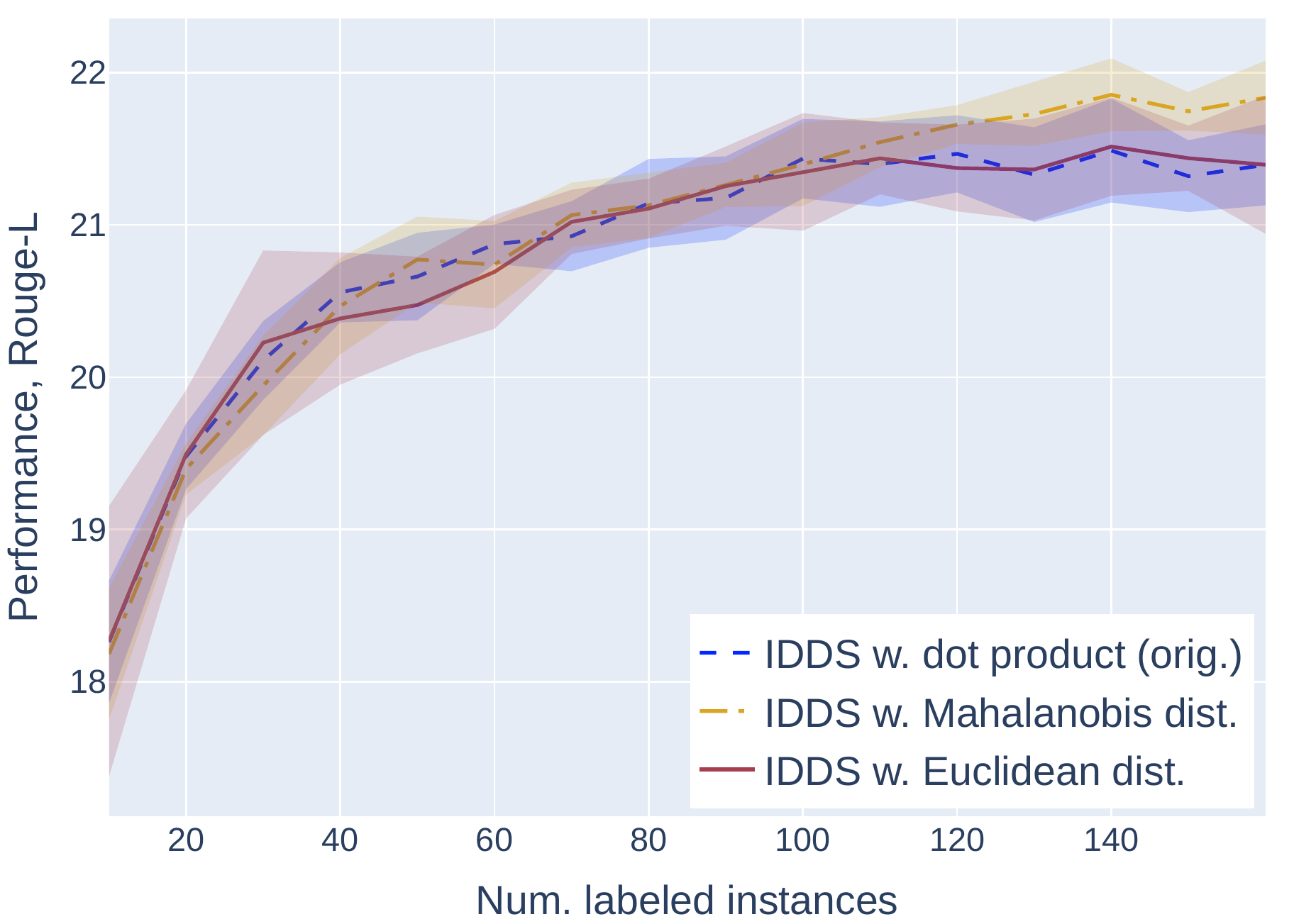} c) ROUGE-L}
    \vspace{0.3cm}
    \end{minipage}

    \vspace{-0.2cm}
    \caption{The performance of IDDS with different similarity functions with BART-base on WikiHow.}
    \label{fig:as_sim_wiki}
    \vspace{-0.4cm}
\end{figure*}

%% file: figures/ablation_2_normalization_aeslc.tex
\begin{figure*}[ht]
    \footnotesize
    \centering
    \begin{minipage}[ht]{0.32\linewidth}
    \vspace{-0.3cm}
    \center{\includegraphics[width=1\linewidth]{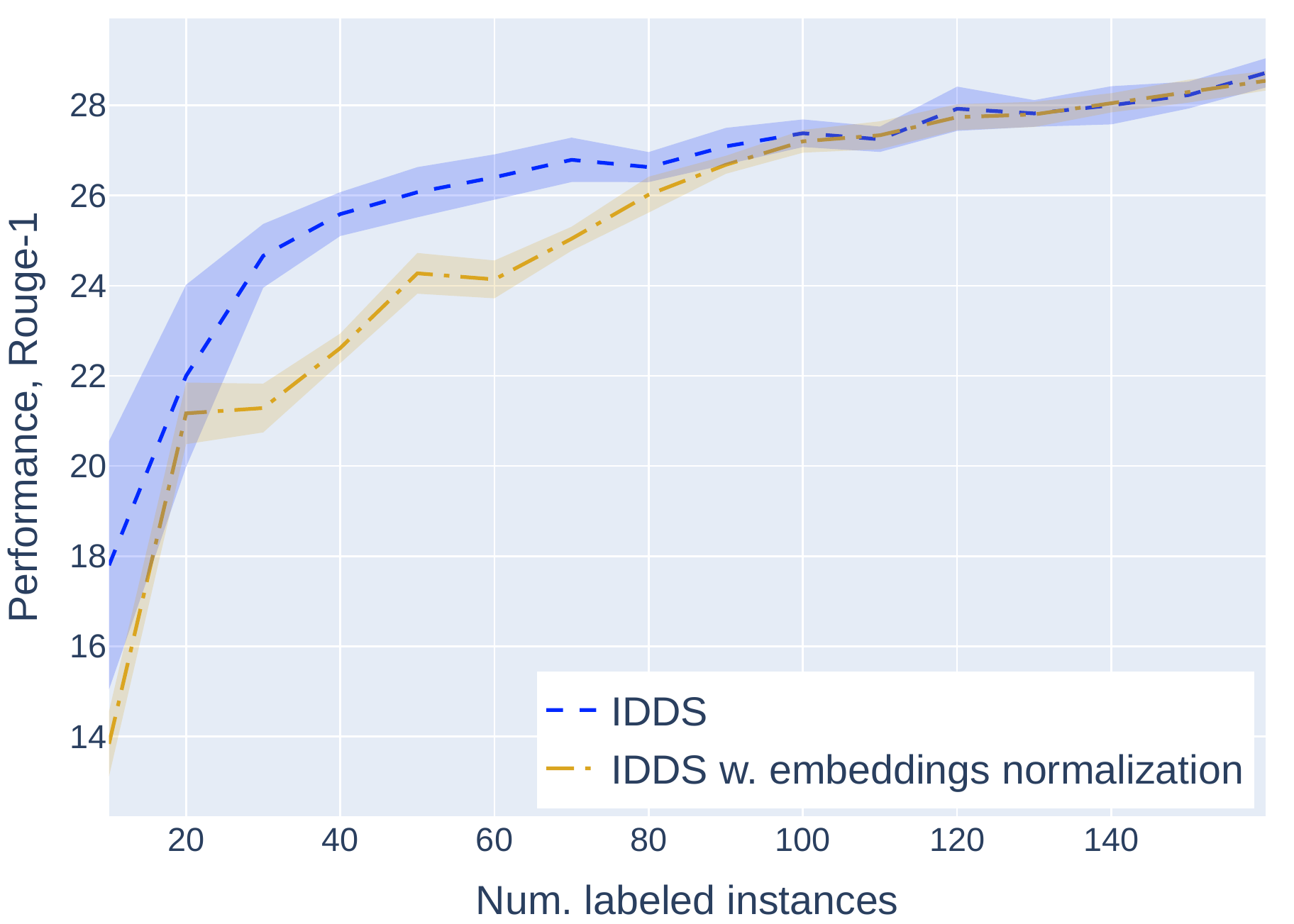} a) ROUGE-1}
    \end{minipage}
    \hspace{0.1cm}
    \begin{minipage}[ht]{0.32\linewidth}
    \center{\includegraphics[width=1\linewidth]{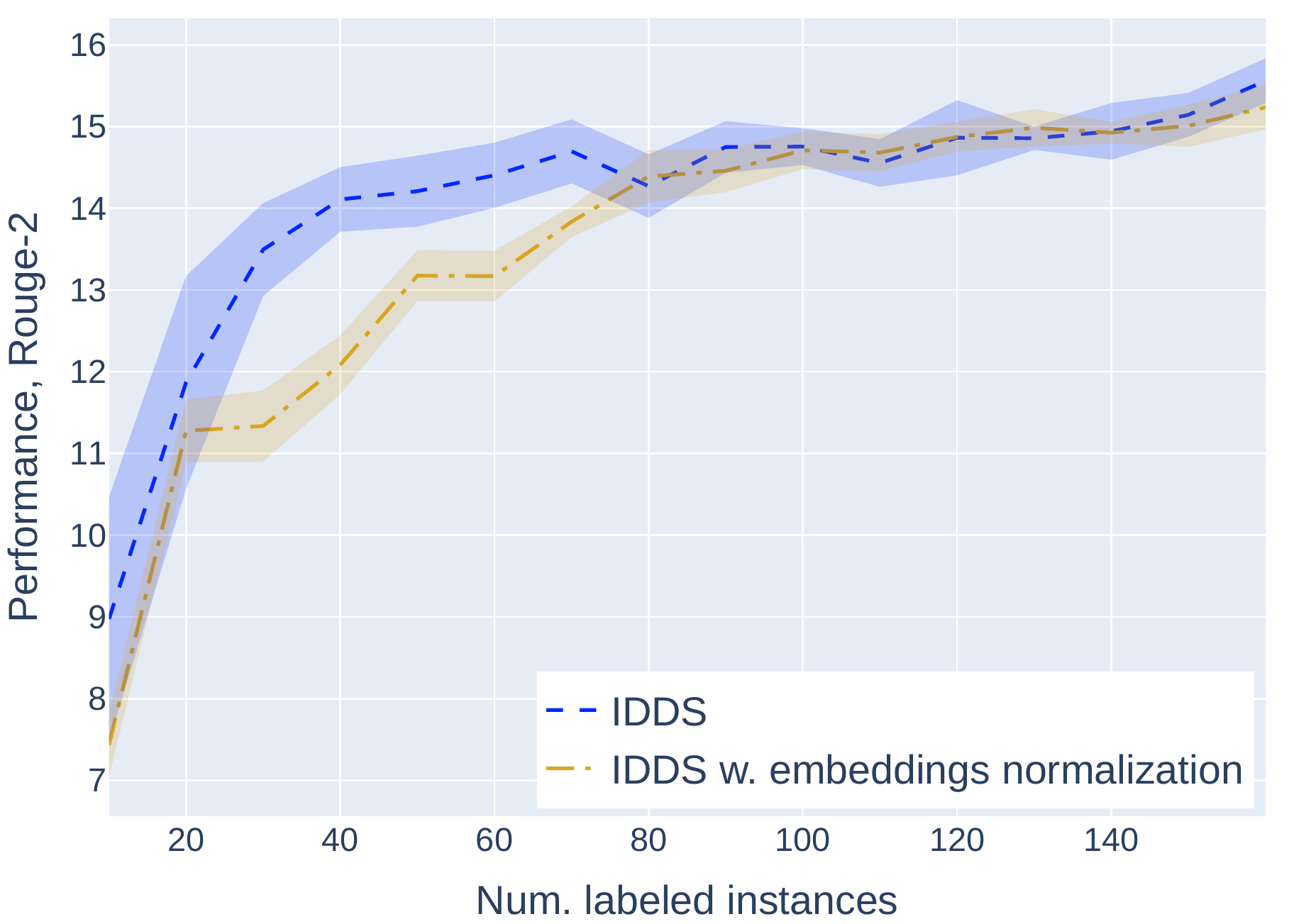} b) ROUGE-2}
    \vspace{0.3cm}
    \end{minipage}
    \hspace{0.1cm}
    \begin{minipage}[ht]{0.32\linewidth}
    \center{\includegraphics[width=1\linewidth]{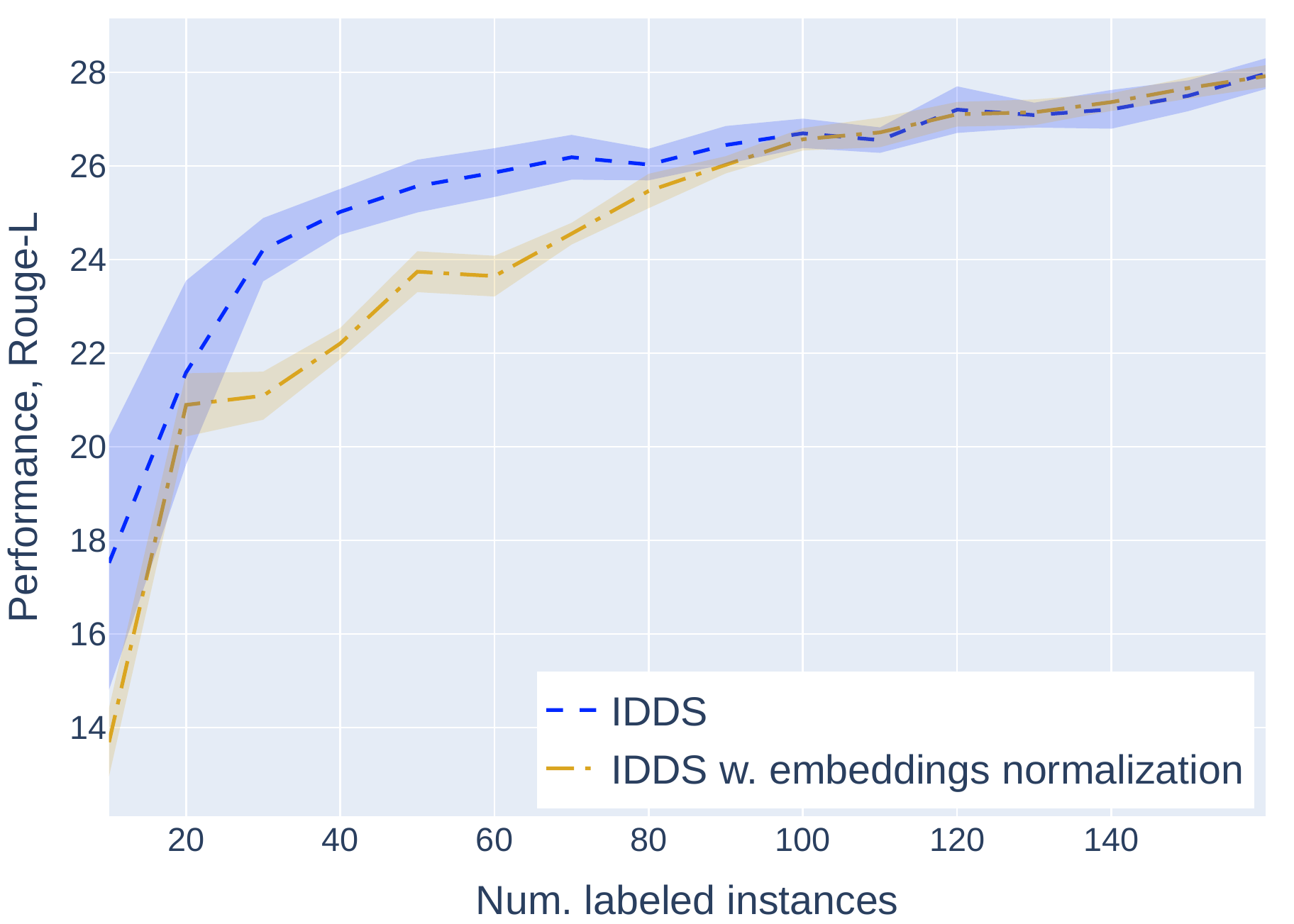} c) ROUGE-L}
    \vspace{0.3cm}
    \end{minipage}
    
    \vspace{-0.2cm}
    \caption{The performance of the BART-base model with the standard IDDS strategy compared with its modification when embeddings are normalized on AESLC.}
    \label{fig:as_norm_aeslc}
\end{figure*}

%% file: figures/ablation_2_normalization_pubmed.tex
\begin{figure*}[ht!]
    \footnotesize
    \centering
    \begin{minipage}[ht]{0.32\linewidth}
    \vspace{-0.3cm}
    \center{\includegraphics[width=1\linewidth]{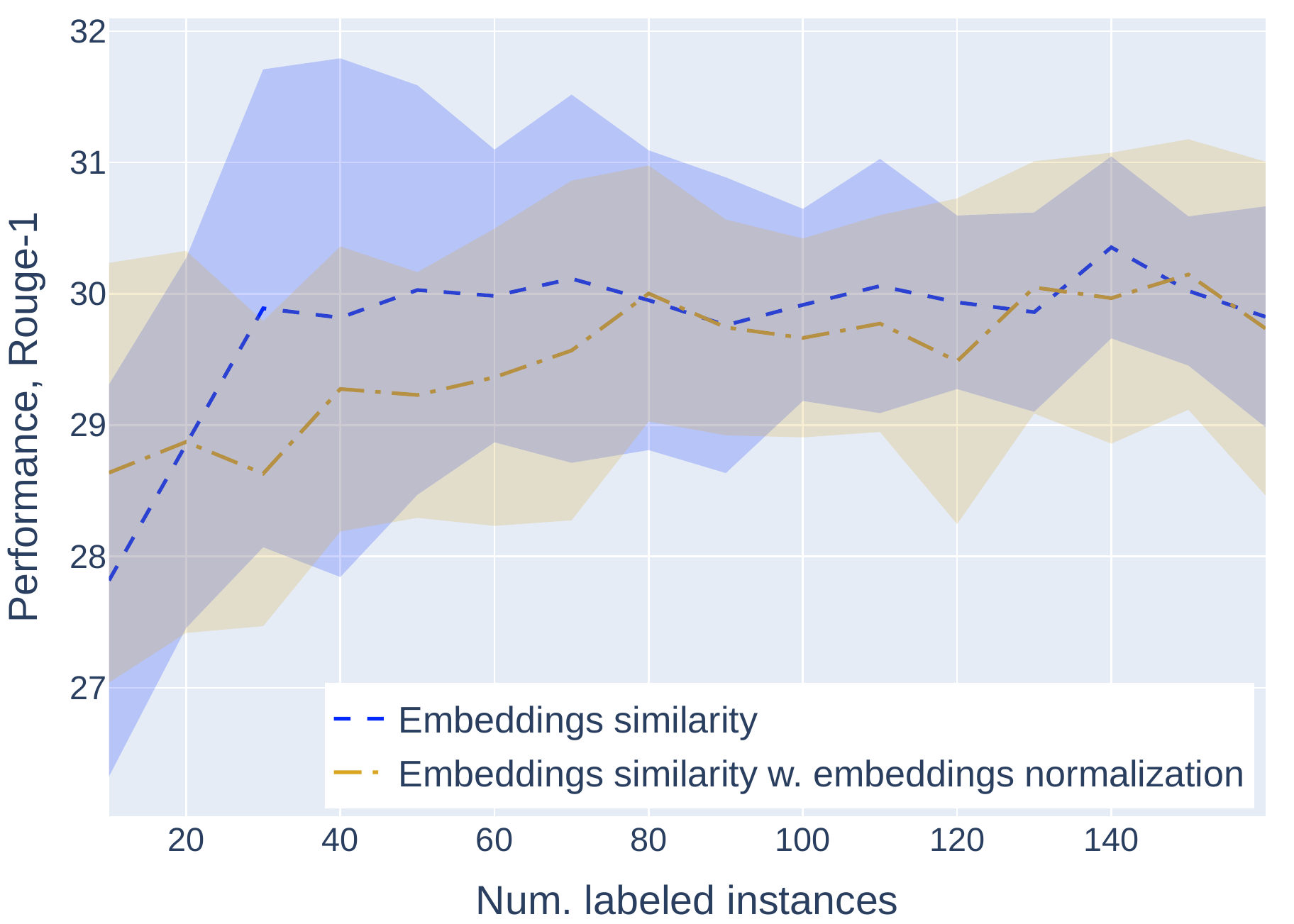} a) ROUGE-1}
    \end{minipage}
    \hspace{0.1cm}
    \begin{minipage}[ht]{0.32\linewidth}
    \center{\includegraphics[width=1\linewidth]{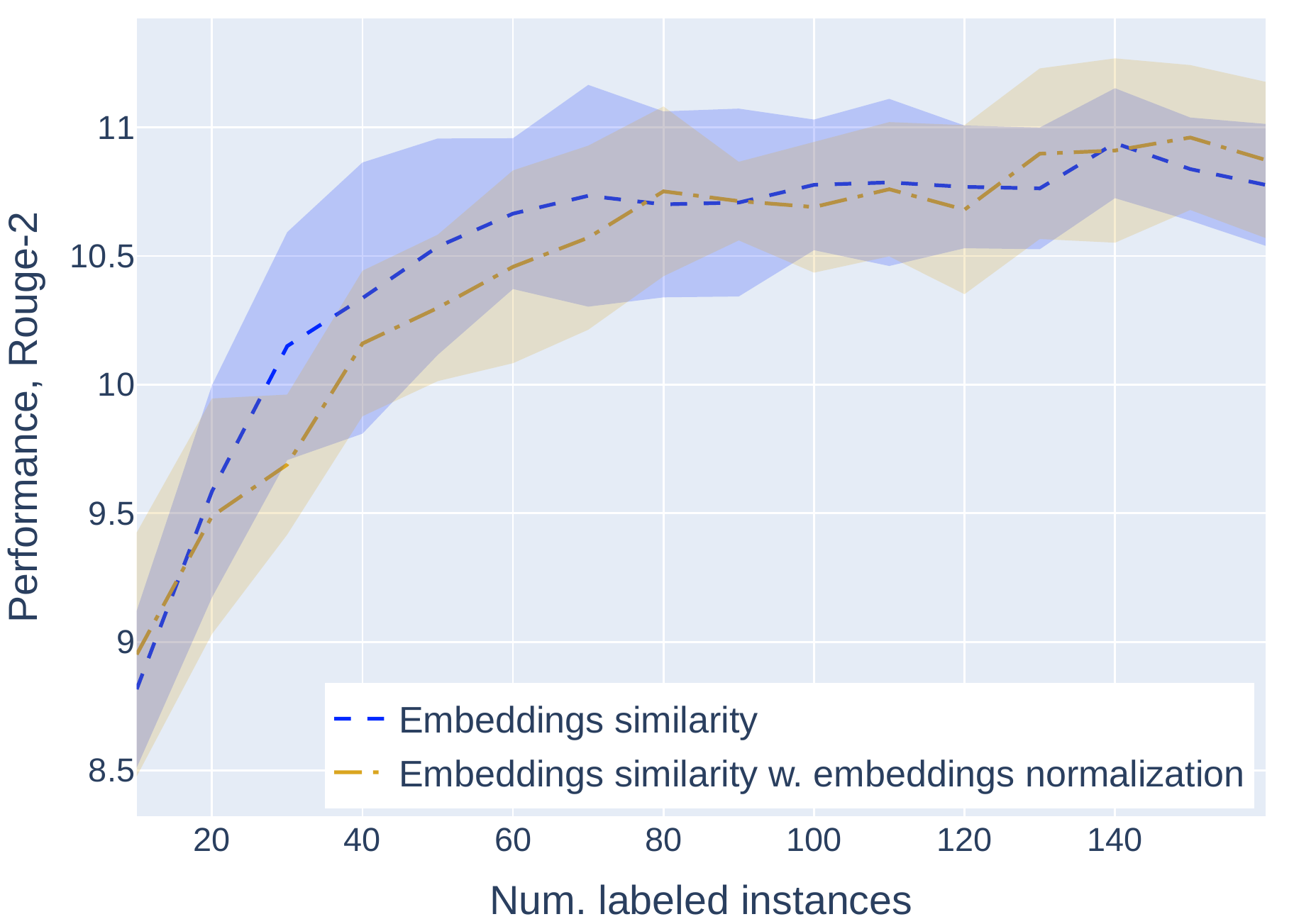} b) ROUGE-2}
    \vspace{0.3cm}
    \end{minipage}
    \hspace{0.1cm}
    \begin{minipage}[ht]{0.32\linewidth}
    \center{\includegraphics[width=1\linewidth]{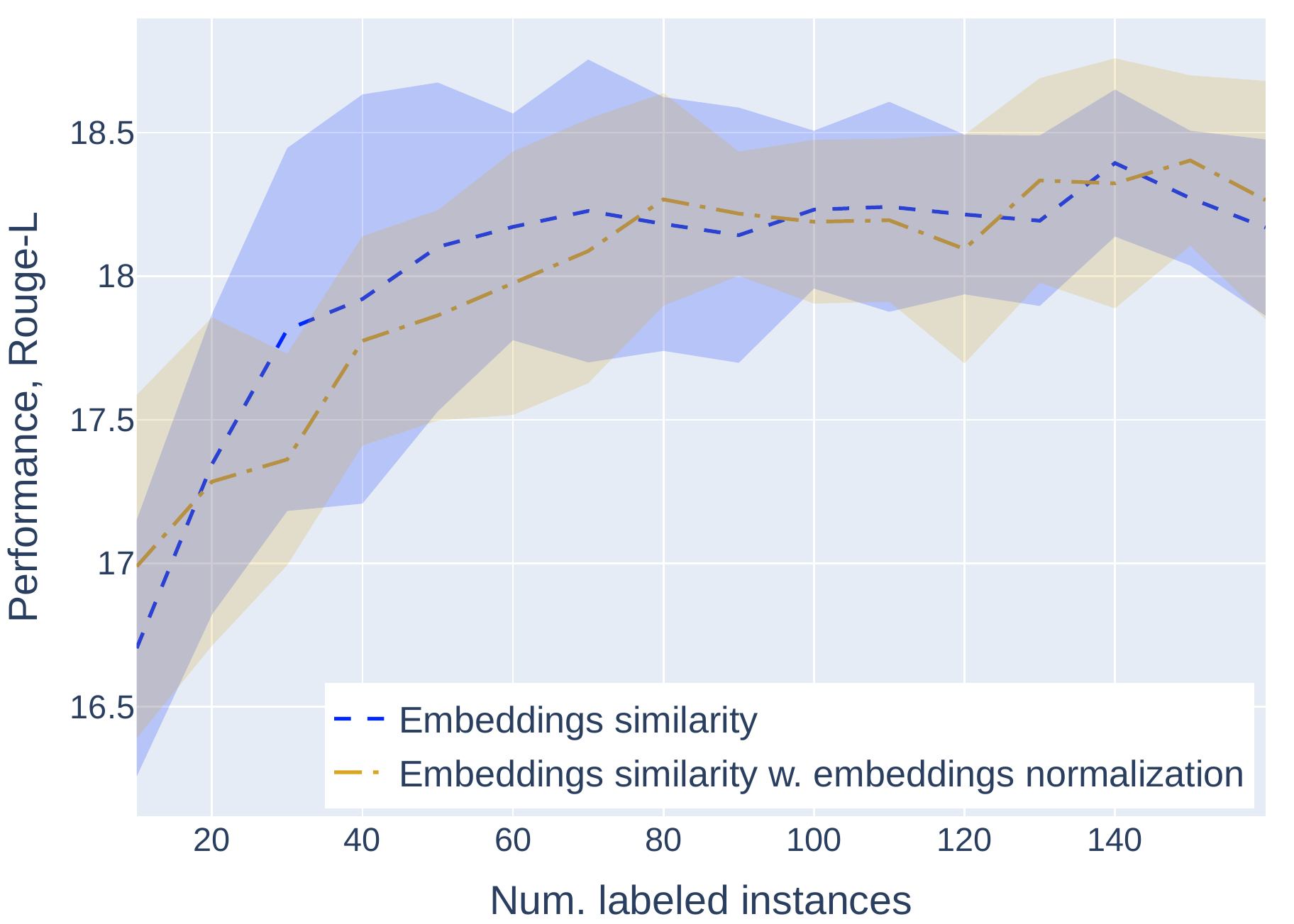} c) ROUGE-L}
    \vspace{0.3cm}
    \end{minipage}
    
    \vspace{-0.2cm}
    \caption{The performance of the BART-base model with the standard IDDS strategy compared with its modification when embeddings are normalized on PubMed.}
    \label{fig:as_norm_pubmed}
    \vspace{-0.4cm}
\end{figure*}

%% file: figures/ablation_3_lambda_aeslc.tex
\begin{figure*}[ht!]
    \footnotesize
    \centering
    \begin{minipage}[ht]{0.32\linewidth}
    \vspace{-0.3cm}
    \center{\includegraphics[width=1\linewidth]{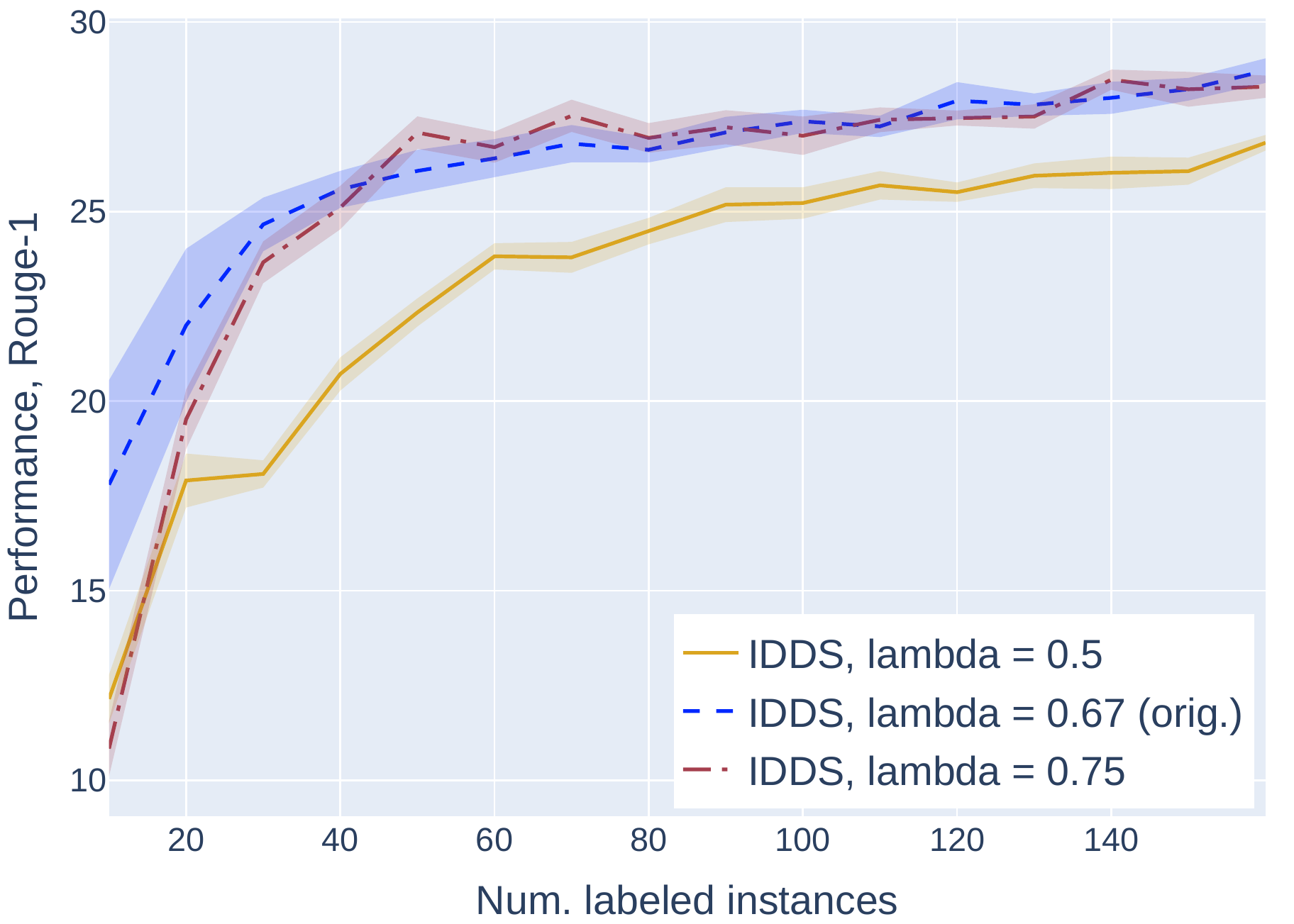} a) ROUGE-1}
    \end{minipage}
    \hspace{0.1cm}
    \begin{minipage}[ht]{0.32\linewidth}
    \center{\includegraphics[width=1\linewidth]{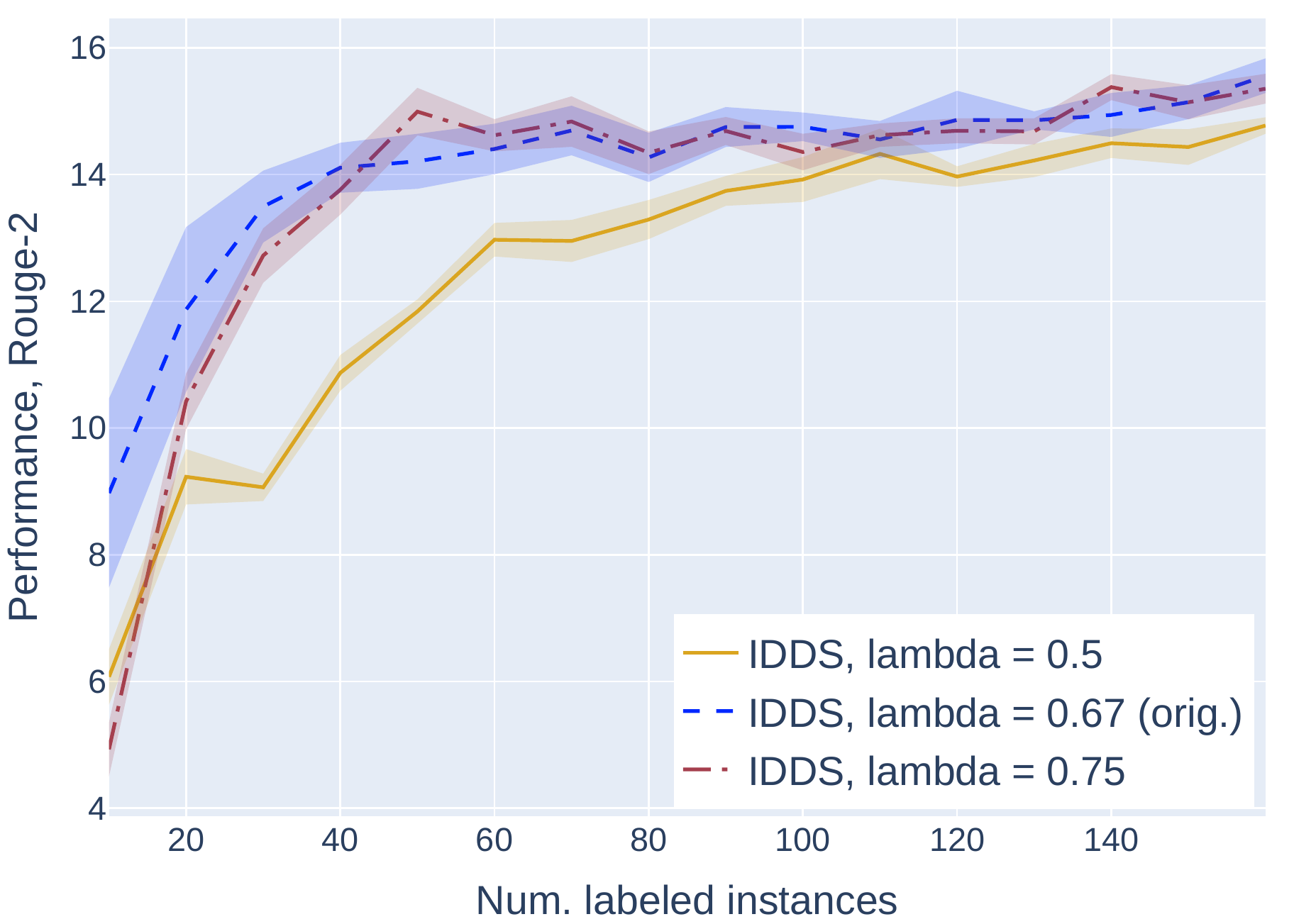} b) ROUGE-2}
    \vspace{0.3cm}
    \end{minipage}
    \hspace{0.1cm}
    \begin{minipage}[ht]{0.32\linewidth}
    \center{\includegraphics[width=1\linewidth]{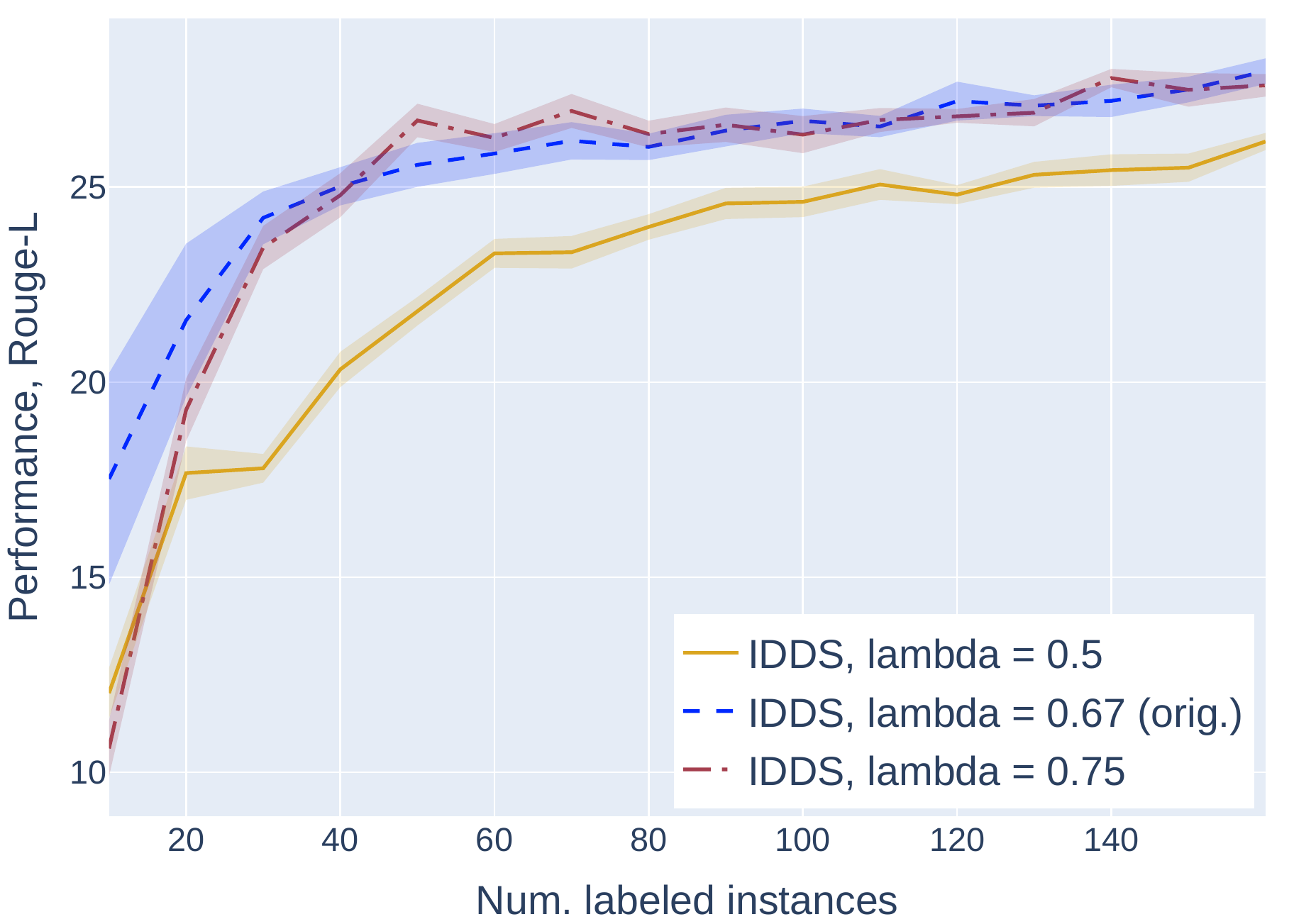} c) ROUGE-L}
    \vspace{0.3cm}
    \end{minipage}
    
    \vspace{-0.2cm}
    \caption{Ablation study for the hyperparameter $\lambda$ in the IDDS strategy with BART-base on AESLC.}
    \label{fig:as_lambda_aeslc}
    \vspace{-0.4cm}
\end{figure*}

%% file: figures/aeslc_table_lambda.tex
\begin{table}[ht!]
\centering
\scriptsize
\begin{tabular}{l|ccccc}
\toprule
\textbf{AL Strategy} &     \textbf{Iter. 0} &     \textbf{Iter. 5} &    \textbf{Iter. 10} &    \textbf{Iter. 15}  & \textbf{Average} \\
\midrule
$\lambda$ = 0.           &    9.08 / 4.8 / 8.79 &   19.6 / 10.87 / 19.29 &    22.6 / 12.58 / 22.1 &  23.68 / 13.32 / 23.23 &  21.25 / 11.72 / 20.88 \\
$\lambda$ = 0.33         &  15.77 / 7.67 / 15.46 &  22.47 / 12.18 / 22.07 &  23.98 / 13.54 / 23.51 &  24.68 / 13.81 / 24.21 &  23.19 / 12.88 / 22.78 \\
$\lambda$ = 0.5          &   12.15 / 6.07 / 12.03 &   23.82 / 12.97 / 23.3 &  25.69 / 14.33 / 25.06 &  26.81 / 14.77 / 26.17 &  23.84 / 12.94 / 23.31 \\
\textbf{$\lambda$ = 0.67 (orig.)} &   \textbf{17.8 / 8.97 / 17.52} &    \textbf{26.4 / 14.4 / 25.86} &  \textbf{27.25 / 14.55 / 26.55} &  \textbf{28.72 / 15.56 / 27.97} &   \textbf{26.7 / 14.43 / 26.07} \\
$\lambda$ = 0.75         &  10.84 / 4.93 / 10.61 &   \textbf{26.7 / 14.62 / 26.26} &  \textbf{27.42 / 14.62 / 26.72} &  28.29 / \textbf{15.36} / 27.61 &   \textbf{26.54 / 14.31 / 26.0} \\
$\lambda$ = 0.83         &   16.47 / 7.84 / 16.03 &  26.06 / \textbf{14.42} / 25.59 &  26.57 / 14.12 / 25.92 &    \textbf{28.7} / 15.22 / \textbf{28.0} &   26.0 / 13.93 / 25.46 \\
$\lambda$ = 1.           &   16.41 / \textbf{8.66} / 16.23 &   25.2 / 13.66 / 24.72 &   26.74 / 14.4 / 26.04 &  27.44 / 14.66 / 26.67 &  25.73 / 13.81 / 25.16 \\
\bottomrule
\end{tabular}
\vspace{-0.1cm}

\caption{ROUGE scores on AL iterations for different values of the $lambda$ hyperparameter in IDDS. We select with \textbf{bold} the largest values w.r.t. the confidence intervals.}
\label{tab:lamb_ae}

\vspace{-0.5cm}
\end{table}

%% file: figures/ablation_4_mmr_aeslc.tex
\begin{figure*}[ht!]
    \footnotesize
    \centering
    \begin{minipage}[ht]{0.32\linewidth}
    \vspace{-0.3cm}
    \center{\includegraphics[width=1\linewidth]{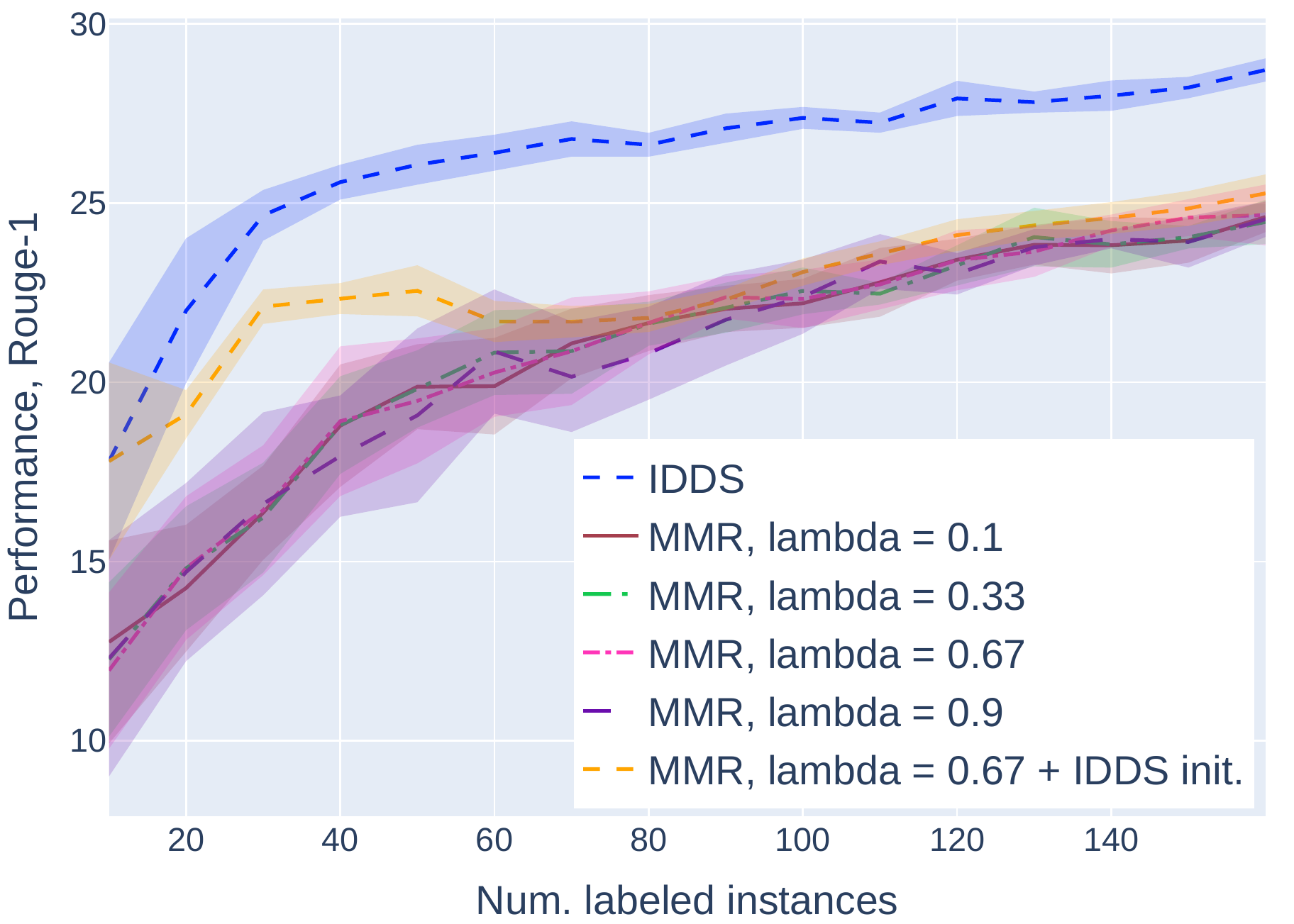} a) ROUGE-1}
    \end{minipage}
    \hspace{0.1cm}
    \begin{minipage}[ht]{0.32\linewidth}
    \center{\includegraphics[width=1\linewidth]{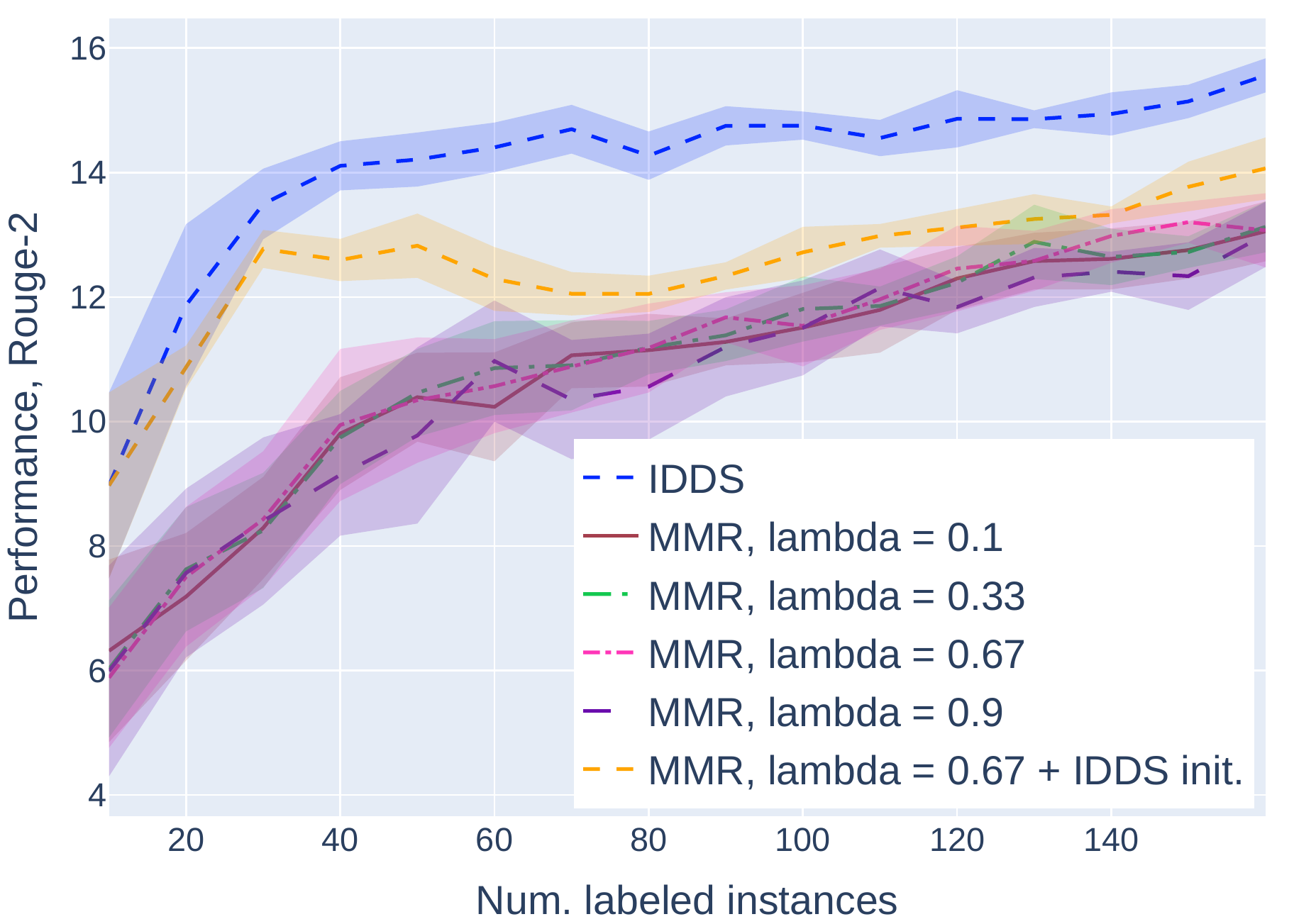} b) ROUGE-2}
    \vspace{0.3cm}
    \end{minipage}
    \hspace{0.1cm}
    \begin{minipage}[ht]{0.32\linewidth}
    \center{\includegraphics[width=1\linewidth]{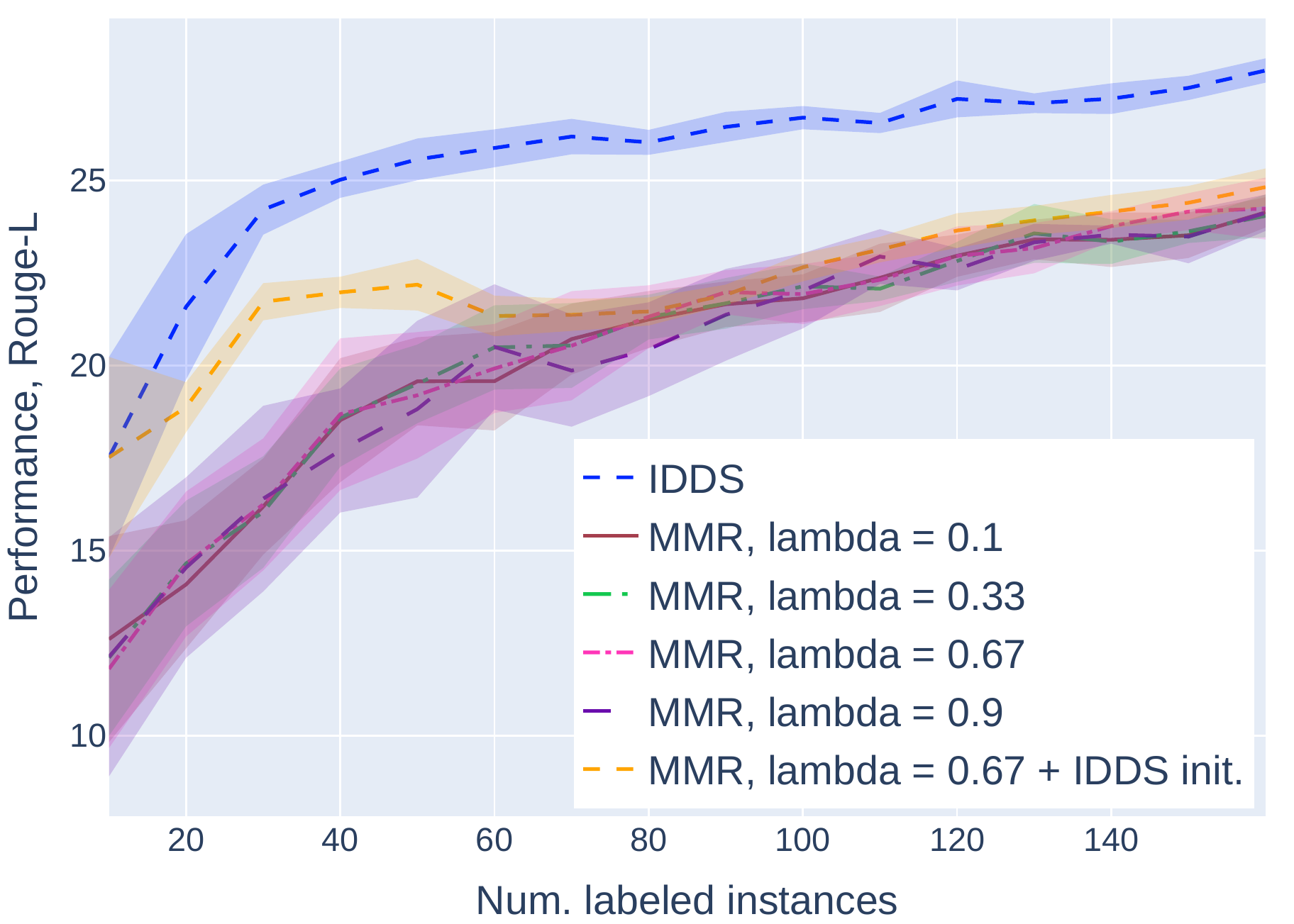} c) ROUGE-L}
    \vspace{0.3cm}
    \end{minipage}
    
    \vspace{-0.2cm}
    \caption{Comparison of IDDS with the MMR-based strategy suggested in \cite{mmr_al} with BART-base on AESLC. We experiment with different $\lambda$ values in MMR and the initialization schemes.}
    \label{fig:as_mmr_aeslc}
    \vspace{-0.4cm}
\end{figure*}

%% file: figures/ablation_4_aggregation_aeslc.tex
\begin{figure*}[ht!]
    \footnotesize
    \centering
    \begin{minipage}[ht]{0.32\linewidth}
    \vspace{-0.3cm}
    \center{\includegraphics[width=1\linewidth]{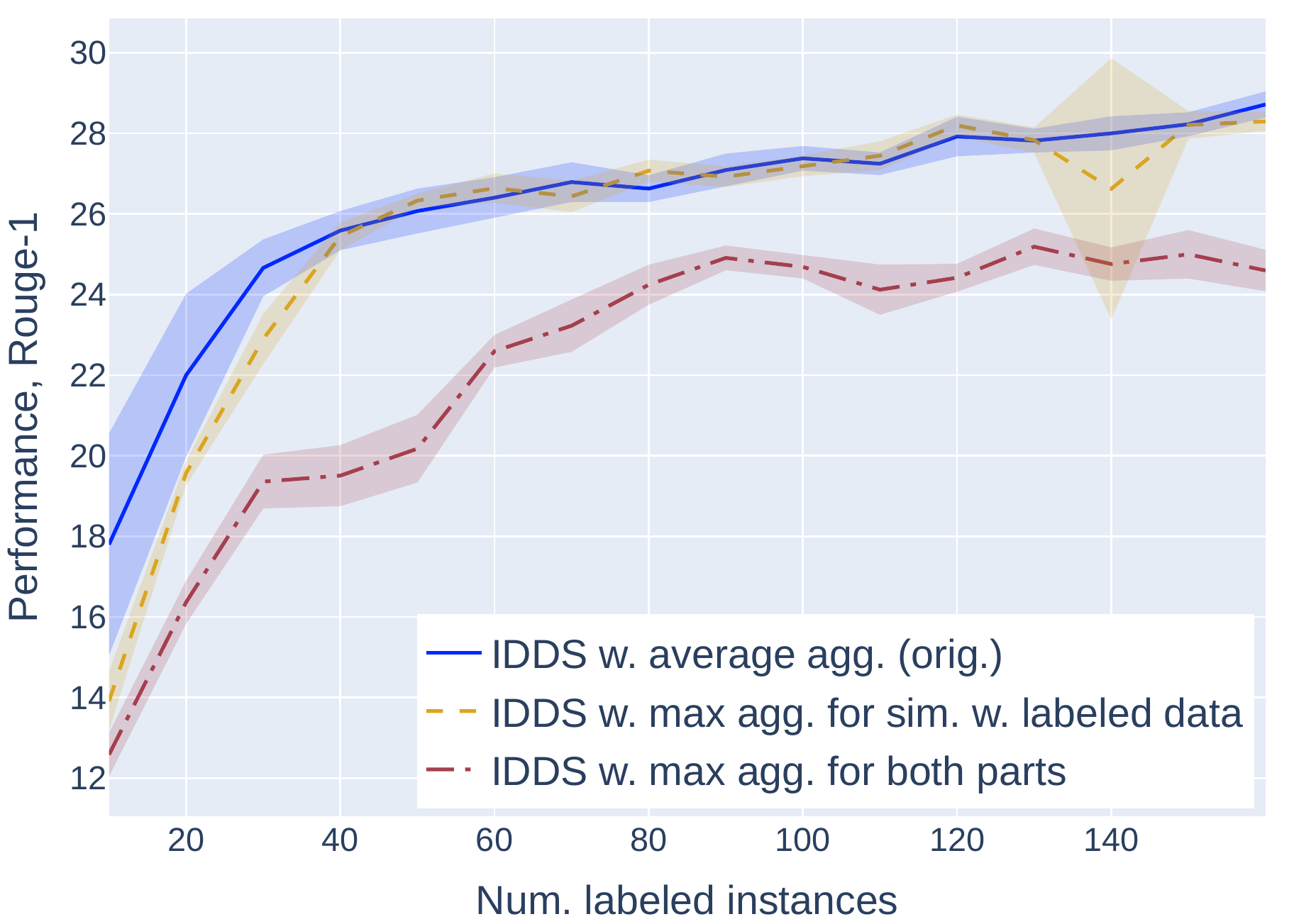} a) ROUGE-1}
    \end{minipage}
    \hspace{0.1cm}
    \begin{minipage}[ht]{0.32\linewidth}
    \center{\includegraphics[width=1\linewidth]{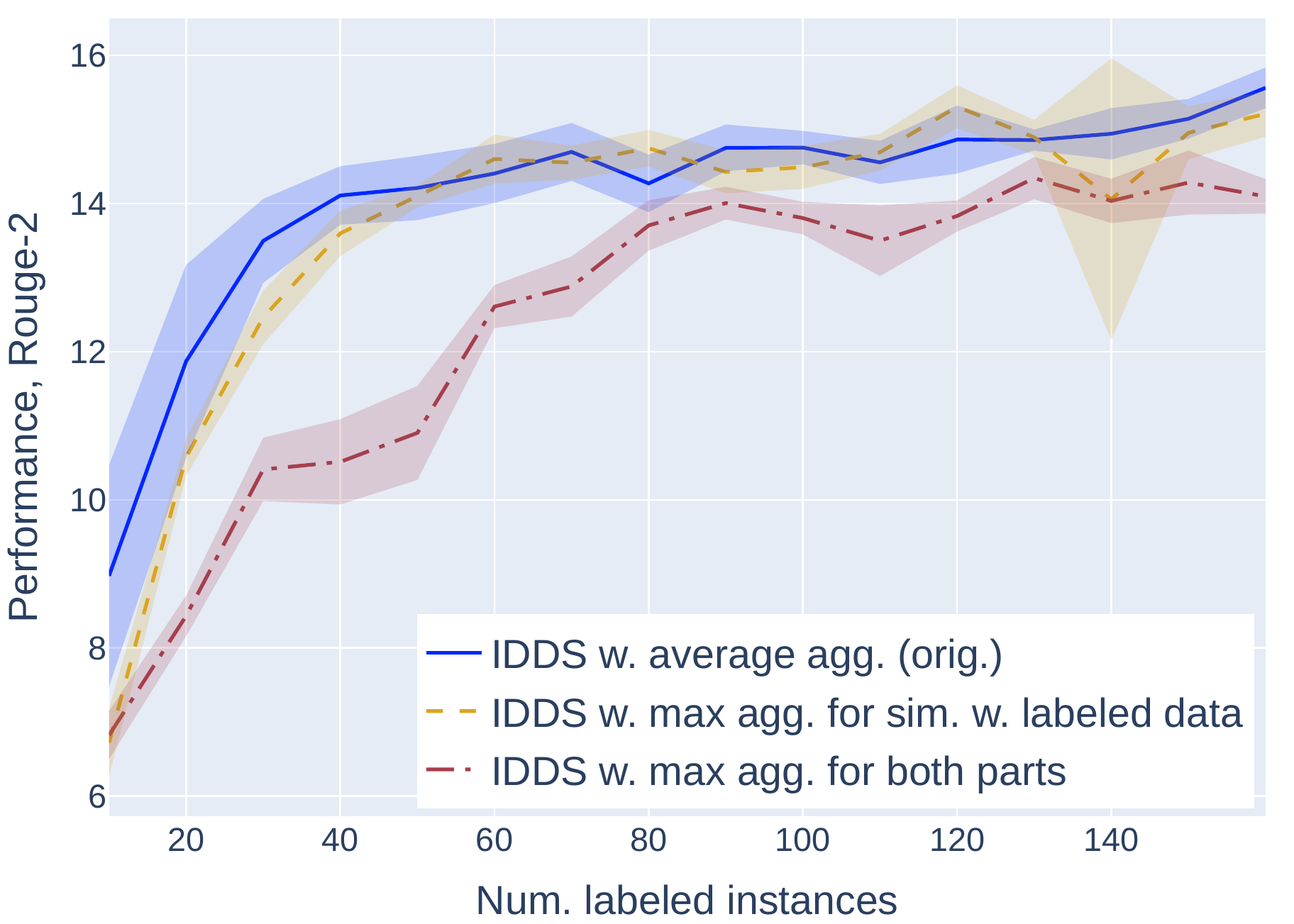} b) ROUGE-2}
    \vspace{0.3cm}
    \end{minipage}
    \hspace{0.1cm}
    \begin{minipage}[ht]{0.32\linewidth}
    \center{\includegraphics[width=1\linewidth]{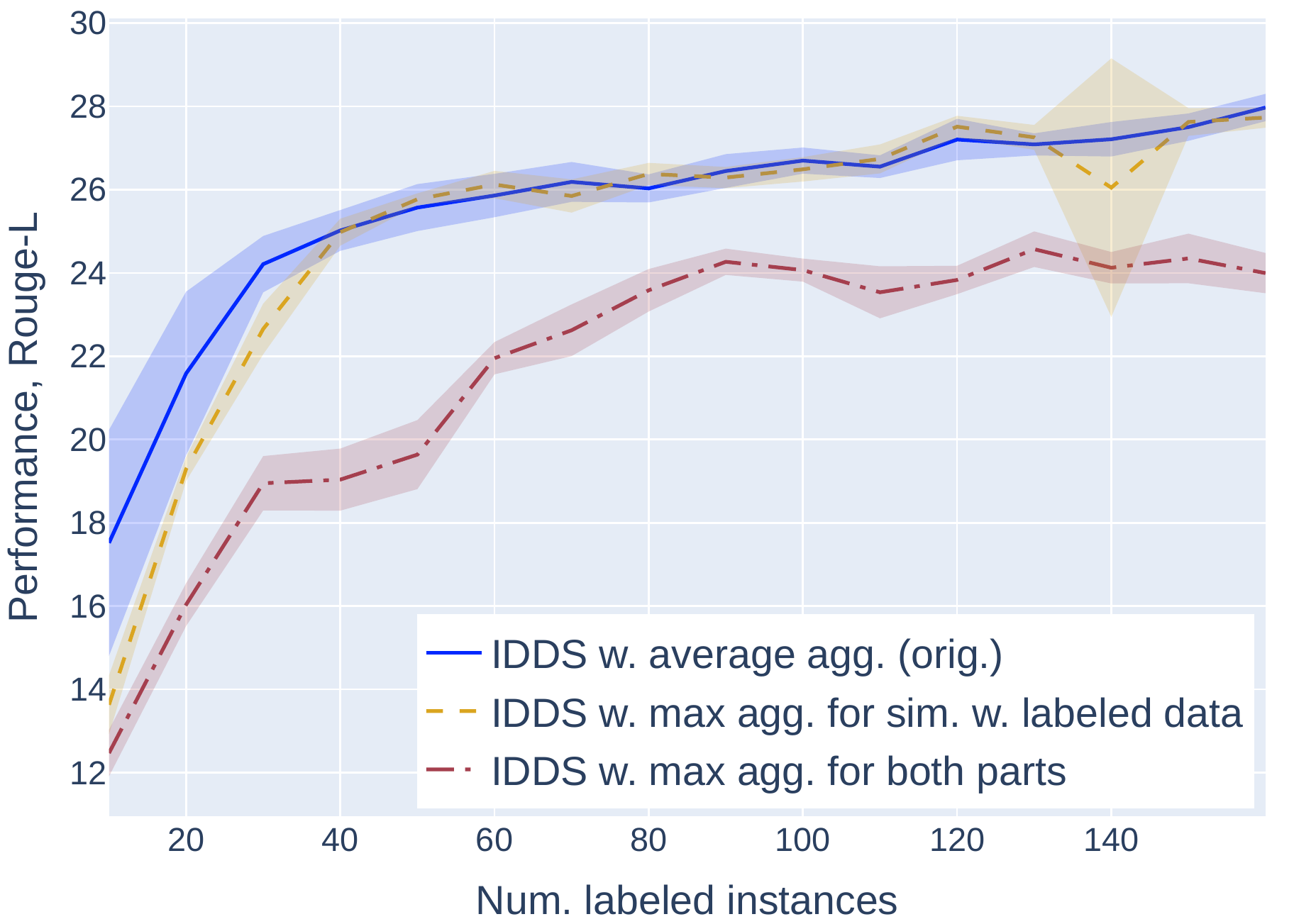} c) ROUGE-L}
    \vspace{0.3cm}
    \end{minipage}
    
    \vspace{-0.2cm}
    \caption{Comparison of the \textit{average} and \textit{maximum} aggregation functions in IDDS with BART-base on AESLC.}
    \label{fig:as_agg_aeslc}
    \vspace{-0.4cm}
\end{figure*}

%% file: figures/ablation_4_aggregation_wikihow.tex
\begin{figure*}[ht!]
    \footnotesize
    \centering
    \begin{minipage}[ht]{0.32\linewidth}
    \vspace{-0.3cm}
    \center{\includegraphics[width=1\linewidth]{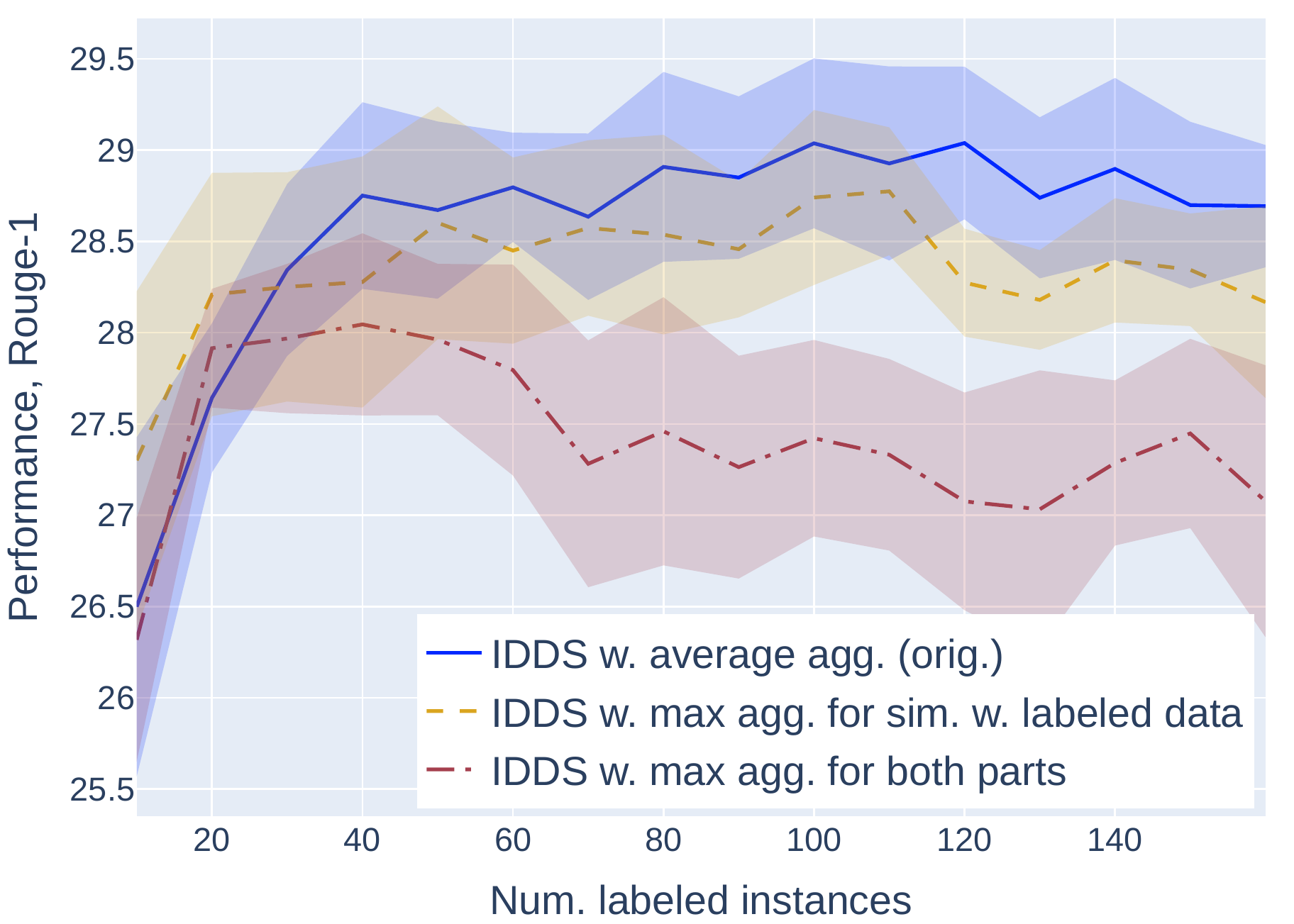} a) ROUGE-1}
    \end{minipage}
    \hspace{0.1cm}
    \begin{minipage}[ht]{0.32\linewidth}
    \center{\includegraphics[width=1\linewidth]{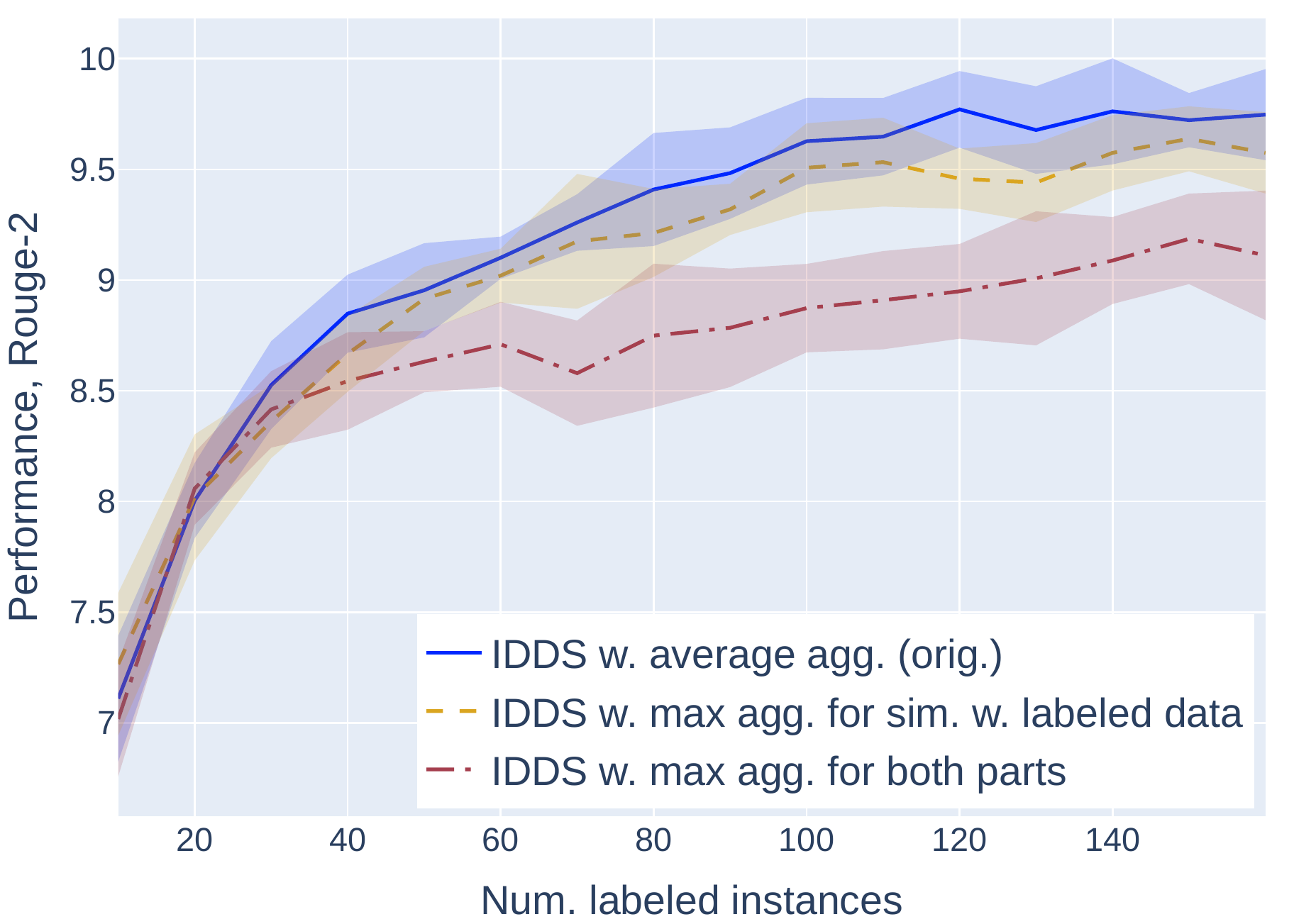} b) ROUGE-2}
    \vspace{0.3cm}
    \end{minipage}
    \hspace{0.1cm}
    \begin{minipage}[ht]{0.32\linewidth}
    \center{\includegraphics[width=1\linewidth]{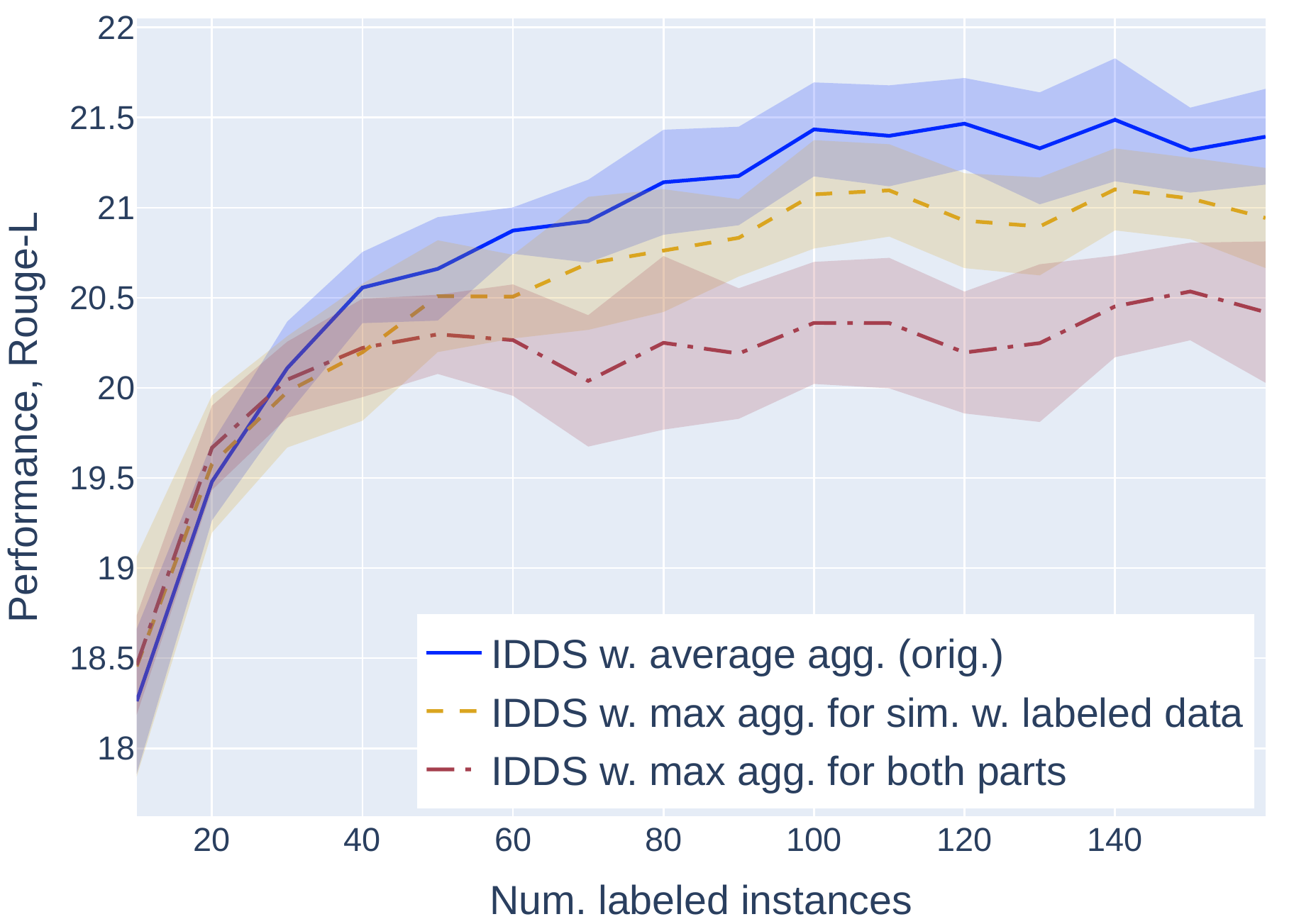} c) ROUGE-L}
    \vspace{0.3cm}
    \end{minipage}
    
    \vspace{-0.2cm}
    \caption{Comparison of the \textit{average} and \textit{maximum} aggregation functions in IDDS with BART-base on WikiHow.}
    \label{fig:as_agg_wiki}
    \vspace{-0.4cm}
\end{figure*}

%% file: figures/gigaword_consistency_pl.tex
\begin{figure}[ht]
    \vspace{-0.1cm}
	\center{\includegraphics[width=0.55\linewidth]{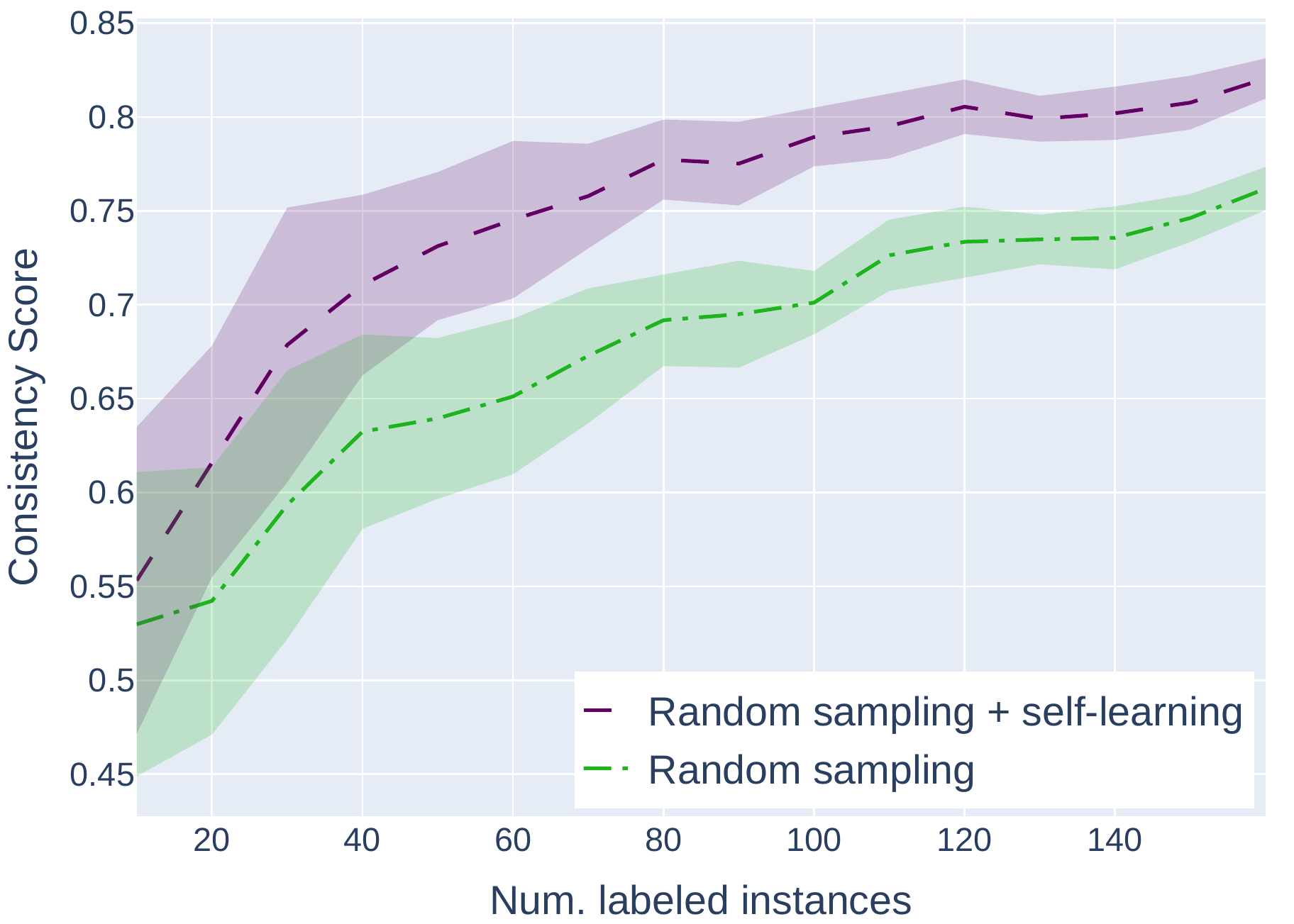}}
	\vspace{-0.1cm}
	\caption{The consistency score calculated via SummaC with BART-base on Gigaword  without AL (random sampling) with and without self-learning.}
	\label{fig:gigaword_consistency_pl}
	\vspace{-0.3cm}
	
\end{figure}

%% file: figures/aeslc_consistency_pl.tex
\begin{figure}[ht!]
    \vspace{-0.1cm}
	\center{\includegraphics[width=0.55\linewidth]{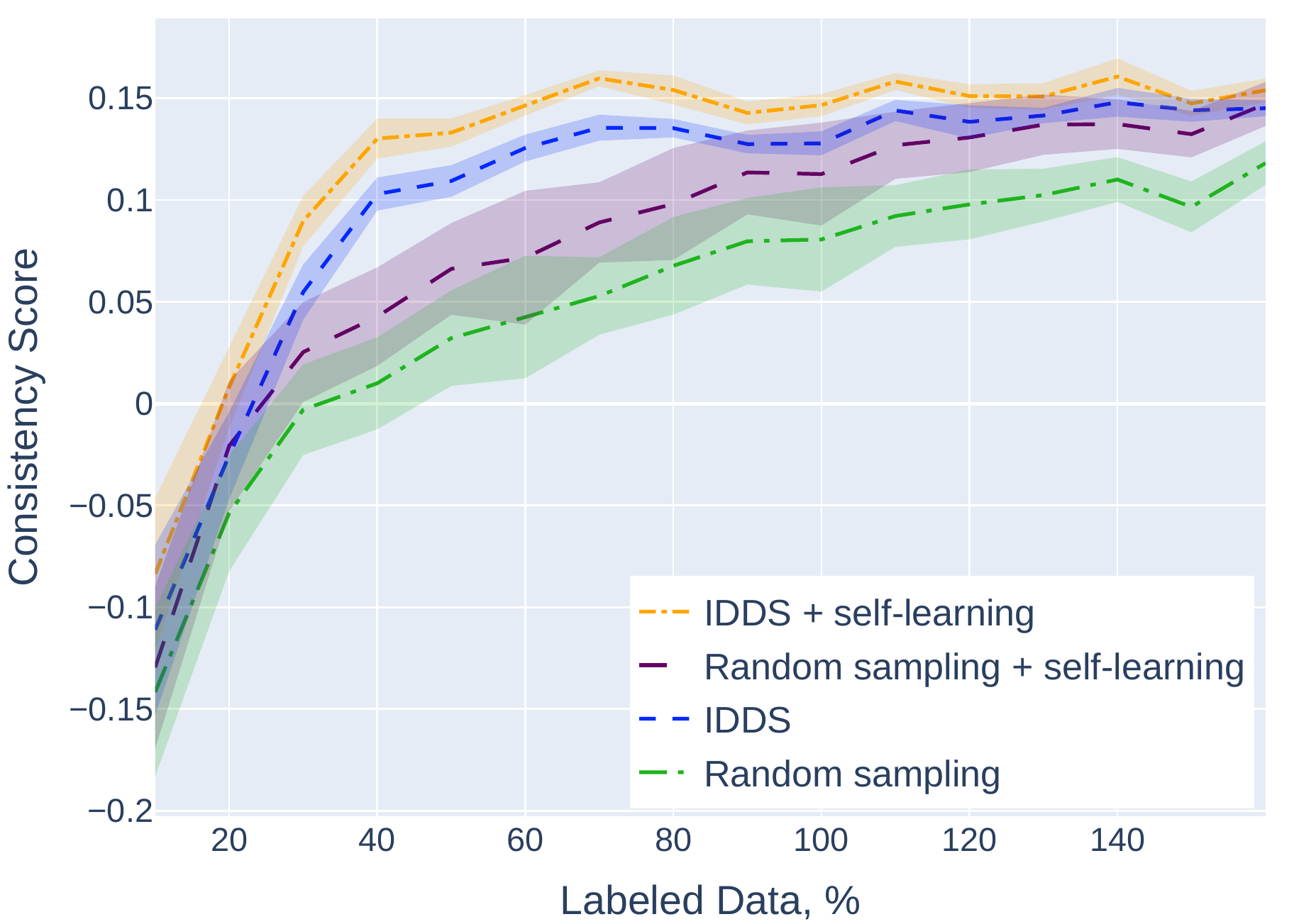}}
	\vspace{-0.1cm}
	\caption{The consistency score calculated via SummaC on the test sample on AESLC for BART-base with the IDDS and random sampling strategies with and without self-learning.}
	\label{fig:aeslc_consistency_pl}
	\vspace{-0.1cm}
	
\end{figure}

%% file: figures/wikihow_consistency_model_bart.tex
\begin{figure}[ht!]
    \vspace{-0.1cm}
	\center{\includegraphics[width=0.55\linewidth]{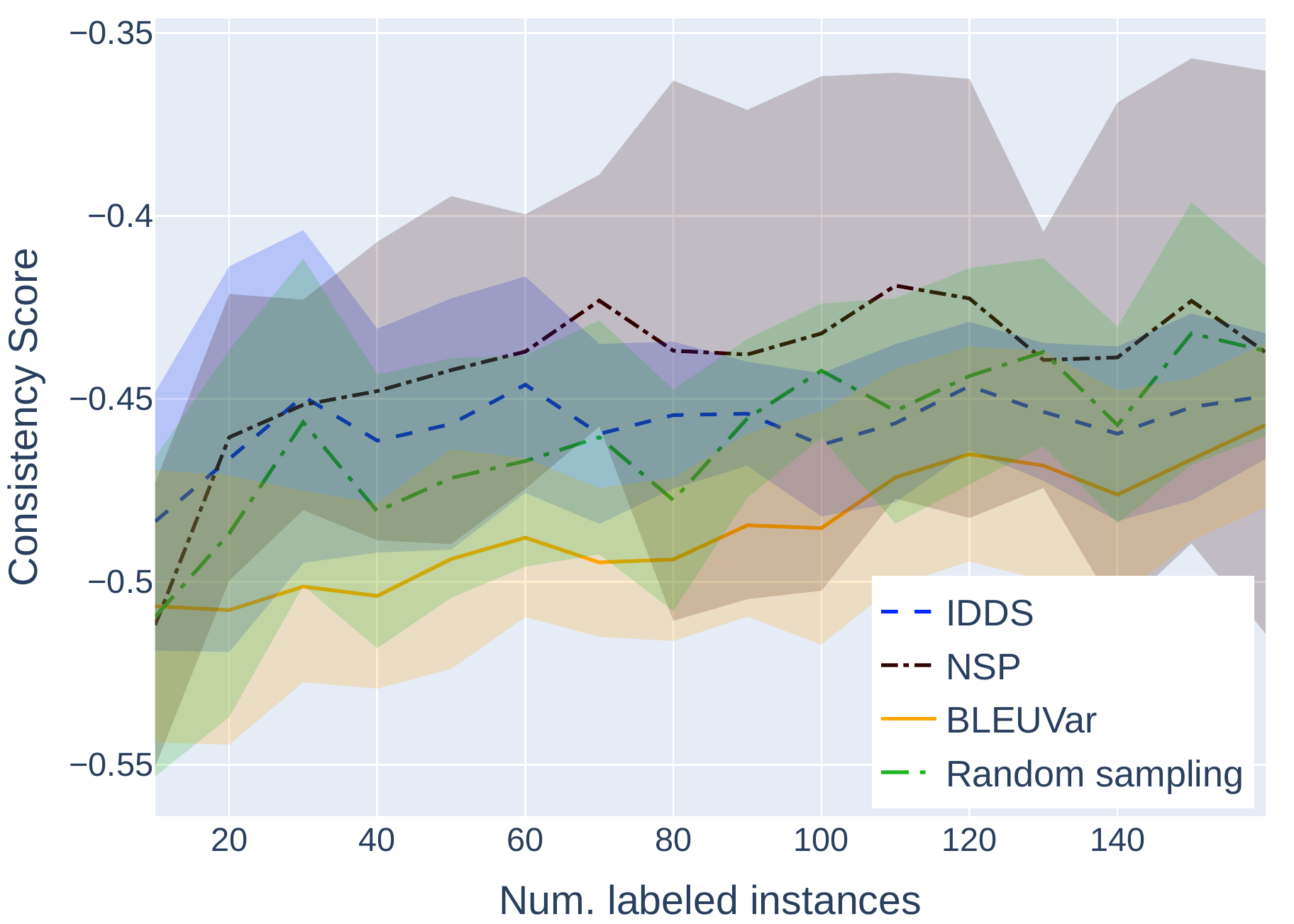}}
	\vspace{-0.1cm}
	\caption{The consistency score calculated via SummaC on the test subset of WikiHow for the BART-base model with various AL strategies.}
	\label{fig:wikihow_consistency_model_bart}
	\vspace{-0.1cm}
	
\end{figure}

%% file: figures/time_strategy_tex.tex
\begin{figure}[!ht]
    \vspace{-0.1cm}
	\center{\includegraphics[width=0.67 \linewidth]{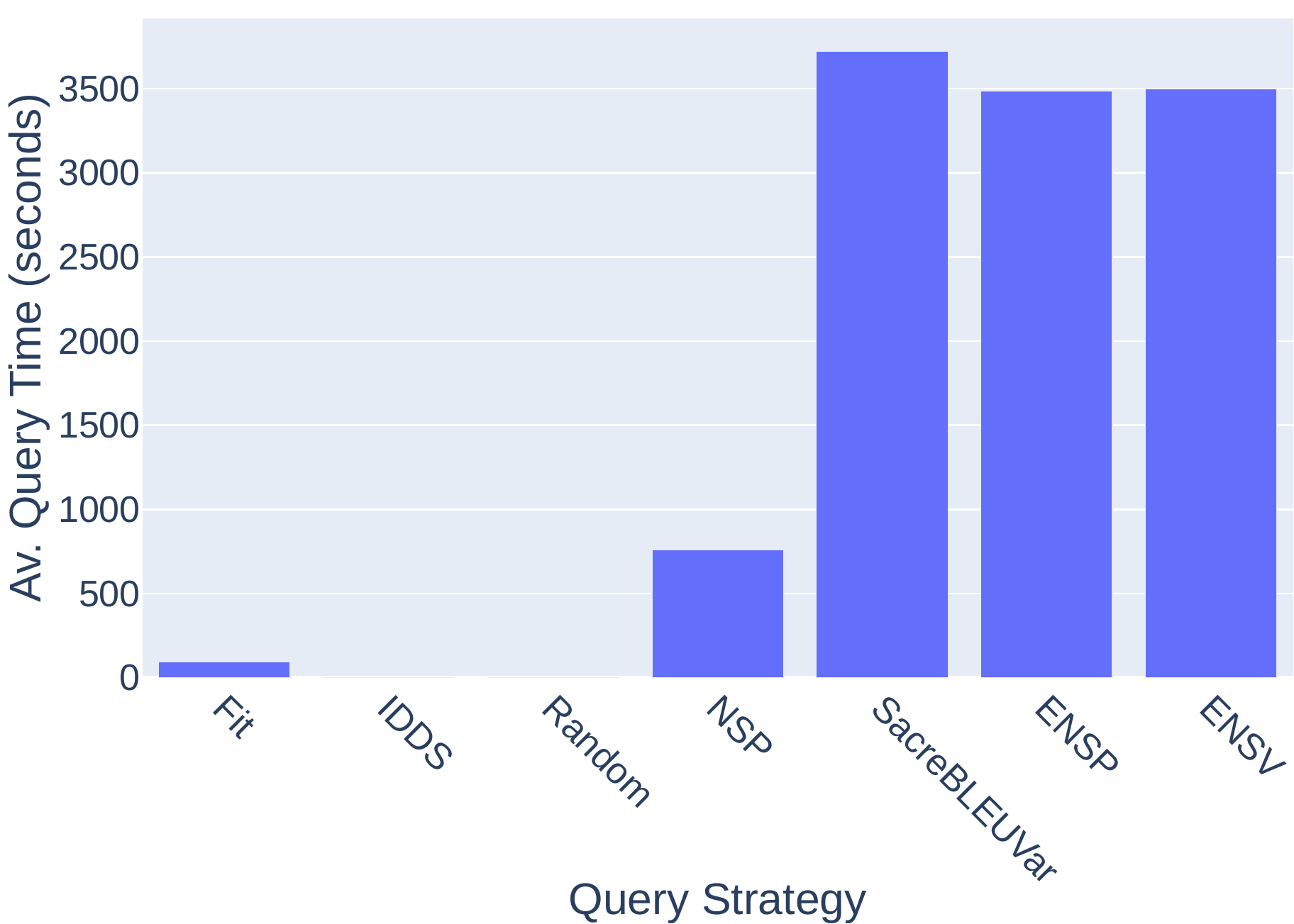}}
	\caption{Average duration in seconds of one AL query of 10 instances with different strategies on the AESLC dataset with BART-base as an acquisition model. \textit{Train} refers to the average time required for training the model throughout the AL cycle. Hardware configuration: 2 Intel Xeon Platinum 8168, 2.7 GHz,
24 cores CPU; NVIDIA Tesla v100 GPU, 32 Gb of VRAM.}
	\label{fig:time_strategy_aeslc}
	\vspace{-0.1cm}
	
\end{figure}

